%% file: arxiv_spirit.tex
\documentclass[times,twocolumn,final]{elsarticle}
\usepackage{fancyhdr}
\usepackage[ruled,vlined]{algorithm2e}
\usepackage{arxiv}
\usepackage{graphicx}%
\usepackage{multirow}%
\usepackage{amsmath,amssymb,amsfonts}%
\usepackage{amsthm}%
\usepackage{mathrsfs}%
\usepackage[title]{appendix}%
\usepackage{xcolor}%
\usepackage{textcomp}%
\usepackage{manyfoot}%
\usepackage{booktabs}%
\usepackage{algpseudocode}%
\usepackage{listings}%
\usepackage{multirow}
\usepackage{pifont}%
\usepackage{amssymb}
\usepackage{adjustbox}
\usepackage{graphicx}
\usepackage{hyperref}
\usepackage{float}
\usepackage{placeins}
\usepackage{fancyhdr}
\usepackage{dsfont}
\usepackage{threeparttable}
\usepackage{color,colortbl}
\usepackage{makecell}
\definecolor{LightCyan}{rgb}{0.88,1,1}
\newcommand{\cmark}{\ding{51}}
\newcommand{\xmark}{\ding{55}}

\begin{document}
\fancypagestyle{firstpagestyle}{
    \fancyhf{} 
    \renewcommand{\headrulewidth}{0pt} 
    \fancyhead{} 
    \fancyfoot{} 
    \fancyhead[CO]{\em \fontsize{9pt}{8pt}\selectfont}
}

\fancypagestyle{default}{
    \fancyhf{}
    \fancyhead[R]{\thepage} 
    \renewcommand{\headrulewidth}{0.4pt} 
    \fancyhead[LO]{\em \fontsize{9pt}{8pt}\selectfont}
}


    
\verso{Saurav Sharma \textit{et~al.}}
\title{{\bf SPIRIT: Spatio-temporal Pairwise Relational Modeling of Instrument-Tissue Interactions for Surgical Action Triplet Recognition}}
\author[1,2]{ 
Saurav \snm{Sharma}} \corref{cor1} \cortext[cor1]{Corresponding author: saurav.sharma@ihu-strasbourg.eu
}
\author[1,2]{Lorenzo \snm{Arboit}}
\author[1,2]{Nabani \snm{Banik}}
\author[1,3]{Sarah \snm{Meuli}}
\author[1,2]{Julia \snm{Alekseenko}}
\author[4]{Jan \snm{Liechti}}
\author[4]{Franziska \snm{Heitzinger}}
\author[1,5]{Michela \snm{Orsi}}
\author[2,6]{Didier \snm{Mutter}}
\author[7]{Daniel \snm{Gero}}
\author[8]{Philipp C. \snm{Nett}}
\author[4,9]{Beat P. \snm{Müller}}
\author[4,9]{Joël L. \snm{Lavanchy}\fnref{fnlast}}
\author[1,2]{Nicolas \snm{Padoy}\fnref{fnlast}}
\fntext[fnlast]{Shared last authorship.}
\address[1]{University of Strasbourg, CNRS, INSERM, ICube, UMR7357, France}
\address[2]{IHU Strasbourg, France}
\address[3]{University of Pavia, Italy}
\address[4]{University Digestive Health Care Center – Clarunis, Basel, Switzerland}
\address[5]{Department of Surgery, Università di Roma Tor Vergata, Rome, Italy}
\address[6]{University Hospital of Strasbourg, France}
\address[7]{University Hospital Zurich, Switzerland}
\address[8]{University Hospital Bern, Switzerland}
\address[9]{Department of Biomedical Engineering, University of Basel, Allschwil, Switzerland}
\received{: }
\finalform{:}
\accepted{:}
\availableonline{:}
\communicated{:}

\input{sections/00-abstract}

\maketitle
\thispagestyle{firstpagestyle}

\input{sections/01-introduction}
\input{sections/02-literature}

\input{sections/03a-dataset}

\input{sections/03b-methods}

\input{sections/04-experiments}

\input{sections/05-results}
\input{sections/06-conclusion}
\input{sections/07-acknowledgement}

\bibliographystyle{model2-names.bst}\biboptions{authoryear}
\bibliography{main}


\input{sections/supp}

\end{document}

%% file: sections/00-abstract.tex
\begin{abstract}
Fine-grained understanding of surgical activity is essential for context-aware assistance in the operating room, including safety monitoring, adverse event identification, and skill assessment. Surgical action triplets, defined as tuples of the form $\langle \emph{instrument}, \emph{verb}, \emph{target} \rangle$, provide a structured description of instrument-tissue interactions. A key open problem, however, is how to learn triplet representations that remain reliable across institutions, where surgical video varies in acquisition conditions, surgeon style, tool usage, and tissue handling, while existing triplet datasets do not support explicit evaluation of center-wise transfer.
To address this problem, we propose \textbf{SPIRIT}, a structured framework for surgical action triplet recognition designed to learn interaction representations that transfer more reliably across centers. Instead of treating each triplet as a flat class label, SPIRIT first learns spatio-temporal representations for instruments, verbs, and targets, then models their pairwise relations, and finally composes them into coherent triplet predictions, with multi-head distillation used to stabilize learning.
To evaluate this setting, we establish \textbf{MultiBypass-4C-T40}, a multi-centric dataset for dense surgical action triplet recognition in Roux-en-Y gastric bypass across four geographically distinct centers, with auxiliary phase and step annotations. Across multiple evaluation protocols, SPIRIT consistently outperforms strong recent baselines, highlighting the value of explicit relational reasoning for multi-centric triplet recognition. Code will be available at \url{https://github.com/CAMMA-public/multibypass-4c-t40}.

\end{abstract}


%% file: sections/01-introduction.tex
\section{Introduction}

Surgical data science seeks to build computational systems that can interpret the intraoperative scene and support safer, more effective interventions~\citep{maier2017surgical}. For many downstream applications, however, coarse workflow labels are insufficient. Tasks such as safety monitoring, adverse event analysis, and fine-grained skill assessment require knowing not only the stage of a procedure, but also which instrument is acting, what action it is performing, and which anatomical structure is involved. This need has driven surgical workflow analysis from phase and step recognition~\citep{twinanda2016endonet,czempiel2020tecno,jin2028svrcnet,funke2018temporal,ramesh2021multi,Lavanchy2024} toward surgical action triplet recognition~\citep{nwoye2022rendezvous,nwoye2020recognition}, where each interaction is represented as an $\langle \emph{instrument}, \emph{verb}, \emph{target} \rangle$ triplet.

Despite this progress, a central question remains unresolved: \emph{how well do triplet recognition models generalize across centers?} This is a critical gap because surgical video is not collected under a single controlled distribution. Across hospitals, substantial variation arises from differences in imaging systems, resolution and color characteristics, trocar placement, camera navigation, surgeon-specific dissection style, preferred instruments, exposure strategy, and tissue handling. These factors change not only how the scene looks, but also how the same surgical action is expressed visually over time. Prior studies have already shown that such variation degrades coarse-grained workflow recognition in both laparoscopic cholecystectomy~\citep{kassem2022federated} and Roux-en-Y gastric bypass~\citep{Lavanchy2024}. For triplet recognition, the challenge is likely more severe, because the task depends on distinguishing subtle interaction-level differences rather than broad procedural stages. Yet current triplet benchmarks do not allow this question to be studied rigorously. CholecT45~\citep{nwoye2020recognition} and CholecT50~\citep{nwoye2022rendezvous} established the task in laparoscopic cholecystectomy, but focus on a relatively standardized procedure and do not support systematic evaluation under institutional variation. ProstaTD~\citep{chen2026prostatd} broadened fine-grained interaction annotation to robot-assisted radical prostatectomy, but was not introduced as a benchmark for multi-centric generalization. Consequently, the field still lacks a benchmark specifically designed to test whether surgical action triplet recognition remains reliable when models are deployed beyond the centers from which they were trained.

This missing evaluation setting is especially consequential in Roux-en-Y gastric bypass (RYGB), a long and procedurally complex laparoscopic operation with dense, repeated, and temporally extended instrument-tissue interactions. Unlike shorter and more standardized procedures, RYGB contains many visually similar yet semantically distinct actions whose interpretation depends on context, temporal evolution, and the precise relation between instrument and tissue. A model may correctly recognize the presence of a grasper, a stapler, or the stomach, yet still fail to determine whether the instrument is retracting, dissecting, clipping, or stapling, and on which structure the action is being applied. This makes RYGB particularly demanding for triplet recognition under center shift. Prior work on MultiBypass140~\citep{Lavanchy2024} showed that even phase and step recognition deteriorate under multi-centric variation in this procedure; at the triplet level, where the model must resolve fine-grained relational structure, the problem is expected to be harder still.
This leads to two research questions. \textbf{RQ1:} \emph{What representation and modeling strategy best preserves the semantics of surgical action triplets while enabling reliable transfer to unseen institutions?} \textbf{RQ2:} \emph{How should such transfer be benchmarked so that robustness to realistic institutional variation can be evaluated explicitly in a multi-centric setting?}

We argue that these two questions are tightly connected. If the goal is to learn triplet representations that transfer reliably across institutions, then the model should not treat surgical action triplets as flat output labels. A triplet is inherently compositional: it depends on how an instrument interacts with tissue over time and on whether the implied action is compatible with the instrument itself. In this sense, the triplet label reflects a structured interaction rather than a single isolated visual category. In particular, instrument-target relations encode the semantics of physical interaction, while instrument-verb relations capture affordance constraints that help distinguish actions with similar local appearance. Under center shift, such relational structure may provide a more stable basis for recognition than appearance cues alone, which are more easily affected by differences in imaging, exposure, and surgical style. This suggests that approaches based primarily on direct triplet prediction may be poorly matched to the transfer problem, whereas models that explicitly organize and reason over intermediate relations may be better suited for robust multi-centric generalization.

Motivated by this perspective, we propose \textbf{SPIRIT}, a structured framework for surgical action triplet recognition based on \textbf{S}patio-temporal \textbf{P}airw\textbf{I}se \textbf{R}elational modeling of \textbf{I}nstrument-\textbf{T}issue interactions. The central idea is that robust triplet recognition should be learned progressively, from identifying the components of an interaction to modeling how they relate and combine. SPIRIT therefore consists of three modules. The first is a \textbf{text-conditioned unary feature learning (TUF)} module, which learns spatio-temporal representations for instruments, verbs, and targets from temporally evolving visual evidence. The second is a \textbf{pairwise interaction coupling (PIC)} module, which refines these unary features by explicitly modeling the two key relation spaces of surgical actions: \emph{instrument-target} relations, which encode where the physical interaction occurs, and \emph{instrument-verb} relations, which encode what action the instrument can plausibly perform through its affordance. The third is a \textbf{triplet graph reasoning (TGR)} module, which treats the learned instrument-target and instrument-verb relations as nodes in a higher-order graph and composes them into coherent instrument-verb-target predictions. By organizing recognition in this way, SPIRIT is designed to learn interaction representations whose semantics remain more stable under the ambiguity and variability introduced by cross-center surgical data.

\begin{figure*}[!htbp]
\centering
    \includegraphics[width=0.80\textwidth]{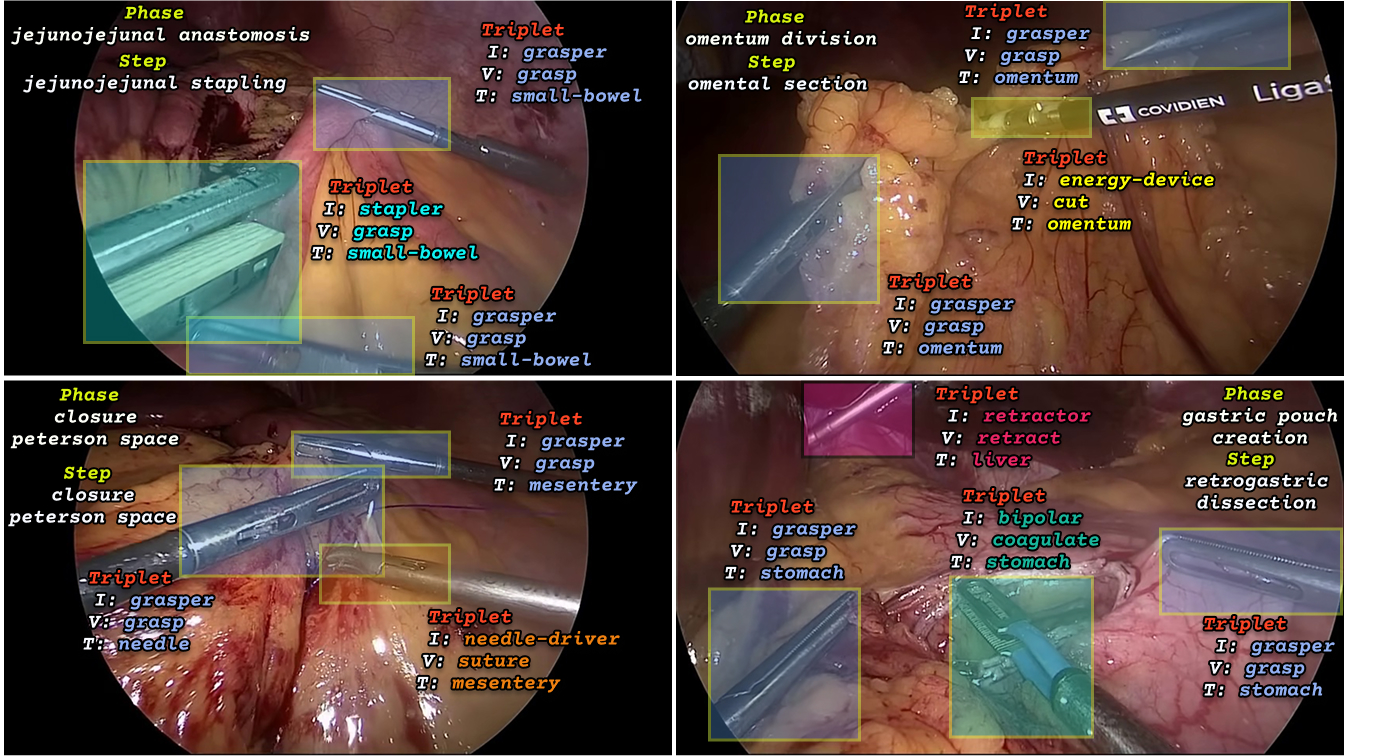} 
    \caption{Illustrative examples from the MultiBypass-4C-T40 dataset showing surgical action triplets together with accompanying phase and step-level labels for fine-grained surgical workflow analysis.}
    \label{fig:dataset_overview}
\end{figure*}

To evaluate this hypothesis, we introduce \textbf{MultiBypass-4C-T40}, a new multi-centric dataset and benchmark for dense surgical action triplet recognition in RYGB, collected across four hospitals and geographic sites. It extends prior multi-centric workflow analysis in RYGB from phase- and step-level recognition to fine-grained action triplets, while retaining phase and step annotations for hierarchical analysis. Its purpose is not merely to provide denser labels, but to enable explicit evaluation of whether triplet recognition methods learn representations that transfer reliably across institutions.

We perform extensive experiments on MultiBypass-4C-T40 under three complementary evaluation settings: the \emph{all-centers} split, which measures standard performance under a fixed multi-centric train-validation-test protocol; the \emph{cross-validation} split, which evaluates robustness across different train-validation partitions; and the \emph{challenge} split, which includes hidden-test evaluation and transfer to unseen centers. Across all three settings, SPIRIT consistently outperforms strong triplet-specific baselines and recent surgical foundation models. The gains are especially pronounced at the pairwise and final triplet levels, supporting the central hypothesis of this work: explicitly modeling intermediate interaction structure yields triplet representations that transfer more reliably across institutions than approaches based primarily on direct triplet prediction from visual appearance alone.




Our contributions are as follows:
\begin{itemize}
\item We propose \textbf{SPIRIT}, a structured framework for surgical action triplet recognition that improves transfer to unseen centers by explicitly modeling unary components, pairwise interaction relations, and triplet composition.

\item We introduce \textbf{MultiBypass-4C-T40}, a multi-centric dataset and benchmark for surgical action triplet recognition in Roux-en-Y gastric bypass, with accompanying phase- and step-level annotations across four geographically distinct centers.

\item We define evaluation protocols that measure both standard triplet recognition performance and robustness to realistic institutional variation, including unseen-center transfer.

\item We show through extensive experiments that SPIRIT consistently outperforms strong triplet-specific baselines and recent surgical foundation models, with especially strong gains in pairwise reasoning and final triplet prediction under center shift.
\end{itemize}

%% file: sections/02-literature.tex
\section{Related work}
\subsection{Surgical Workflow Analysis}
Automatic modelling of surgical workflow has been a central topic in surgical data science because it provides the temporal context required for downstream assistance, monitoring and intraoperative decision support~\citep{maier2017surgical,vercauteren2019cai4cai}. Early work approached this problem through statistical modelling of surgical process signals, often relying on instrument usage patterns as indirect indicators of procedural progress~\citep{padoy2012statistical,blum2008workflow}. In these settings, frameworks based on Hidden Markov Models (HMMs) were commonly used to infer surgical phases, with expert annotations serving as clinical reference.
With the rise of deep learning, surgical workflow analysis increasingly shifted toward direct visual interpretation of endoscopic video. A series of methods demonstrated that phase recognition can be learned from richer scene information rather than from instrument presence alone, leading to substantial improvements in automated workflow understanding~\citep{twinanda2016endonet,dergachyova2016automatic,czempiel2020tecno,funke2018temporal,jin2028svrcnet}. More recent studies have further explored self-supervised representation learning, large-scale pretraining and dataset expansion to improve robustness and transferability, particularly in the context of phase recognition~\citep{ramesh2023dissecting,batic2024endovit,jaspers2025scaling}.
Although phase recognition captures global procedural progression, it does not resolve the finer structure of surgical activity. To address this limitation, later work introduced surgical steps as a more localized description of workflow, corresponding to the intermediate units that compose each phase~\citep{ramesh2021multi,Lavanchy2024,ayobi2024pixelwise,valderrama2022towards,huaulme2021micro}. Step recognition offers a more detailed account of procedural organization and has improved the granularity at which surgical progress can be analyzed.
Nevertheless, both phases and steps remain relatively coarse abstractions of the intraoperative scene. They describe procedural progress, but not the precise interaction between tools and anatomy. Surgical action triplets move to a finer semantic level by explicitly representing the instrument, the action being performed and the anatomical target~\citep{nwoye2022rendezvous}. In this sense, surgical workflow analysis has evolved along a hierarchy of increasing granularity, from phases to steps and ultimately to triplets, each level offering a progressively more detailed representation of operative activity.

\subsection{Surgical Action Triplets}
Fine-grained surgical understanding depends on recognizing how instruments interact with anatomical structures over time. Early efforts to formalize such interactions were grounded in surgical ontologies, in which activities were described through actions performed by specific instruments on specific targets~\citep{neumuth2009validation,katic2014knowledge}. In laparoscopic cholecystectomy, this concept was later operationalized as the \textlangle{}{\textit{instrument, verb, target}}\textrangle{} triplet, which provides a compact and semantically grounded representation of surgical activity.

The first major benchmark for this formulation was CholecT40~\citep{nwoye2020recognition}, which introduced frame-level triplet annotations for 40 videos together with TripNet, a multi-task architecture that incorporated weak instrument localization cues. This benchmark was later extended to CholecT50~\citep{nwoye2022rendezvous}, which increased the dataset size to 50 videos and was accompanied by Rendezvous~\citep{nwoye2022rendezvous}, an attention-based framework that jointly reasons over instrument, verb and target representations for triplet prediction. The MICCAI CholecTriplet challenges further established surgical action triplet recognition as a benchmark task for causal frame-level understanding.
Building on these foundations, several methods have sought to enrich triplet prediction by incorporating temporal context or spatial instance information. Rendezvous-in-Time~\citep{sharma2023rendezvous} extends triplet recognition with temporal modeling to better capture the evolution of surgical interactions over time. In a related direction, MCIT-IG~\citep{ssharma2023mcitig} addresses the absence of instance-level supervision by combining instrument detection with graph-based association between detected instrument instances and target embeddings, enabling triplet detection on the instance-level benchmark of the MICCAI 2022 CholecTriplet challenge~\citep{nwoye2023cholectriplet2022}.
Another line of work has focused on the optimization challenges posed by triplet recognition, particularly severe class imbalance and the long-tail distribution of interaction classes. SelfD~\citep{yamlahi2023self} introduced self-distillation to improve learning under imbalanced triplet labels, and subsequent work extended this idea through teacher-selection and ensemble-based distillation strategies~\citep{yamlahi2025smarter}. MT4MTL-KD~\citep{gui2023mt4mtl} similarly explored teacher-student distillation across spatial and temporal learning stages. Beyond distillation, TDN~\citep{chen2023surgical} proposed disentanglement-oriented supervision through auxiliary soft labels that capture interaction-level properties, whereas TERL~\citep{gui2024tail} used contrastive learning to improve recognition of tail classes with limited training samples. More recently, CurConMix~\citep{jeon2025curconmix} introduced a curriculum-based compositional learning strategy that progressively learns from targets to instrument-target pairs and then to full triplets, complemented by hard negative pair sampling, distillation and feature mixup. Taken together, these methods indicate that triplet prediction benefits from supervision beyond flat classification, although explicit modeling of intermediate pairwise interaction structure remains comparatively limited.

\textbf{Extensions to other surgical domains.} Related formulations have also been explored beyond laparoscopic cholecystectomy, although often with different annotation granularities. In robot-assisted radical prostatectomy, SARAS-ESAD~\citep{bawa2021saras} represents interactions using \textlangle{}{\textit{verb, target}}\textrangle{} pairs with bounding-box supervision, whereas other prostatectomy datasets focus on action recognition or pixel-level workflow analysis rather than full triplet recognition~\citep{ayobi2024pixelwise,valderrama2022towards}. In cataract surgery, triplet-style annotations have likewise been combined with spatial localization cues~\citep{lin2022instrument}. Collectively, these studies reflect growing interest in fine-grained surgical interaction modeling, while also showing that current benchmarks remain limited in procedural scope, annotation setting and evaluation under institutional variation.

\begin{figure*}[!htbp]
\centering
    \includegraphics[width=2.05\columnwidth]{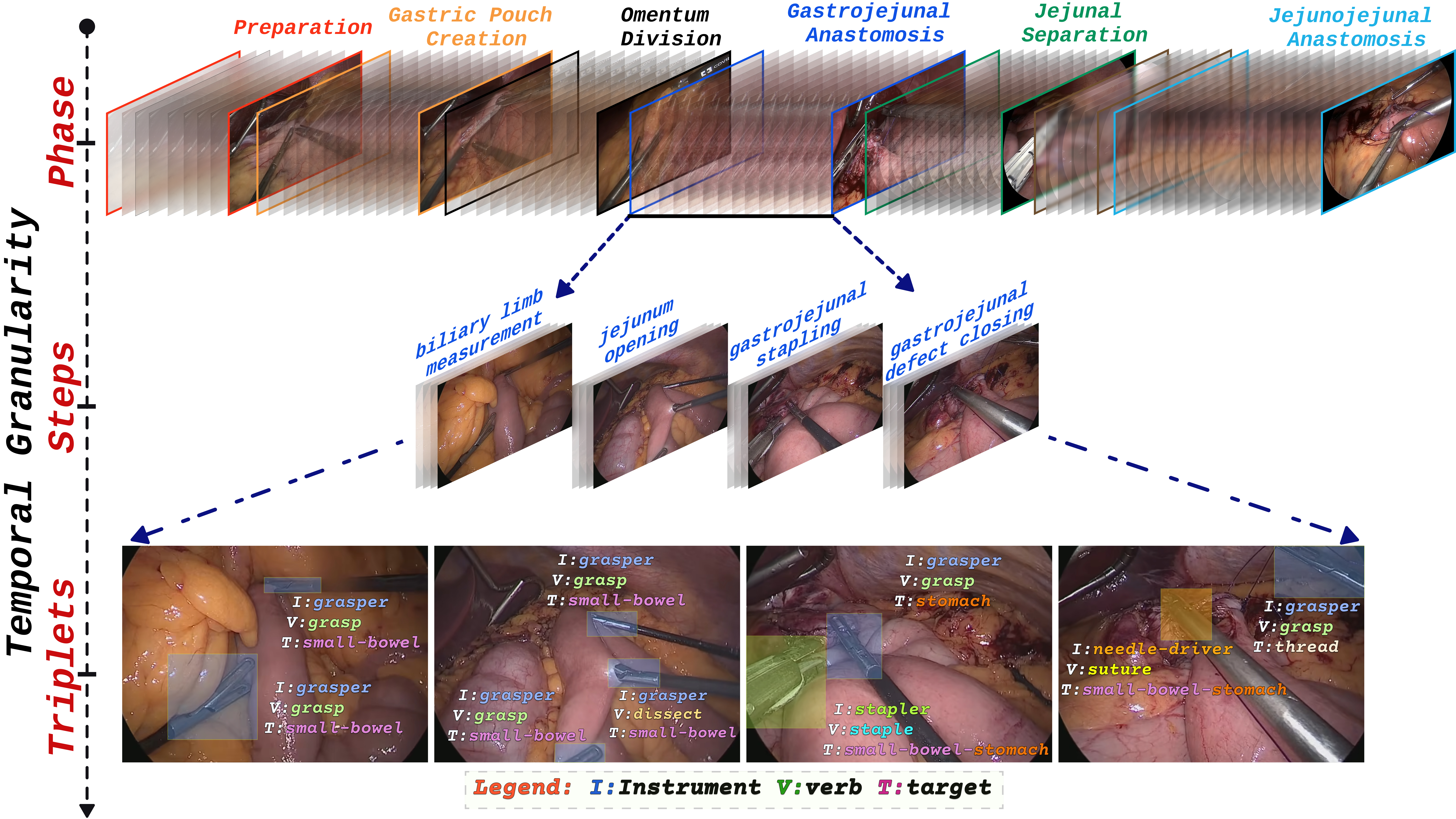} 
    \caption{Illustration of the temporal evolution of hierarchical workflow annotations, from phases and steps to fine-grained surgical action triplets, in Roux-en-Y gastric bypass. The figure shows how labels at different semantic levels are temporally aligned within the same surgical sequence.}
    \label{fig:dataset_granularity}
\end{figure*}

\subsection{Multi-centric Generalization}

The generalizability of surgical video models has become increasingly important as the field moves beyond retrospective benchmarking toward clinically useful deployment. Models developed on narrowly curated datasets may perform well in-distribution while remaining sensitive to differences in institutional practice, acquisition characteristics and procedural style. This has motivated growing interest in evaluating surgical video understanding under realistic distribution shifts rather than relying solely on random train-test splits from a single source.

Many widely used surgical video benchmarks were collected from a single institution or a tightly controlled setting. More recent efforts have begun to address this limitation at coarser levels of workflow analysis and quality assessment. FedCy \citep{kassem2022federated} introduced a large international multicenter dataset for surgical phase recognition with 180 videos collected from five hospitals, highlighting the importance of cross-site learning for workflow analysis. In laparoscopic cholecystectomy, the SAGES Critical View of Safety challenge \citep{alapatt2025sages} further expanded this direction by assembling a globally sourced benchmark for AI-assisted surgical quality assessment, emphasizing robustness to heterogeneous clinical variability. In Roux-en-Y gastric bypass, MultiBypass140~\citep{Lavanchy2024} established multicenter phase and step recognition across two centers and showed that center shift degrades performance even for coarse semantic tasks.

This concern has also begun to shape methodological work. For example, recent domain-transfer studies \citep{satyanaik2024optimizing} have explicitly examined zero-shot transfer of surgical scene models across centers, showing that distribution shift materially affects performance and motivating approaches that improve robustness under unseen domains. Related efforts have also expanded fine-grained annotation in other domains: ProstaTD~\citep{chen2026prostatd}, for example, introduced triplet detection instances for robot-assisted radical prostatectomy using videos aggregated from three data sources, but did not study multicenter triplet generalization explicitly. Taken together, these studies show that multicenter variation is a central challenge in surgical AI, yet this question remains insufficiently explored for surgical action triplet recognition in complex laparoscopic procedures such as Roux-en-Y gastric bypass.

\subsection{Relational Modeling}
Recent work in surgical AI has increasingly explored relational representations of the surgical scene, motivated by the idea that workflow understanding depends not only on visual appearance and temporal dynamics, but also on structured interactions among tools, anatomy and semantic workflow entities. In surgical workflow analysis, MURPHY~\citep{zhao2022murphy} showed that modeling relations within hierarchical annotations improves prediction. Related scene-graph approaches likewise represent tools, anatomy and their interactions explicitly for surgical scene understanding and workflow prediction~\citep{holm2023dynamic,koksal2024sangria,satyanaik2024optimizing}.

A related line of work has focused on \emph{instance-centric} relational representations derived from detections or segmentations. Endoscapes~\citep{murali2023latent}, CAT-SG~\citep{holm2025cat} and Endoscapes-SG201~\citep{shin2025towards} demonstrate the value of modeling component instances and their relations for fine-grained surgical understanding. However, these approaches rely on explicit instance-level supervision and graph construction. By contrast, surgical action triplet recognition requires reasoning over interaction semantics from spatio-temporal video features, often without reliable instance-level annotations.

Within triplet analysis, structured learning has so far been explored more selectively. MCIT-IG~\citep{ssharma2023mcitig} introduced graph-based association for triplet detection, and CurConMix~\citep{jeon2025curconmix} exploited compositional structure through curriculum learning over targets, pairs and triplets. These studies suggest that triplet prediction benefits from supervision beyond flat classification, but explicit reasoning over intermediate pairwise interaction structure before final triplet inference remains comparatively underexplored, especially in multi-centric settings.

%% file: sections/03a-dataset.tex

\section{Dataset}
In this section, we present the MultiBypass-4C-T40 dataset and its design rationale. We first describe the acquisition process and the annotation protocol used to define fine-grained surgical action triplets. We then introduce the annotation process and the label mediation strategy used to consolidate a broad set of raw interaction labels into a clinically meaningful triplet vocabulary. Next, we position MultiBypass-4C-T40 relative to existing datasets for fine-grained surgical workflow analysis, and finally summarize the split design and dataset statistics. Figure~\ref{fig:dataset_granularity} provides an overview of the temporal relationship between the hierarchical labels in MultiBypass-4C-T40, highlighting how phase, step and triplet-level annotations evolve jointly over time within the same surgical sequence.

\subsection{Objective}
Existing datasets for surgical action triplet analysis are either effectively single-center or provide only limited diversity in the number of centers and videos available for benchmark evaluation. This restricts the ability to assess how fine-grained surgical recognition models behave under variation in acquisition conditions, camera framing and surgeon-dependent interaction patterns. To address this gap, we introduce \textbf{MultiBypass-4C-T40}, a new dataset for surgical action triplet recognition in Roux-en-Y gastric bypass spanning four centers with diverse procedural styles. In addition to triplet annotations, MultiBypass-4C-T40 also provides phase and step labels, following the MultiBypass140 protocol~\citep{Lavanchy2024}, to support a more comprehensive evaluation of surgical workflow understanding across multiple levels of semantic granularity. 

\begin{table}[!htbp]
    \centering
    \caption{Comparison of datasets for workflow and triplet analysis. Proc.: procedure; MC: multi-centric; P/S/T: phase, step and triplet; \#Vid: number of videos; \#Cls: number of triplet classes; \#Inst: number of triplet instances. LC: laparoscopic cholecystectomy; RARP: robot-assisted radical prostatectomy; RYGB: Roux-en-Y gastric bypass.}
    \label{tab:dataset_comparison}
    \setlength{\tabcolsep}{4pt}
    \resizebox{\columnwidth}{!}{%
    \begin{tabular}{lcccccccc}
        \toprule
        \textbf{Dataset} & \textbf{Proc.} & \textbf{MC} & \textbf{P} & \textbf{S} & \textbf{T} & \textbf{\#Vid} & \textbf{\#Cls} & \textbf{\#Inst} \\
        \midrule
        CholecT50 & LC & \xmark & \cmark & \xmark & \cmark & 50 & 100 & 150532 \\
        MultiBypass140 & RYGB & \cmark & \cmark & \cmark & \xmark & 140 & -- & -- \\
        ProstaTD & RARP & \cmark & \xmark & \xmark & \cmark & 21 & 89 & 196490 \\
        MultiBypass-4C-T40 & RYGB & \cmark & \cmark & \cmark & \cmark & 40 & 85 & 420911 \\
        \bottomrule
    \end{tabular}%
    }
\end{table}

Table~\ref{tab:dataset_comparison} summarizes the position of MultiBypass-4C-T40 relative to existing datasets for surgical workflow and triplet analysis. In contrast to CholecT50 \citep{nwoye2022rendezvous}, which provides triplet annotations for laparoscopic cholecystectomy in a single-center setting, and MultiBypass140 \citep{Lavanchy2024}, which provides multi-centric phase and step annotations without triplets, MultiBypass-4C-T40 combines multi-centric acquisition with hierarchical phase, step and triplet labels in Roux-en-Y gastric bypass. Compared with ProstaTD \citep{chen2026prostatd}, it also provides a larger number of triplet instances while retaining explicit workflow annotations, thereby enabling a broader evaluation of fine-grained surgical scene understanding across multiple levels of semantic granularity.

\subsection{Data Acquisition}
The MultiBypass-4C-T40 dataset comprises surgical videos collected from four centers: University Hospital of Strasbourg, France (C1), Bern University Hospital, Bern, Switzerland (C2), University Digestive Health Care Center--Clarunis, University Hospital of Basel, Basel, Switzerland (C3), and University Hospital of Zurich, Zurich, Switzerland (C4). At each site, patients who underwent Roux-en-Y gastric bypass surgery were retrospectively selected from institutional databases. 
The corresponding surgical videos were then retrospectively retrieved from institutional repositories, and associated clinical variables were collected from the electronic health records, except for C1. These variables included age, sex, body-mass index, date and duration of surgery, as well as outcome measures such as hospital length of stay and postoperative complications.
Before inclusion in the dataset, all retrieved data were fully de-identified and all videos were anonymized to remove patient and institution-specific information, thereby minimizing any possibility of information leakage during dataset construction, annotation, or downstream analysis.
The study was conducted following institutional review board approval, with waiver of informed consent (\#2021-01666). For videos from C1, no separate IRB approval was required because the data were considered anonymous.
All videos were recorded at their native frame rates and spatial resolutions using endoscopic camera systems. Native acquisition settings differed between sites, reflecting real-world variation in recording pipelines. For analysis, all videos were standardized to 25 frames per second and a spatial resolution of 480p.
After de-identification, anonymization, and curation, the video files were imported into the in-house MoSaiC annotation platform~\citep{mazellier2023mosaic}, where triplet annotation was performed according to the protocol described next.

\begin{algorithm}[!h]
\caption{Temporal annotation protocol for surgical action triplets}
\label{alg:triplet_annotation_clean}
\KwIn{Surgical video $V$}
\KwOut{Annotated triplet intervals $\mathcal{T}$}

Initialize $\mathcal{T} \leftarrow \emptyset$\;

\ForEach{candidate surgical action in $V$}{

    \textbf{Step 1: Identify the action apex.}\;
    
    Select the time point or interval where the triplet 
    $\langle$instrument, verb, target$\rangle$ is most visually unambiguous\;

    \textbf{Step 2: Define the initial temporal window.}\;

    Set the preliminary start to $\sim 3\,\mathrm{s}$ before the apex\;
    
    Set the preliminary end to $\sim 3\,\mathrm{s}$ after the apex\;

    \textbf{Step 3: Refine the start boundary.}\;
    
    If the instrument is not visible at the preliminary start, move the start to the first frame where the instrument becomes visible before the interaction\;
    
    If another action is still ongoing, move the start to immediately after the previous action ends\;

    \textbf{Step 4: Refine the end boundary.}\;
    
    If the instrument becomes non-visible before the preliminary end, move the end to the last frame where the interaction remains visually supported\;
    
    If a new action starts before the preliminary end, move the end to the transition point\;

    \textbf{Step 5: Apply continuity rules.}\;
    
    Preserve the interval if the instrument is briefly occluded but the action remains visually evident\;
    
    Preserve the interval if the instrument exits the field for less than $\sim 3\,\mathrm{s}$ and the same static action remains visually implied\;
    
    Terminate the interval if the instrument is out of frame for more than $\sim 3\,\mathrm{s}$ or if visual evidence no longer supports continuation\;

    \textbf{Step 6: Assign the triplet label.}\;
    
    Assign the corresponding $\langle$instrument, verb, target$\rangle$ label and add the interval to $\mathcal{T}$\;
}
\Return{$\mathcal{T}$}\;
\end{algorithm}

\subsection{Annotation Protocol}
To annotate surgical action triplets, we developed a dedicated protocol through a multi-year effort involving iterative protocol design, pilot labeling, and repeated expert review. We adapted the protocols used in CholecT50~\citep{nwoye2022rendezvous} and ProstaTD ~\citep{chen2026prostatd}. to reflect our view that surgical actions are inherently \emph{temporal events}, rather than isolated moments of maximal contact between an instrument and an anatomical structure. Event perception research has shown that human observers parse ongoing activity into temporally extended events with meaningful onset, execution and offset phases~\citep{zacks2001human,zacks2007event,zacks2020event}, and a similar perspective is reflected in computer vision benchmarks for everyday activity understanding, where actions are typically annotated as temporal segments with explicit start and end times~\citep{damen2018scaling,Damen2021PAMI}. 
Guided by this view, annotators first identify the apex of an interaction, defined as the phase of strongest and clearest instrument-tissue contact, and then extend the interval backward and forward to capture the beginning and end of the action. In standard cases, the annotated interval spans approximately 3\,s before and 3\,s after the apex, yielding a temporally smooth interaction envelope that retains the approach, peak interaction, and disengagement phases relevant for learning fine-grained surgical actions. The full pipeline is described in Algorithm~\ref{alg:triplet_annotation_clean}.

The annotation process was carried out by surgical experts involved in both clinical practice and surgical data science research. Two medical students and two surgical residents underwent structured training in surgical triplet annotation following the protocol and review process. Their annotations were supervised by a clinical expert in surgical data science, with final mediation and adjudication performed by an expert surgeon.

\subsection{Label Mediation and Consolidation}
Label mediation and consolidation were conducted through a sustained multi-stage process, since raw triplet annotations frequently contained ambiguities that could not be resolved through guideline application alone. The most common ambiguities arose from two sources: the instrument identity and the interaction type expressed by the verb. These cases were reviewed and mediated by an expert surgeon. After this initial mediation stage, we obtained a provisional pool of 212 triplet labels.

This pool still included triplets that were overly center-specific, such as those involving \emph{circular stapler}, as well as triplets that occurred only very rarely, i.e. with fewer than 10 instances. A further aim of consolidation was to remove \emph{dormant} triplets, namely interactions that were technically present but not clinically informative for fine-grained action understanding. For example, \emph{\textlangle{}retractor, retract, liver\textrangle{}} was excluded because it consistently appeared as a static background interaction rather than a meaningful procedural action. In the next stage, two medical experts independently performed a binary relevance assessment for each remaining triplet. Triplets receiving at least one positive judgment were then forwarded to a final verification round conducted by the domain-specific clinical expert responsible for mediation. Through this consolidation process, the final vocabulary was reduced to 85 triplet classes, comprising 12 instrument classes, 13 verb classes, and 15 target classes.

\begin{table*}[h]
\centering
\caption{Dataset statistics showing the number of occurrences of the triplets.}
\label{tab:mbt40_triplet_freq}
\small
\setlength{\tabcolsep}{4pt}
\begin{tabular}{p{0.24\linewidth}r p{0.24\linewidth}r p{0.24\linewidth}r}
\toprule
\textbf{Triplet} & \textbf{Count} & \textbf{Triplet} & \textbf{Count} & \textbf{Triplet} & \textbf{Count} \\
\midrule
grasper,grasp,small\_bowel & 59,984 & grasper,grasp,thread & 50,084 & grasper,grasp,stomach & 49,470 \\
grasper,null\_verb,null\_target & 30,137 & needle\_driver,suture,SBS-A & 27,997 & grasper,grasp,omentum & 25,387 \\
needle\_driver,suture,SB-A & 20,377 & grasper,suture,SBS-A & 17,759 & grasper,suture,SB-A & 12,009 \\
needle\_driver,suture,mesentery & 9,805 & grasper,grasp,needle & 7,781 & grasper,retract,stomach & 7,602 \\
IA,aspirate,fluid & 6,835 & grasper,grasp,mesentery & 6,722 & energy\_device,cut,omentum & 6,248 \\
grasper,suture,mesentery & 5,660 & stapler,grasp,small\_bowel & 5,067 & grasper,retract,small\_bowel & 4,463 \\
grasper,dissect,stomach & 4,458 & stapler,grasp,stomach & 4,283 & grasper,retract,omentum & 4,178 \\
gauze,clean,stomach & 4,035 & grasper,retract,liver & 3,239 & energy\_device,dissect,stomach & 2,873 \\
stapler,staple,stomach & 2,661 & energy\_device,dissect,omentum & 2,539 & grasper,dissect,omentum & 2,334 \\
needle\_driver,grasp,needle & 2,295 & stapler,null\_verb,null\_target & 2,209 & energy\_device,null\_verb,null\_target & 2,195 \\
needle\_driver,grasp,thread & 1,791 & stapler,grasp,SB-A & 1,371 & electric\_hook,dissect,omentum & 1,283 \\
grasper,retract,mesentery & 1,237 & electric\_hook,dissect,stomach & 1,217 & needle\_driver,suture,small\_bowel & 1,198 \\
bipolar\_forceps,coagulate,stomach & 1,192 & grasper,suture,small\_bowel & 1,118 & energy\_device,cut,mesentery & 937 \\
electric\_hook,dissect,small\_bowel & 921 & grasper,dissect,mesentery & 796 & stapler,grasp,SBS-A & 779 \\
suture\_passer,pass,thread & 749 & electric\_hook,null\_verb,null\_target & 708 & stapler,staple,SB-A & 691 \\
needle\_driver,null\_verb,null\_target & 677 & energy\_device,dissect,small\_bowel & 665 & energy\_device,coagulate,stomach & 637 \\
energy\_device,dissect,mesentery & 607 & energy\_device,retract,omentum & 587 & energy\_device,retract,stomach & 582 \\
grasper,dissect,small\_bowel & 567 & energy\_device,cut,small\_bowel & 552 & energy\_device,cut,adhesion & 543 \\
swab\_forceps,clean,small\_bowel & 542 & RAD,dissect,small\_bowel & 540 & energy\_device,coagulate,omentum & 537 \\
stapler,staple,small\_bowel & 522 & energy\_device,cut,stomach & 501 & swab\_forceps,clean,stomach & 481 \\
clipper,clip,stomach & 437 & IA,null\_verb,null\_target & 432 & IA,irrigate,fluid & 383 \\
stapler,staple,SBS-A & 368 & swab\_forceps,retract,liver & 343 & bipolar\_forceps,null\_verb,null\_target & 339 \\
swab\_forceps,null\_verb,null\_target & 305 & electric\_hook,coagulate,spleen & 272 & RAD,dissect,stomach & 271 \\
RAD,grasp,stomach & 263 & grasper,grasp,colon & 230 & needle\_driver,retract,small\_bowel & 218 \\
IA,retract,small\_bowel & 211 & swab\_forceps,retract,small\_bowel & 190 & grasper,grasp,liver & 180 \\
IA,retract,stomach & 177 & RAD,null\_verb,null\_target & 168 & clipper,null\_verb,null\_target & 135 \\
energy\_device,grasp,omentum & 132 & grasper,retract,sponge & 126 & clipper,clip,small\_bowel & 110 \\
bipolar\_forceps,grasp,sponge & 107 & energy\_device,retract,small\_bowel & 105 & grasper,retract,colon & 98 \\
\cmidrule(lr){5-6}
stapler,retract,stomach & 97 &  &  & \textbf{Total} & \textbf{420,911} \\
\midrule
\multicolumn{6}{p{0.95\linewidth}}{\footnotesize\textit{Acronyms:} SB-A=small\_bowel\_anastomosis, SBS-A=small\_bowel\_stomach\_anastomosis, RAD=right\_angle\_dissector, IA=irrigator\_aspirator.} \\
\bottomrule
\end{tabular}
\end{table*}

\begin{figure*}[!htbp]
\centering
    \includegraphics[width=1.75\columnwidth]{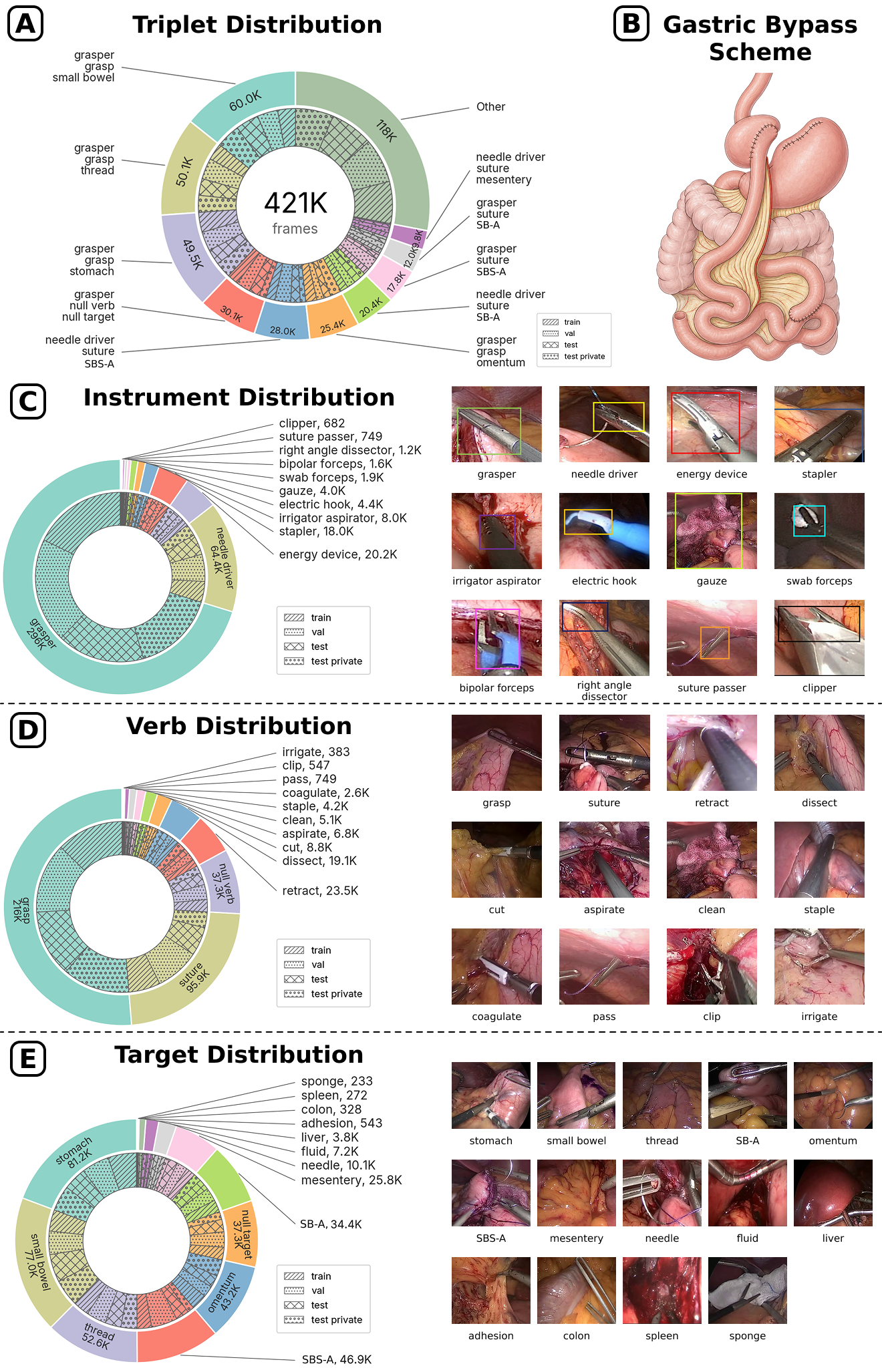} 
    \caption{Overview of the MultiBypass-4C-T40 dataset, illustrating the distribution of fine-grained surgical action triplets and their constituent instrument, verb, and target categories.}
    \label{fig:dataset_overview}
\end{figure*}

\subsection{Dataset Statistics}
Following the protocol adopted in CholecT50~\citep{nwoye2022rendezvous}, we extract frames at 1\,FPS and assign triplet labels across all 40 videos in MultiBypass-4C-T40. Table~\ref{tab:mbt40_triplet_freq} summarizes the frequency of the consolidated triplet labels, while Table~\ref{tab:mbt40_ivt_component_counts} reports the corresponding distributions for instruments, verbs and targets. Together, these statistics show that MultiBypass-4C-T40 spans a broad range of fine-grained surgical interactions, including both frequent actions and rarer long-tail triplets.

The class distribution is notably imbalanced, which is expected in a complex procedure such as Roux-en-Y gastric bypass. A relatively small number of interactions account for a large proportion of the annotations, particularly those involving grasping and retraction of major anatomical structures such as the stomach, small bowel and omentum. At the same time, many clinically meaningful triplets occur much less frequently, often because they are tied to specific procedural stages, uncommon anatomical configurations, or shorter but important instrument maneuvers. This long-tail behavior makes the benchmark more realistic, but also substantially more difficult, as models must learn to recognize both dominant interaction patterns and sparse, less repetitive actions.

The component-level statistics further show that the imbalance is not uniform across the triplet structure. Certain instruments, such as the grasper and needle driver, appear far more often than others, and verbs such as \emph{grasp} and \emph{suture} dominate the action vocabulary. Similarly, targets such as the stomach, small bowel and anastomotic structures are more frequently represented than smaller or less consistently visible anatomical entities. These trends indicate that the difficulty of triplet recognition does not arise only from the number of triplet classes, but also from the uneven composition of the underlying instrument, verb and target spaces. In practice, this means that rare triplets may be difficult either because the full combination is uncommon or because one of their constituent components is itself underrepresented.

Beyond frequency imbalance, the dataset also captures interactions that are temporally extended and compositionally complex. For example, suturing often unfolds over long intervals and may involve coordinated use of multiple instruments, such as a grasper and a needle driver, within the same procedural segment. More broadly, MultiBypass-4C-T40 reflects substantial variation in interaction duration, maneuver style and anatomical context across videos and centers. This distributional and structural diversity makes the dataset not only clinically meaningful, but also methodologically demanding, and further motivates the need for models that can capture fine-grained relational structure rather than relying only on isolated appearance cues. These characteristics are summarized visually in Fig.~\ref{fig:dataset_overview}, which highlights the distribution of triplet annotations and their underlying components across the dataset.

\begin{table}[!htbp]
\centering
\caption{Statistics of the triplet’s component labels in the dataset.}
\label{tab:mbt40_ivt_component_counts}
\scriptsize
\setlength{\tabcolsep}{2pt}
\resizebox{\columnwidth}{!}{%
\begin{tabular}{p{0.15\linewidth}r p{0.15\linewidth}r p{0.15\linewidth}r}
\toprule
\textbf{Instrument} & \textbf{Count} & \textbf{Verb} & \textbf{Count} & \textbf{Target} & \textbf{Count} \\
\midrule
grasper & 295,619 & grasp & 215,926 & stomach & 81,237 \\
needle\_driver & 64,358 & suture & 95,923 & small\_bowel & 76,973 \\
energy\_device & 20,240 & null\_verb & 37,305 & thread & 52,624 \\
stapler & 18,048 & retract & 23,453 & SBS-A & 46,903 \\
IA & 8,038 & dissect & 19,071 & omentum & 43,225 \\
electric\_hook & 4,401 & cut & 8,781 & null\_target & 37,305 \\
gauze & 4,035 & aspirate & 6,835 & SB-A & 34,448 \\
swab\_forceps & 1,861 & clean & 5,058 & mesentery & 25,764 \\
bipolar\_forceps & 1,638 & staple & 4,242 & needle & 10,076 \\
RAD & 1,242 & coagulate & 2,638 & fluid & 7,218 \\
suture\_passer & 749 & pass & 749 & liver & 3,762 \\
clipper & 682 & clip & 547 & adhesion & 543 \\
 &  & irrigate & 383 & colon & 328 \\
 &  &  &  & spleen & 272 \\
 &  &  &  & sponge & 233 \\
\bottomrule
\end{tabular}
}
\begin{flushleft}
{\footnotesize\textit{Acronyms:} SB-A=small\_bowel\_anastomosis, RAD=right\_angle\_dissector, SBS-A=small\_bowel\_stomach\_anastomosis, IA=irrigator\_aspirator.}
\end{flushleft}
\end{table}

\begin{table}[h]
    \centering
    \caption{Statistics of the evaluation splits introduced in this work. Each entry reports the number of videos / frames.}
    \label{tab:dataset_splits}
    \setlength{\tabcolsep}{6pt}
    \resizebox{\columnwidth}{!}{%
    \begin{tabular}{lcccc}
        \toprule
        \textbf{Split} & \textbf{Train} & \textbf{Val} & \textbf{Test} & \textbf{Hidden Test} \\
        \midrule
        \emph{all-centers}    & 22 / 117796 & 8 / 36734 & 10 / 54466 & -- \\
        \emph{crossval-fold1} & 24 / 123693 & 6 / 29972 & 10 / 55331 & -- \\
        \emph{crossval-fold2} & 24 / 120739 & 6 / 28092 & 10 / 60165 & -- \\
        \emph{crossval-fold3} & 24 / 132770 & 6 / 32709 & 10 / 43517 & -- \\
        \emph{challenge}      & 14 / 73855 & 2 / 9723 & 9 / 50572 & 15 / 74846 \\
        \bottomrule
    \end{tabular}%
    }
\end{table}

\begin{figure}[h]
\centering
    \includegraphics[width=0.50\textwidth]{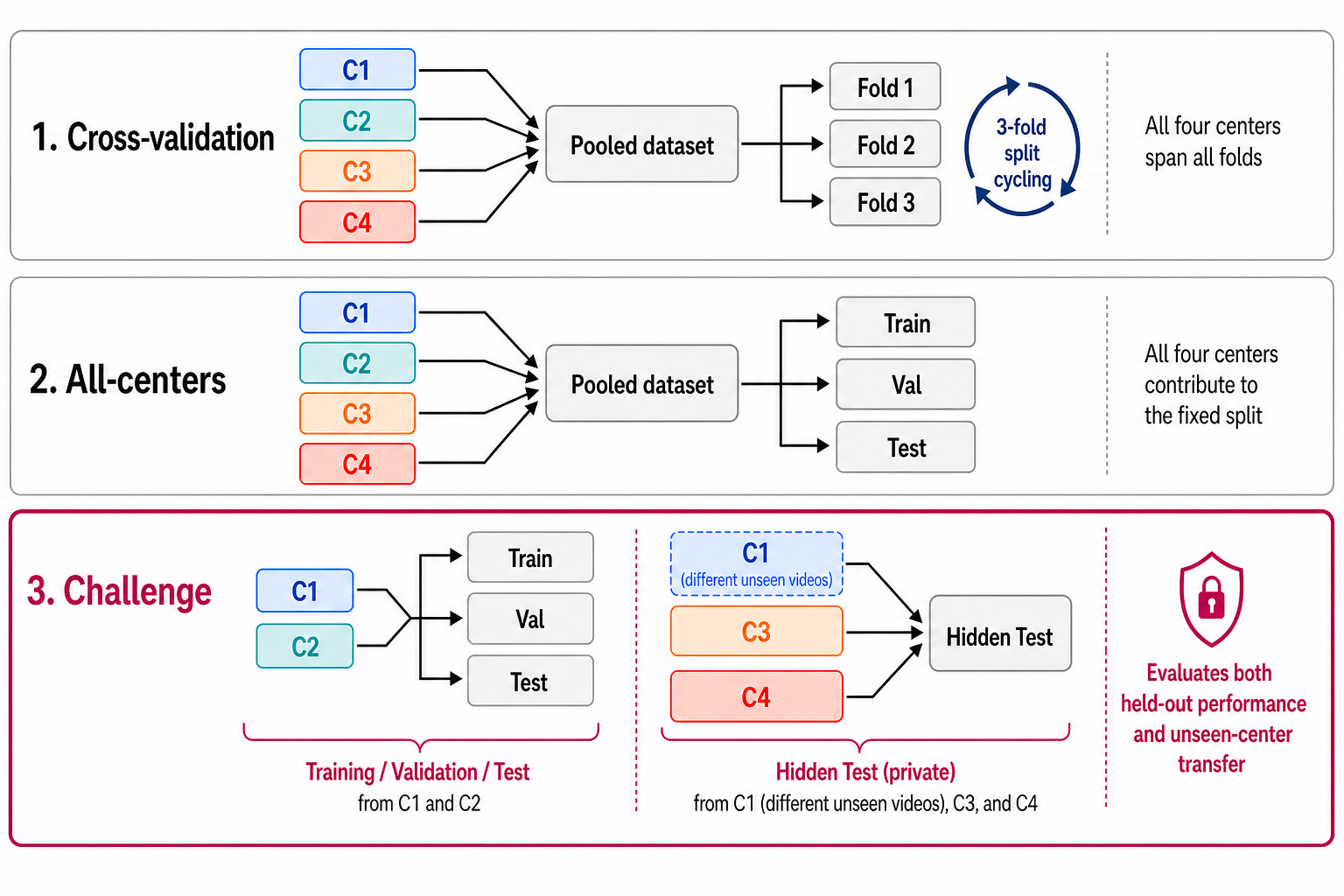} 
    \caption{\label{fig:mbt40_data_splits} Overview of the three evaluation settings used in MultiBypass-4C-T40 and their corresponding center-wise data allocation.}
    \label{fig:mbt40_data_splits}
\end{figure}



\subsection{Data Splits}
To support a comprehensive evaluation of surgical action triplet recognition, we define three benchmark settings following the general evaluation protocol established in CholecT50~\citep{nwoye2022rendezvous}. After downsampling each video to 1\,FPS, we construct an \emph{all-centers} split, in which videos from all four centers are used in a standard fixed training, validation, and test partition, and a \emph{cross-validation} setting, which likewise spans all four centers but uses three folds to assess robustness to different train-validation partitions. In addition, we define a dedicated \emph{challenge} split for the MultiBypassTriplets2026 MICCAI Challenge. In this protocol, videos from C1 and C2 are used for training, validation, and test, while a separate hidden test set is formed from unseen videos from C1 together with videos from C3 and C4. As shown in Fig.~\ref{fig:mbt40_data_splits}, this design allows us to distinguish performance on a familiar institutional distribution from transfer to genuinely unseen centers.

Together, these three settings provide complementary views of model behavior: standard fixed-split evaluation across all centers, robustness across alternative data partitions, and a stricter transfer setting that explicitly probes generalization beyond the centers used for training. Table~\ref{tab:dataset_splits} summarizes the allocation of videos and frames across these evaluation protocols.



%% file: sections/03b-methods.tex
\section{Methodology}
\begin{figure*}[!htbp]
\centering
    \includegraphics[width=1.00\textwidth]{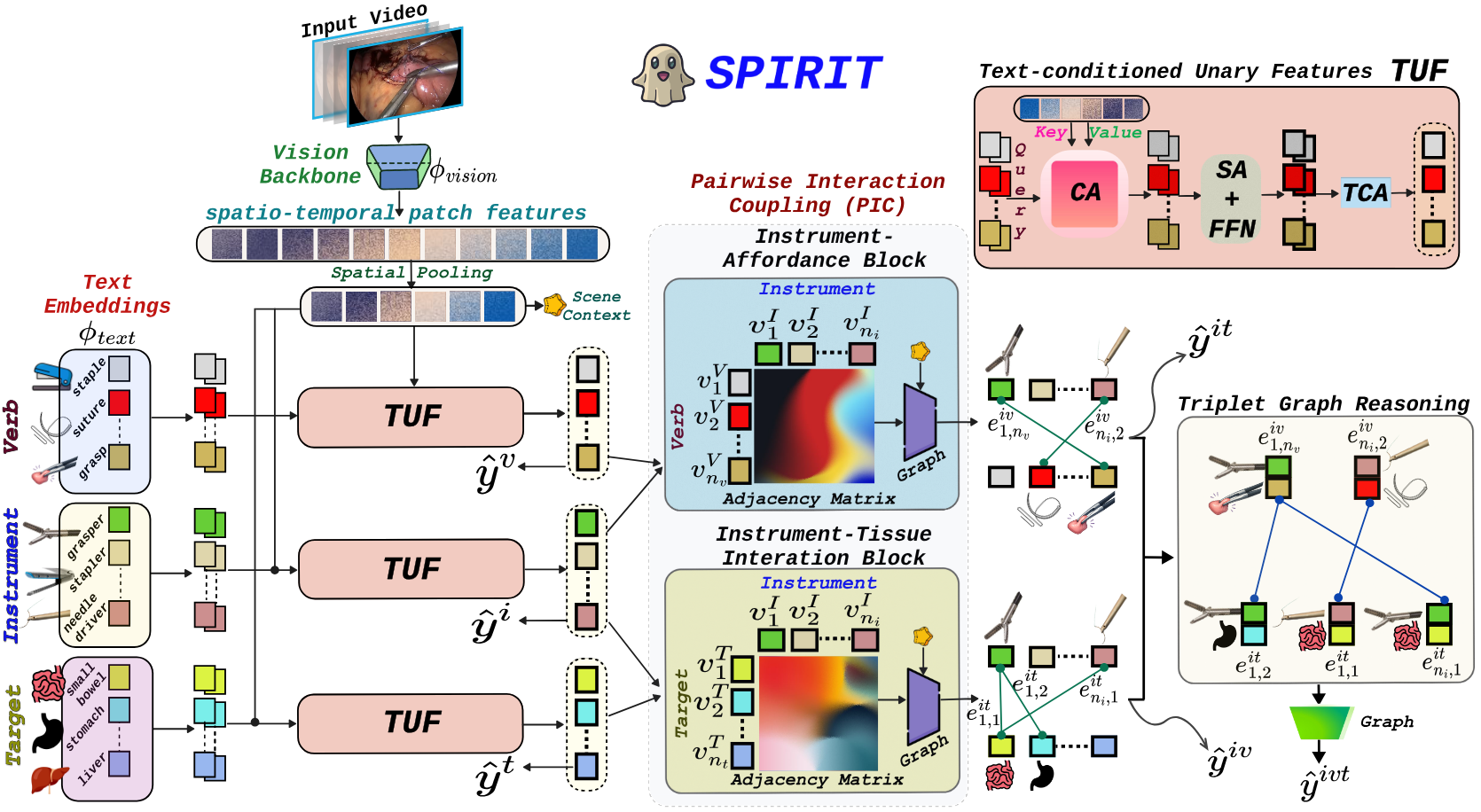} 
    \caption{Overview of SPIRIT, which consists of three modules: text-conditioned unary feature learning (TUF), pairwise interaction coupling (PIC), and triplet graph reasoning (TGR). Within the TUF module, class queries are updated through cross-attention to spatio-temporal visual tokens (CA), self-attention (SA), feed-forward networks (FFN), and temporal cross-attention to frame-level descriptors (TCA).}
    \label{fig:SPIRIT_model}
\end{figure*}

In this section, we introduce SPIRIT, short for \textbf{S}patio-temporal \textbf{P}airw\textbf{I}se \textbf{R}elational modeling of \textbf{I}nstrument-\textbf{T}issue interactions. SPIRIT, as shown in Figure~\ref{fig:SPIRIT_model}, is designed for surgical action triplet recognition, where the objective is to predict a triplet of the form \textlangle{}\emph{instrument}, \emph{verb}, \emph{target}\textrangle{} for each surgical interaction. Rather than treating triplets as flat labels, the proposed framework explicitly models their underlying structure: surgical actions arise through interactions between instruments and anatomical targets, while the verb depends on the affordance of the instrument. Accordingly, SPIRIT decomposes triplet recognition into three stages: \emph{unary representation learning}, \emph{pairwise relational modeling}, and \emph{final triplet composition}. 

Given an input video clip, the model first builds unary representations for \emph{instrument}, \emph{verb} and \emph{target} from visual tokens and text-conditioned class queries. It then constructs two dedicated pairwise branches: an \emph{instrument-target} branch for interaction semantics, and an \emph{instrument-verb} branch for affordance modeling. Finally, these pairwise representations are integrated through a triplet reasoning module that predicts coherent \emph{instrument-verb-target} combinations under compositional constraints. Next, we describe each stage in detail.

\subsection{Spatio-Temporal Video Backbone}

Let an input video clip be denoted by $\mathbf{X}=\{I_t\}_{t=1}^{T}$, where $T$ is the temporal extent of the clip. Each frame $I_t$ is processed by a visual encoder $\phi_{\mathrm{vision}}$ to extract spatial feature maps,
\begin{equation}
\mathbf{F}_t=\phi_{\mathrm{vision}}(I_t)\in\mathbb{R}^{H_b\times W_b\times D},
\end{equation}
where $H_b$ and $W_b$ denote the spatial resolution of the backbone features and $D$ is the feature dimension. Rather than retaining all backbone patch tokens, we apply spatial pooling to obtain a more compact feature grid while preserving the layout needed for downstream reasoning. This yields a pooled representation $\tilde{\mathbf{F}}_t\in\mathbb{R}^{H\times W\times D}$, where $H=W=S$ denotes the pooled spatial resolution. Stacking these representations over time gives $\tilde{\mathbf{F}}\in\mathbb{R}^{T\times H\times W\times D}$.

We then flatten the pooled grid within each frame into $N=HW$ patch tokens and project each token into a shared latent space of dimension $d$ using a learned visual projection map $\Pi_{\mathrm{vision}}$, producing
\begin{equation}
\mathbf{Z}^{0}\in\mathbb{R}^{T\times N\times d}, \qquad 
\mathbf{Z}^{0}_{t,n}=\Pi_{\mathrm{vision}}(\tilde{\mathbf{F}}_{t,n}),
\end{equation}
where $t\in\{1,\dots,T\}$ indexes time and $n\in\{1,\dots,N\}$ indexes pooled spatial locations. To encode spatial layout and temporal order, we add spatial and temporal positional embeddings,
\begin{equation}
\mathbf{Z}_{t,n}=\mathbf{Z}^{0}_{t,n}+\mathbf{P}_{s,n}+\mathbf{P}_{t,t}, \qquad
\mathbf{Z}\in\mathbb{R}^{T\times N\times d},
\end{equation}
where $\mathbf{P}_{s,n}$ and $\mathbf{P}_{t,t}$ denote the spatial and temporal positional embeddings, respectively. The resulting token set $\mathbf{Z}$ serves as the visual memory for downstream class-query refinement, retaining both the spatial evidence of where an interaction occurs and the temporal evidence of how it evolves over the clip.

\subsection{Text-conditioned Unary Features}
To explicitly model the three components of a surgical triplet, SPIRIT first learns unary representations for instrument, verb and target through separate component-specific branches. We refer to this stage as \emph{Text-conditioned Unary Features (TUF)}, since it constructs unary node representations by refining text-conditioned class queries with visual and temporal evidence. Each branch is initialized with semantic priors derived from its class vocabulary using a text encoder $\phi_{\mathrm{text}}$. Concretely, we denote the instrument, verb and target vocabularies by $\mathcal{C}^i=\{c^i_m\}_{m=1}^{n_i}$, $\mathcal{C}^v=\{c^v_m\}_{m=1}^{n_v}$ and $\mathcal{C}^t=\{c^t_m\}_{m=1}^{n_t}$, where $n_i$, $n_v$ and $n_t$ denote the number of instrument, verb and target classes, respectively.
For each component type $\kappa\in\{i,v,t\}$ and class label $c^\kappa_m$, we construct a generic text prompt of the form \texttt{`a photo of \{c\}'}, where $\{c\}$ is instantiated with the corresponding class name. The resulting prompt is encoded by the text encoder $\phi_{\mathrm{text}}$ to produce a semantic embedding
\begin{equation}
\mathbf{e}^\kappa_m=\phi_{\mathrm{text}}(p(c^\kappa_m))\in\mathbb{R}^{d_{\mathrm{text}}},
\end{equation}
where $p(\cdot)$ denotes the prompt template.
These embeddings are projected into the same latent space as the visual tokens through a shared projection map $\Pi_{\mathrm{text}}$, yielding the initial class queries
\begin{equation}
\mathbf{q}^{\kappa,0}_m=\Pi_{\mathrm{text}}(\mathbf{e}^\kappa_m),
\qquad
\mathbf{Q}^{\kappa,0}\in\mathbb{R}^{n_\kappa\times d},
\end{equation}
where $n_\kappa$ denotes the number of classes in branch $\kappa$. These text-derived queries provide semantic initialization before video evidence is incorporated.

Given the spatio-temporal visual memory $\mathbf{Z}\in\mathbb{R}^{T\times N\times d}$, we reshape it into $\mathbf{M}\in\mathbb{R}^{TN\times d}$, which serves as the key-value memory for query refinement. For each branch $\kappa\in\{i,v,t\}$, the initial queries are updated by a decoder-style attention block. Cross-attention first injects spatio-temporal visual evidence into the queries:
\begin{equation}
\begin{aligned}
\mathbf{A}^{\kappa}
&=
\mathrm{softmax}\!\left(
\frac{
(\mathbf{Q}^{\kappa,0}\mathbf{W}^{\kappa}_{Q})
(\mathbf{M}\mathbf{W}^{\kappa}_{K})^{\top}
}{
\sqrt{d_h}
}
\right), \\
\mathbf{U}^{\kappa}
&=
\mathbf{A}^{\kappa}(\mathbf{M}\mathbf{W}^{\kappa}_{V}), \qquad
\bar{\mathbf{Q}}^{\kappa}
=
\mathrm{LN}\!\left(\mathbf{Q}^{\kappa,0}+\mathbf{U}^{\kappa}\right).
\end{aligned}
\end{equation}
Here, $\mathbf{W}^{\kappa}_{Q}$, $\mathbf{W}^{\kappa}_{K}$ and $\mathbf{W}^{\kappa}_{V}$ are learned linear projections for the query, key and value spaces of branch $\kappa$, and $d_h$ denotes the attention head dimension. This operation allows each query to gather evidence from all spatial locations across all frames, aligning class semantics with the observed content of the clip.

The visually updated queries are then refined through self-attention within each branch,
\begin{equation}
\mathbf{S}^{\kappa}
=
\mathrm{MHA}\!\left(
\bar{\mathbf{Q}}^{\kappa},
\bar{\mathbf{Q}}^{\kappa},
\bar{\mathbf{Q}}^{\kappa}
\right),
\qquad
\tilde{\mathbf{Q}}^{\kappa}
=
\mathrm{LN}\!\left(\bar{\mathbf{Q}}^{\kappa}+\mathbf{S}^{\kappa}\right),
\end{equation}
followed by a feed-forward transformation
\begin{equation}
\mathbf{Q}^{\kappa,*}
=
\mathrm{LN}\!\left(
\tilde{\mathbf{Q}}^{\kappa}
+
\mathrm{FFN}\!\left(\tilde{\mathbf{Q}}^{\kappa}\right)
\right).
\end{equation}
Here, $\mathrm{MHA}(\cdot)$ denotes multi-head self-attention, $\mathrm{LN}(\cdot)$ denotes layer normalization, and $\mathrm{FFN}(\cdot)$ denotes a position-wise feed-forward network. This ordering first updates each class query with visual evidence from the clip, and then allows the queries within the same branch to interact, so that their representations are refined in a context-aware manner.

The preceding refinement stages capture visual evidence at the token level, but do not explicitly model how that evidence evolves across frames. To encode this temporal structure, we compute a frame-level descriptor for each time step by averaging the visual tokens over the spatial dimension, yielding $\mathbf{g}_t=\frac{1}{N}\sum_{n=1}^{N}\mathbf{Z}_{t,n}$ and the stacked temporal memory $\mathbf{G}\in\mathbb{R}^{T\times d}$. For each branch, a temporal cross-attention module (TCA) then uses the refined class queries $\mathbf{Q}^{\kappa,*}$ as queries and $\mathbf{G}$ as temporal memory:
\begin{equation}
\begin{aligned}
\mathbf{B}^{\kappa}
&=
\mathrm{softmax}\!\left(
\frac{
(\mathbf{Q}^{\kappa,*}\mathbf{W}^{\kappa}_{Q,\tau})
(\mathbf{G}\mathbf{W}^{\kappa}_{K,\tau})^{\top}
}{
\sqrt{d_h}
}
\right), \\
\mathbf{T}^{\kappa}
&=
\mathbf{B}^{\kappa}(\mathbf{G}\mathbf{W}^{\kappa}_{V,\tau}), \qquad
\mathbf{Q}^{\kappa}
=
\mathrm{LN}\!\left(\mathbf{Q}^{\kappa,*}+\mathbf{T}^{\kappa}\right).
\end{aligned}
\end{equation}
Here, $\mathbf{W}^{\kappa}_{Q,\tau}$, $\mathbf{W}^{\kappa}_{K,\tau}$ and $\mathbf{W}^{\kappa}_{V,\tau}$ denote the learned query, key and value projection matrices of the temporal attention module for branch $\kappa$, where the subscript $\tau$ indicates temporal refinement. This step enables each class query to aggregate evidence across the clip duration, which is especially important for motion-sensitive verb semantics while also improving temporal stability for instrument and target representations.

The resulting unary node features are $\mathbf{Q}^{i}\in\mathbb{R}^{n_i\times d}$, $\mathbf{Q}^{v}\in\mathbb{R}^{n_v\times d}$ and $\mathbf{Q}^{t}\in\mathbb{R}^{n_t\times d}$. In addition to serving as inputs to the downstream relational branches, these unary features are mapped directly to branch-wise logits through lightweight classification heads,
\begin{equation}
\hat{\mathbf{y}}^{\kappa}
=
f_{\kappa}\!\left(\mathbf{Q}^{\kappa}\right),
\qquad \kappa\in\{i,v,t\},
\end{equation}
where $f_{\kappa}(\cdot)$ denotes the unary classification head for branch $\kappa$. These predictions provide explicit supervision for the individual triplet components, while the corresponding unary features form the semantic foundation of SPIRIT by combining class-level prior knowledge with video-specific spatio-temporal evidence for downstream relational reasoning.

\subsection{Learning Interaction Semantics}

A surgical interaction is fundamentally defined by the coupling between an instrument and an anatomical target. To model this explicitly, SPIRIT constructs an \emph{instrument-target} relational branch over the unary node features $\mathbf{Q}^{i}\in\mathbb{R}^{n_i\times d}$ and $\mathbf{Q}^{t}\in\mathbb{R}^{n_t\times d}$, where $n_i$ and $n_t$ denote the numbers of instrument and target classes, respectively. In addition, we derive a clip-level scene descriptor from the visual memory and project it into a global context vector $\mathbf{s}\in\mathbb{R}^{d_s}$, which provides complementary scene-level information for pairwise reasoning.

For each candidate instrument-target pair $(m,n)$, with $m\in\{1,\dots,n_i\}$ and $n\in\{1,\dots,n_t\}$, let $\mathbf{q}^{i}_{m}\in\mathbb{R}^{d}$ and $\mathbf{q}^{t}_{n}\in\mathbb{R}^{d}$ denote the corresponding unary node features. To capture both local interaction structure and global surgical context, we form a pair descriptor by concatenating the two endpoint features together with the scene descriptor,
$
\mathbf{r}^{it}_{m,n}
=
\left[
\mathbf{q}^{i}_{m};
\mathbf{q}^{t}_{n};
\mathbf{s}
\right].
$
This descriptor is then encoded by a learnable edge map $\phi_{it}$ to produce an edge embedding
\begin{equation}
\mathbf{e}^{it}_{m,n}
=
\phi_{it}\!\left(\mathbf{r}^{it}_{m,n}\right)
\in\mathbb{R}^{d_e}.
\end{equation}
Collecting all such embeddings yields the dense interaction tensor $\mathbf{E}^{it}\in\mathbb{R}^{n_i\times n_t\times d_e}$. In this way, each pairwise relation is conditioned jointly on the instrument node, the target node and the global scene context.

We organize these relations as a dense bipartite graph, where $\mathcal{V}^{i}$ and $\mathcal{V}^{t}$ denote the instrument-node and target-node sets, respectively, and the full node set is given by $\mathcal{V}^{it}=\mathcal{V}^{i}\cup\mathcal{V}^{t}$. The corresponding edge set, $\mathcal{E}^{it}=\{(u,v)\mid u\in\mathcal{V}^{i},\, v\in\mathcal{V}^{t}\}$, contains all cross-partition instrument-target pairs. We perform message passing over the full instrument-target bipartite graph, allowing the model to learn interaction compatibility across all candidate instrument-target pairs rather than restricting reasoning to a predefined subset.
Let $\mathbf{X}^{it}=[\mathbf{Q}^{i};\mathbf{Q}^{t}]$ denote the stacked node features. Using the learned edge attributes $\mathbf{E}^{it}$, we propagate relational messages through graph attention layer GATv2 \citep{brody2021attentive} as
\begin{equation}
\mathbf{X}^{it}_{\mathrm{msg}}
=
\mathrm{GATv2}\!\left(\mathbf{X}^{it}, \mathcal{E}^{it}, \mathbf{E}^{it}\right).
\end{equation}
Because the graph output dimension differs from the input node dimension, we first project the node features into the graph space using a learned residual projection $\Pi_{\mathrm{res}}$, and then apply residual fusion followed by layer normalization:
\begin{equation}
\tilde{\mathbf{X}}^{it}
=
\mathrm{LN}\!\left(
\Pi_{\mathrm{res}}(\mathbf{X}^{it})+\mathbf{X}^{it}_{\mathrm{msg}}
\right).
\end{equation}
Here, $\Pi_{\mathrm{res}}$ denotes the residual projection map and $\mathrm{LN}(\cdot)$ denotes layer normalization. We then split the updated node tensor back into the instrument and target partitions, yielding $\tilde{\mathbf{Q}}^{i,\mathrm{gnn}}\in\mathbb{R}^{n_i\times d_o}$ and $\tilde{\mathbf{Q}}^{t,\mathrm{gnn}}\in\mathbb{R}^{n_t\times d_o}$, where $d_o$ is the output dimension of the relational branch.

Graph message passing updates the node features through the bipartite interaction structure, but the learned edge embeddings themselves also contain explicit pair-specific information about instrument-target compatibility. To inject this information back into the node states, we aggregate the edge embeddings to their endpoint nodes. For each instrument node $m$, we summarize all incident instrument-target edges by averaging transformed edge features over the target partition. Similarly, for each target node $n$, we average transformed edge features over the instrument partition:
\begin{equation}
\mathbf{a}^{i}_{m}
=
\frac{1}{n_t}\sum_{n=1}^{n_t}\psi^{i}_{it}\!\left(\mathbf{e}^{it}_{m,n}\right),
\qquad
\mathbf{a}^{t}_{n}
=
\frac{1}{n_i}\sum_{m=1}^{n_i}\psi^{t}_{it}\!\left(\mathbf{e}^{it}_{m,n}\right),
\end{equation}
where $\mathbf{a}^{i}_{m}$ and $\mathbf{a}^{t}_{n}$ denote the aggregated relational summaries for the $m$-th instrument node and the $n$-th target node, respectively, and $\psi^{i}_{it}$ and $\psi^{t}_{it}$ are learned side-specific edge projection maps. These summaries are then added to the graph-updated node features,
\begin{equation}
\hat{\mathbf{Q}}^{i}
=
\tilde{\mathbf{Q}}^{i,\mathrm{gnn}}+\mathbf{a}^{i},
\qquad
\hat{\mathbf{Q}}^{t}
=
\tilde{\mathbf{Q}}^{t,\mathrm{gnn}}+\mathbf{a}^{t},
\end{equation}
and further refined through a feed-forward update with residual connection and layer normalization:
\begin{equation}
\tilde{\mathbf{Q}}^{i}_{it}
=
\hat{\mathbf{Q}}^{i}
+
\mathrm{FFN}\!\left(\mathrm{LN}(\hat{\mathbf{Q}}^{i})\right),
\qquad
\tilde{\mathbf{Q}}^{t}_{it}
=
\hat{\mathbf{Q}}^{t}
+
\mathrm{FFN}\!\left(\mathrm{LN}(\hat{\mathbf{Q}}^{t})\right).
\end{equation}
This yields interaction-aware node representations that combine unary semantic priors, graph-level relational context and explicit pairwise evidence from the learned instrument-target edge embeddings.

In addition to updating the node features, the learned edge embeddings are used to predict pairwise interaction scores through an edge classifier,
\begin{equation}
\hat{y}^{it}_{m,n}
=
g_{it}\!\left(\mathbf{e}^{it}_{m,n}\right),
\end{equation}
where $g_{it}(\cdot)$ denotes the instrument-target edge classifier. Collecting these scores over all candidate pairs yields the interaction matrix $\hat{\mathbf{Y}}^{it}\in\mathbb{R}^{n_i\times n_t}$, which can be interpreted as an interaction heatmap over the instrument-target space. This heatmap quantifies how likely each instrument class is to interact with each target class in the observed clip.

Overall, the instrument-target branch serves two complementary roles. First, it produces explicit pairwise interaction scores over all candidate instrument-target combinations. Second, it uses these learned relations to refine the underlying node representations through edge-conditioned graph reasoning. The resulting features provide a structured notion of which anatomical target a given instrument is most likely to act on, which is essential in surgical scenes where the same instrument may engage different tissues depending on the procedural context.

\subsection{Learning Instrument Affordance}

A valid surgical triplet is determined not only by \emph{where} an instrument acts, but also by \emph{what} action that instrument can plausibly perform. This second aspect corresponds to \emph{instrument affordance}, which is especially important because surgical verbs are not interchangeable across tools: the same anatomical target may support multiple actions, but only a subset of these are compatible with the physical role and manipulation capability of a given instrument. For example, a grasper may support actions such as grasping or retracting, whereas cutting is typically associated with instruments designed for dissection. We therefore introduce a dedicated \emph{instrument-verb} relational branch to model \emph{affordance} explicitly. To our knowledge, this type of affordance-aware reasoning has not been explored in detail for surgical action triplet recognition, where most existing methods focus on direct triplet prediction without disentangling the compatibility structure between tools and actions.

This branch is formulated analogously to the instrument-target interaction branch, but replaces target nodes with verb nodes so as to capture the action space supported by each instrument under the observed scene context. It operates on the unary node features $\mathbf{Q}^{i}\in\mathbb{R}^{n_i\times d}$ and $\mathbf{Q}^{v}\in\mathbb{R}^{n_v\times d}$ together with the clip-level scene descriptor $\mathbf{s}\in\mathbb{R}^{d_s}$. For each candidate instrument-verb pair $(m,n)$, with $m\in\{1,\dots,n_i\}$ and $n\in\{1,\dots,n_v\}$, we form the pair descriptor
\begin{equation}
\mathbf{r}^{iv}_{m,n}
=
\left[
\mathbf{q}^{i}_{m};
\mathbf{q}^{v}_{n};
\mathbf{s}
\right],
\end{equation}
and encode it through a learnable edge encoder $\phi_{iv}$ to obtain
\begin{equation}
\mathbf{e}^{iv}_{m,n}
=
\phi_{iv}\!\left(\mathbf{r}^{iv}_{m,n}\right)
\in\mathbb{R}^{d_e}.
\end{equation}
Collecting all pairwise relations yields the affordance tensor $\mathbf{E}^{iv}\in\mathbb{R}^{n_i\times n_v\times d_e}$. We then construct the corresponding dense bipartite instrument-verb graph and apply the same edge-conditioned graph reasoning as in the interaction branch, yielding the affordance-aware node features
\begin{equation}
\tilde{\mathbf{Q}}^{i}_{iv}\in\mathbb{R}^{n_i\times d_o},
\qquad
\tilde{\mathbf{Q}}^{v}_{iv}\in\mathbb{R}^{n_v\times d_o}.
\end{equation}
In parallel, we classify each edge to obtain instrument-verb compatibility scores,
\begin{equation}
\hat{y}^{iv}_{m,n}
=
g_{iv}\!\left(\mathbf{e}^{iv}_{m,n}\right),
\end{equation}
which together form the affordance matrix $\hat{\mathbf{Y}}^{iv}\in\mathbb{R}^{n_i\times n_v}$. This matrix quantifies which actions are supported by each instrument in the observed clip. In this sense, the instrument-verb branch complements the instrument-target branch: the latter models \emph{where} the action is applied, whereas the former constrains \emph{what} action the instrument can plausibly perform.

\subsection{Learning Triplet Composition via Graph Reasoning}

The two pairwise branches capture complementary aspects of a surgical action: the instrument-target branch models \emph{where} the action is applied, whereas the instrument-verb branch models \emph{what} action the instrument can plausibly perform. Triplet recognition, however, requires these two relation spaces to be composed into a unified prediction over valid instrument-verb-target combinations. To this end, we introduce a graph-based triplet composition module that reasons jointly over the learned interaction and affordance relations. The central idea is to elevate pairwise relations themselves to graph entities and to compose triplets only through structurally compatible instrument-target and instrument-verb pairs.

Starting from the learned pairwise edge tensors $\mathbf{E}^{it}\in\mathbb{R}^{n_i\times n_t\times d_e}$ and $\mathbf{E}^{iv}\in\mathbb{R}^{n_i\times n_v\times d_e}$, we reinterpret each instrument-target edge and each instrument-verb edge as a relation node. These edge embeddings are first projected into a shared hidden space,
\begin{equation}
\mathbf{H}^{it}_{m,n}=\psi_{it}\!\left(\mathbf{e}^{it}_{m,n}\right),
\qquad
\mathbf{H}^{iv}_{m,n}=\psi_{iv}\!\left(\mathbf{e}^{iv}_{m,n}\right),
\end{equation}
where $\psi_{it}$ and $\psi_{iv}$ are learned relation-node projection maps. Flattening the pairwise index spaces yields $\mathbf{H}^{it}\in\mathbb{R}^{N_{it}\times d_h}$ and $\mathbf{H}^{iv}\in\mathbb{R}^{N_{iv}\times d_h}$, where $N_{it}=n_in_t$ and $N_{iv}=n_in_v$. The triplet module therefore operates not on unary nodes, but on a \emph{higher-order graph} whose nodes correspond to pairwise relations.

We then construct a bipartite relation graph between the instrument-target and instrument-verb relation nodes. A connection is introduced only when the two relation nodes participate in a valid triplet defined by the dataset ontology. Let $\mathcal{M}_{ivt}$ denote the predefined mapping from each triplet class to its associated instrument, verb and target indices. Using this mapping, we define the relation-graph edge set as
\begin{equation}
\mathcal{E}_{\mathrm{rel}}
=
\left\{
\big((i,t),(i,v)\big)
\;|\;
(i,v,t)\in\mathcal{M}_{ivt}
\right\},
\end{equation}
together with the reverse edges to enable bidirectional message passing. This construction is important because it restricts reasoning to structurally valid triplet compositions rather than allowing arbitrary mixing between unrelated pairwise relations.

Let
$
\mathbf{H}^{0}
=
\left[
\mathbf{H}^{it};\mathbf{H}^{iv}
\right]
$
denote the stacked relation-node features. We perform graph reasoning over the relation graph using graph attention, followed by residual fusion and normalization,
\begin{equation}
\begin{aligned}
\mathbf{H}^{1}
&=
\mathrm{GATv2}\!\left(\mathbf{H}^{0},\mathcal{E}_{\mathrm{rel}}\right), \\
\tilde{\mathbf{H}}
&=
\mathrm{LN}\!\left(\mathbf{H}^{0}+\mathbf{H}^{1}\right).
\end{aligned}
\end{equation}
In this way, each relation node is updated not only by its own pairwise evidence but also by messages from the complementary relation space. An instrument-target relation can therefore be reinforced or suppressed by the corresponding instrument-verb evidence, and vice versa.

For each triplet class $k$, we then retrieve the updated instrument-target and instrument-verb relation nodes specified by $\mathcal{M}_{ivt}$. Let $\tilde{\mathbf{h}}^{it}_k$ and $\tilde{\mathbf{h}}^{iv}_k$ denote the corresponding relation embeddings. We form a triplet composition feature by combining them through concatenation, element-wise interaction and absolute difference,
\begin{equation}
\mathbf{z}^{ivt}_k
=
\left[
\tilde{\mathbf{h}}^{it}_k;
\tilde{\mathbf{h}}^{iv}_k;
\tilde{\mathbf{h}}^{it}_k\odot\tilde{\mathbf{h}}^{iv}_k;
\left|\tilde{\mathbf{h}}^{it}_k-\tilde{\mathbf{h}}^{iv}_k\right|
\right].
\end{equation}
This design preserves the identity of the two relation types while also capturing their agreement and discrepancy. A lightweight triplet classifier then predicts the logit for triplet class $k$,
\begin{equation}
\hat{y}^{ivt}_k
=
g_{ivt}\!\left(\mathbf{z}^{ivt}_k\right),
\end{equation}
where $g_{ivt}(\cdot)$ denotes the triplet classification head. Collecting all logits yields $\hat{\mathbf{y}}^{ivt}\in\mathbb{R}^{n_{ivt}}$, where $n_{ivt}$ is the number of valid triplet classes.

Finally, prediction remains grounded in the dataset-defined triplet ontology: only valid instrument-verb-target combinations are used for supervision, while invalid compositions are excluded through a validity mask. Triplet recognition is therefore not treated as unconstrained flat classification, but as the composition of two semantically meaningful pairwise relation spaces. This graph formulation provides two main benefits. First, it enforces compositional consistency between pairwise relations and final triplet predictions by explicitly linking interaction and affordance evidence. Second, it restricts reasoning to structurally valid triplets, thereby reducing ambiguity and improving data efficiency. In this sense, the triplet graph forms the final compositional layer that integrates \emph{where} the action is applied and \emph{what} action the instrument can perform into a coherent instrument-verb-target prediction.

\subsection{Adaptive Teacher-Student Distillation}
To further improve generalization, we adopt a teacher-student distillation framework with adaptive sample weighting. Our formulation is inspired by KDAS-style distillation \citep{chae2025distill}, but is adapted here to the multi-label, multi-head setting of surgical triplet recognition. Training proceeds in two stages. First, a teacher model is trained using standard supervised learning and stored as a fixed checkpoint. Second, a student model is optimized under both ground-truth supervision and soft supervision from the frozen teacher.

Since surgical action triplet recognition is formulated as a multi-label prediction problem, we use binary cross-entropy (BCE) loss with logits at each prediction head. For a head with logits $\hat{\mathbf{y}}_h$ and ground-truth labels $\mathbf{y}_h$, the supervised loss is written compactly as
\begin{equation}
L_h = \mathrm{BCE}\!\left(\hat{\mathbf{y}}_h,\mathbf{y}_h\right),
\end{equation}
where $h\in\{i,v,t,it,iv,ivt\}$ indexes the unary, pairwise and triplet heads.

Distillation is applied to all prediction heads. Let $\mathbf{s}_n$ and $\mathbf{t}_n$ denote the student and teacher logits for sample $n$ at a given head, and let $\mathbf{y}_n$ denote the corresponding ground-truth label vector. We first define a teacher-student disagreement score
\begin{equation}
TS_n = \mathrm{BCE}\!\left(\mathbf{s}_n,\sigma(\mathbf{t}_n)\right),
\end{equation}
which measures how strongly the student deviates from the teacher’s soft predictions. We also define a teacher-ground-truth reliability score
\begin{equation}
TG_n = \mathrm{BCE}\!\left(\mathbf{t}_n,\mathbf{y}_n\right),
\end{equation}
which measures how well the teacher agrees with the ground-truth labels. Intuitively, samples with large teacher-student disagreement are more informative for knowledge transfer, whereas samples with large teacher-ground-truth disagreement indicate a less reliable teacher signal. We therefore assign each sample a distillation weight
\begin{equation}
w_n \propto TS_n \exp(-TG_n/\beta),
\end{equation}
where $\beta$ controls how strongly teacher-ground-truth disagreement suppresses the contribution of a sample to the distillation loss. The weights are normalized within each batch, and the final distillation loss is computed as a weighted average of the per-sample teacher-student disagreement terms.
Distillation is applied jointly across the unary, pairwise and triplet heads, encouraging the student to inherit not only the teacher’s final predictions but also its intermediate relational structure.

\subsection{Two-stage Training and Loss Functions}

SPIRIT is trained in two stages. In Stage~1, we optimize the base model using supervised losses on the unary, pairwise and triplet outputs. In Stage~2, the student model is trained with the same supervised objective together with the adaptive distillation loss introduced in the previous subsection, using a frozen teacher checkpoint.

Let $\mathcal{L}_i$, $\mathcal{L}_v$, $\mathcal{L}_t$, $\mathcal{L}_{it}$, $\mathcal{L}_{iv}$ and $\mathcal{L}_{ivt}$ denote the supervised losses for the instrument, verb, target, instrument-target, instrument-verb and triplet heads, respectively. The overall supervised objective is
\begin{equation}
\mathcal{L}_{sup}
=
\lambda_i \mathcal{L}_i +
\lambda_v \mathcal{L}_v +
\lambda_t \mathcal{L}_t +
\lambda_{it} \mathcal{L}_{it} +
\lambda_{iv} \mathcal{L}_{iv} +
\lambda_{ivt} \mathcal{L}_{ivt},
\end{equation}
where $\lambda_i,\lambda_v,\lambda_t,\lambda_{it},\lambda_{iv}$ and $\lambda_{ivt}$ are head-specific weights for the supervised losses. In practice, each term is implemented as a binary cross-entropy loss with logits, together with class-dependent weighting to mitigate label imbalance.

During Stage~2, distillation is applied to all prediction heads. Let $\mathcal{L}^{kd}_i$, $\mathcal{L}^{kd}_v$, $\mathcal{L}^{kd}_t$, $\mathcal{L}^{kd}_{it}$, $\mathcal{L}^{kd}_{iv}$ and $\mathcal{L}^{kd}_{ivt}$ denote the corresponding distillation losses. We combine them as
\begin{equation}
\mathcal{L}_{kd}
=
\omega_i \mathcal{L}^{kd}_i +
\omega_v \mathcal{L}^{kd}_v +
\omega_t \mathcal{L}^{kd}_t +
\omega_{it} \mathcal{L}^{kd}_{it} +
\omega_{iv} \mathcal{L}^{kd}_{iv} +
\omega_{ivt} \mathcal{L}^{kd}_{ivt},
\end{equation}
where $\omega_i,\omega_v,\omega_t,\omega_{it},\omega_{iv}$ and $\omega_{ivt}$ are head-specific distillation weights. The total Stage~2 objective is then
\begin{equation}
\mathcal{L}
=
\mathcal{L}_{sup} + \mathcal{L}_{kd}.
\end{equation}

This two-stage strategy first establishes strong unary and relational supervision, and then further improves the student model by transferring structured knowledge from the teacher across the multiple prediction heads.

%% file: sections/04-experiments.tex
\section{Experiments}

\subsection{Task Setup and Evaluation Metrics}
Following CholecT50~\citep{nwoye2022rendezvous}, we formulate surgical action triplet recognition as a multi-label frame-level prediction task, where the goal is to infer the active surgical action triplet for a given frame. MultiBypass-4C-T40 contains 85 triplet classes, and we adopt a causal setting in which the prediction at each frame depends only on the current and preceding frames, without access to future observations.
For evaluation, we follow the CholecT50 protocol and report mean average precision (mAP) as the primary metric. 
Average precision is first computed for each class within each video and then averaged across classes and videos to obtain the final mAP. 
For completeness, and to better characterize performance across different levels of prediction, we additionally report AP for the unary components ($AP_I$, $AP_V$, $AP_T$), the pairwise relations ($AP_{IV}$, $AP_{IT}$), and the final triplet output ($AP_{IVT}$), as presented in the result tables. To complement mAP, we also report Hit@k for $k \in \{5,10,20\}$, which measures whether any ground-truth positive triplet is present among the top-$k$ predicted triplets for a given frame.

\subsection{Data Preprocessing}
We conduct experiments on the MultiBypass-4C-T40 dataset across all predefined evaluation splits. To ensure a consistent visual input space, all frames are resized to a fixed spatial resolution of $224\times224$. 
During training, RandAugment is applied, whereas no data augmentation is used for validation and test samples.

Rather than treating frames independently, we construct fixed-length video clips to provide short-term temporal context. For an annotated frame at time step $t$, we form a causal clip by collecting the preceding frames together with the current frame, so that the prediction remains aligned with frame $t$. When the temporal window extends beyond the beginning of a video, the earliest available frame is repeated to preserve a constant clip length. This yields a temporally ordered clip representation while maintaining a fixed input size for the model.

The original annotations are defined at the triplet level. To obtain supervision for the individual components and their pairwise relations, we follow the same label factorization strategy used in CholecT50~\citep{nwoye2022rendezvous}. Specifically, each triplet annotation is projected onto its associated instrument, verb, and target components, from which we derive binary labels for the unary tasks,
$$
\mathbf{y}^{i}\in\{0,1\}^{C_i},\qquad
\mathbf{y}^{v}\in\{0,1\}^{C_v},\qquad
\mathbf{y}^{t}\in\{0,1\}^{C_t},
$$
as well as for the pairwise and triplet tasks,
$$
\mathbf{y}^{it}\in\{0,1\}^{C_{it}},\qquad
\mathbf{y}^{iv}\in\{0,1\}^{C_{iv}},\qquad
\mathbf{y}^{ivt}\in\{0,1\}^{C_{ivt}}.
$$
In addition, we derive valid-pair masks from the dataset ontology to identify admissible instrument-target and instrument-verb combinations. These preprocessing steps yield a unified multi-level supervision scheme in which each annotated frame contributes structured labels at the unary, pairwise, and triplet levels, while the clip construction provides the temporal context required for spatio-temporal reasoning.

\subsection{Implementation Details}
We implement SPIRIT using the \emph{dino.txt} large vision-language backbone built on DINOv3 ViT-L/16~\citep{simeoni2025dinov3}, where the visual and text towers serve as $\phi_{\mathrm{vision}}$ and $\phi_{\mathrm{text}}$, respectively. The visual backbone outputs frame-wise features of dimension $d_0=1024$, while the text encoder produces features of dimension 2048. During training, all input frames are resized to $224\times224$, and causal clips of temporal extent $T=8$ are constructed. The visual backbone is frozen up to block 15.

Visual features are spatially pooled to an $8\times8$ grid ($S = 8$) and projected to a shared latent space of dimension $d=128$, while text features are projected into the same space through a shared text projection layer. We use sinusoidal spatial and temporal positional embeddings. Unary refinement is performed separately for the instrument, verb and target branches using one decoder-style block per branch with 4 attention heads, self-attention applied after cross-attention, feed-forward expansion ratio 2.0, and dropout 0.05. Temporal refinement is also applied separately to each unary branch using 4 attention heads. For relational reasoning, both the instrument-target and instrument-verb branches use one graph-attention layer GATv2 \citep{brody2021attentive} with 4 heads. Edge embeddings have dimension $d_e=128$, the relational branch outputs node features of dimension $d_o=256$, and scene context is enabled through concatenation with scene feature dimension $d_s=256$. The triplet relation graph uses hidden dimension 128 and one GATv2 layer with 4 heads. The model predicts 12 instrument classes, 13 verb classes, 15 target classes and 85 triplet classes.

We train for 30 epochs with batch size 16 using AdamW, a base learning rate of $10^{-4}$, a backbone learning rate of $2\times10^{-5}$, weight decay 0.06, gradient clipping 1.2, and a cosine learning-rate schedule. 
Training uses binary cross-entropy losses on all unary, pairwise and triplet heads. 
Model selection is based on validation mAP of the triplet head. All models are trained on NVIDIA A100 GPUs, and hyperparameters are selected based on validation performance.

In Stage~2, we initialize a frozen teacher from the best Stage~1 SPIRIT checkpoint and distill all six heads. Distillation uses BCE loss with temperature 1.0 and global weight 1.1. Teacher-reliability weighting is enabled, with reliability scaling parameter $\beta=1.0$ and mid-confidence filtering using quantile thresholds 0.1 and 0.9. The distillation head weights are set to 0.2 for the unary heads, 0.4 for the pairwise heads and 1.6 for the triplet head, placing the strongest emphasis on transferring structured triplet-level knowledge.

%% file: sections/05-results.tex
\begin{table*}[!htbp]
\centering
\setlength{\tabcolsep}{18pt}
\caption{\label{results_crossval}Results on the surgical action triplet recognition task using the official \textit{cross-validation} split of MultiBypass-4C-T40. Values are reported as mean $\pm$ standard deviation AP (\%). Parentheses indicate resolution; default is 224 unless specified.}
\resizebox{\textwidth}{!}{
\begin{tabular}{lcccccc}
\toprule
\multirow{2}{*}{\textbf{Method}} &
\multicolumn{3}{c}{\textbf{Component Detection}} &
\multicolumn{3}{c}{\textbf{Triplet Association}} \\
\cmidrule(lr){2-4} \cmidrule(lr){5-7}
& $AP_I$ & $AP_V$ & $AP_T$ & $AP_{IV}$ & $AP_{IT}$ & $AP_{IVT}$ \\
\midrule
\multicolumn{4}{l}{\textit{Triplet Specific Models}} \\
RDV & 39.9 $\pm$ 17.9 & 31.1 $\pm$ 11.5 & 37.0 $\pm$ 13.5 & 20.2 $\pm$ 9.4 & 26.7 $\pm$ 13.8 & 21.4 $\pm$ 10.8 \\
RiT & 39.5 $\pm$ 16.1 & 30.7 $\pm$ 12.2 & 36.0 $\pm$ 14.4 & 19.4 $\pm$ 8.9 & 26.1 $\pm$ 13.5 & 20.5 $\pm$ 10.7 \\

SelfD (224) & 66.3 $\pm$ 2.6 & 50.5 $\pm$ 1.9 & 55.1 $\pm$ 2.7 & 34.2 $\pm$ 1.1 & 44.8 $\pm$ 1.1 & 36.2 $\pm$ 2.1 \\

SelfD (384) & 72.5 $\pm$ 4.8 & 56.5 $\pm$ 1.0 & 60.5 $\pm$ 3.8 & 39.6 $\pm$ 1.4 & 50.7 $\pm$ 3.4 & 41.5 $\pm$ 1.8 \\

SelfD (Ens) & 75.6 $\pm$ 5.7 & 60.0 $\pm$ 2.7 & 62.9 $\pm$ 4.3 & 41.6 $\pm$ 2.3 & 53.3 $\pm$ 3.5 & 44.1 $\pm$ 2.1 \\

TERL (224) & 61.7 $\pm$ 5.3 & 47.1 $\pm$ 2.1 & 54.2 $\pm$ 2.6 & 30.7 $\pm$ 1.6 & 40.9 $\pm$ 2.3 & 32.6 $\pm$ 1.1 \\

TERL (384) & 66.7 $\pm$ 6.6 & 51.4 $\pm$ 4.1 & 57.0 $\pm$ 2.6 & 34.4 $\pm$ 2.7 & 45.2 $\pm$ 3.3 & 36.3 $\pm$ 2.1 \\

TERL (Ens) & 69.5 $\pm$ 6.0 & 54.0 $\pm$ 3.5 & 59.0 $\pm$ 2.4 & 36.0 $\pm$ 2.4 & 46.8 $\pm$ 2.9 & 38.0 $\pm$ 1.9 \\

CurConMix (224) & 69.7 $\pm$ 2.9 & 53.6 $\pm$ 0.7 & 60.1 $\pm$ 5.8 & 37.9 $\pm$ 2.1 & 48.7 $\pm$ 2.5 & 39.8 $\pm$ 1.3 \\

CurConMix (384) & 74.8 $\pm$ 4.7 & 57.8 $\pm$ 2.1 & 61.4 $\pm$ 4.0 & 40.5 $\pm$ 2.3 & 51.3 $\pm$ 3.2 & 42.4 $\pm$ 2.3 \\

CurConMix (Ens) & 75.4 $\pm$ 4.3 & 58.4 $\pm$ 2.1 & 62.8 $\pm$ 4.9 & 41.8 $\pm$ 2.7 & 52.6 $\pm$ 3.1 & 43.6 $\pm$ 2.1 \\

\midrule
\multicolumn{4}{l}{\textit{Foundation Models}} \\
SurgeNetXL & 64.4 $\pm$ 4.9 & 49.8 $\pm$ 1.7 & 54.9 $\pm$ 3.7 & 34.0 $\pm$ 1.9 & 44.8 $\pm$ 4.4 & 36.6 $\pm$ 2.5 \\
LemonFM & 71.8 $\pm$ 3.6 & 55.7 $\pm$ 2.0 & 58.7 $\pm$ 3.0 & 39.4 $\pm$ 2.6 & 50.4 $\pm$ 3.0 & 41.5 $\pm$ 1.8 \\
LemonFM-T & 68.4 $\pm$ 3.4 & 52.2 $\pm$ 1.9 & 58.1 $\pm$ 1.8 & 37.4 $\pm$ 1.5 & 48.4 $\pm$ 2.7 & 40.6 $\pm$ 1.0 \\
DinoV3-L & 73.4 $\pm$ 5.2 & 57.3 $\pm$ 2.5 & 58.8 $\pm$ 3.3 & 40.7 $\pm$ 2.7 & 51.2 $\pm$ 2.8 & 43.0 $\pm$ 1.7 \\
DinoV3-L-T & 73.8 $\pm$ 3.0 & 57.8 $\pm$ 0.8 & 59.5 $\pm$ 3.2 & 40.8 $\pm$ 2.2 & 51.3 $\pm$ 1.7 & 44.0 $\pm$ 2.1 \\

\midrule
\multicolumn{4}{l}{\textit{Our Models}} \\
SPIRIT-S & 79.6 $\pm$ 3.6 & 63.9 $\pm$ 2.0 & 64.9 $\pm$ 6.7 & 47.8 $\pm$ 1.3 & 57.2 $\pm$ 3.5 & 48.6 $\pm$ 2.0 \\
\rowcolor{magenta!15}
SPIRIT-TS & 82.1 $\pm$ 3.8 & 67.6 $\pm$ 3.4 & 66.9 $\pm$ 5.2 & 49.9 $\pm$ 1.7 & 60.1 $\pm$ 3.0 & 51.5 $\pm$ 2.2 \\
\bottomrule
\end{tabular}
}
\end{table*}

\section{Results}
In this section, we evaluate SPIRIT under the three benchmark settings introduced in this work, each targeting a complementary aspect of performance. The \emph{crossval} setting assesses robustness to different train-validation partitions, the \emph{all-centers} setting measures recognition under the standard fixed evaluation protocol, and the \emph{challenge} setting examines generalization in the more demanding public and hidden test setup, including transfer to unseen centers. We first describe the baseline methods used for comparison, and then present quantitative results and analysis for each setting in turn.

\subsection{Baselines}
We compare SPIRIT against a set of baselines chosen to represent both established triplet-specific architectures and stronger modern visual backbones. As task-specific baselines, we include Rendezvous (RDV)~\citep{nwoye2022rendezvous} and Rendezvous in Time (RiT)~\citep{sharma2023rendezvous}, which are among the standard methods for surgical action triplet recognition. Both are built on a ResNet-18 feature extractor and employ multi-task prediction heads for the triplet components, namely instrument (I), verb (V), and target (T), together with a final triplet classification head. 

To reflect more recent progress in dedicated triplet-recognition models, we further include SelfD~\citep{yamlahi2023self}, TERL~\citep{gui2024tail}, and CurConMix~\citep{jeon2025curconmix}, each evaluated in two single-model settings at $224\times224$ and $384\times384$ resolution, as well as in an ensemble configuration. These methods provide stronger task-specific baselines than RDV and RiT and are particularly relevant because they also aim to improve compositional triplet understanding beyond simple multi-head prediction. Including both their single-resolution and ensemble variants allows us to compare SPIRIT not only against earlier triplet-specific methods, but also against more competitive recent models under stronger evaluation settings.

To assess whether stronger generic visual representations alone are sufficient for the task, we further include SurgeNetXL~\citep{jaspers2025scaling}, a recent surgical foundation model. We also compare against LemonFM~\citep{che2026lemon}, a more recent surgical foundation model pretrained on a substantially larger and more diverse set of surgical videos than SurgeNetXL. In addition to the frame-based version, we report a temporal variant, denoted LemonFM-T. Finally, because SPIRIT is built on DinoV3-L~\citep{simeoni2025dinov3}, we include two corresponding backbone baselines: a frame-based DinoV3-L model and a temporal variant, denoted DinoV3-L-T, which uses the same number of input frames as SPIRIT. Within our own method family, \emph{SPIRIT-S} denotes the student model, whereas \emph{SPIRIT-TS} denotes the corresponding variant trained with teacher-student distillation. This setup allows us to distinguish improvements arising from stronger visual representations alone from those due to the explicit relational modeling and distillation strategy introduced by SPIRIT.

\subsection{Cross-validation Results}
Table~\ref{results_crossval} reports results on the official \emph{cross-validation} setting, which corresponds to the three folds listed as \emph{crossval-fold1}, \emph{crossval-fold2}, and \emph{crossval-fold3} in Table~\ref{tab:dataset_splits}. This protocol is designed to assess robustness to different train-validation partitions rather than performance on a single fixed split.
The results are best understood by considering the three groups of methods shown in the table: dedicated triplet-specific models, foundation-model baselines, and our SPIRIT variants. 
For fair comparison, all foundation-model baselines are fine-tuned on the target training split, rather than used as frozen feature extractors. Additional training details and selected hyperparameters for the non-SPIRIT baselines are summarized in Table~\ref{tab:training_hparams_all} in the appendix.

Among the \textbf{\emph{triplet-specific models}}, a clear progression is visible. The earlier methods RDV and RiT perform substantially worse than all subsequent approaches and also exhibit very large standard deviations across folds, indicating that with weaker visual representations the task remains highly sensitive to partitioning. More recent triplet-specific models, including SelfD, TERL, and CurConMix, improve this picture considerably, especially at higher resolution and in their ensemble variants. In particular, the ensemble versions consistently outperform their corresponding single-resolution models, with CurConMix (Ens) emerging as the strongest model in this category at 43.6 $AP_{IVT}$. This shows that dedicated triplet modeling does benefit from stronger architectures and ensembling, but even these improved task-specific baselines remain clearly below the strongest overall methods.

A second trend is visible among the \textbf{\emph{foundation-model baselines}}. SurgeNetXL already improves markedly over RDV and RiT, while LemonFM and DinoV3-L further raise performance across unary, pairwise, and triplet prediction, confirming the importance of robust pretrained visual representations for this benchmark. At the same time, stronger backbones alone are not sufficient to explain the best results. The temporal variants of LemonFM and DinoV3-L behave differently: LemonFM-T is consistently weaker than LemonFM across all prediction levels, whereas DinoV3-L-T provides a modest gain over DinoV3-L at the triplet level, improving $AP_{IVT}$ from 43.0 to 44.0. This suggests that simply adding temporal context at the backbone level does not automatically yield stronger interaction reasoning.

The strongest performance is obtained by the \textbf{\emph{SPIRIT variants}}, which outperform both the triplet-specific models and the foundation-model baselines across all prediction levels. SPIRIT-S already surpasses the strongest non-SPIRIT baseline, DinoV3-L-T, improving $AP_{IV}$ from 40.8 to 47.8 (+7.0), $AP_{IT}$ from 51.3 to 57.2 (+5.9), and $AP_{IVT}$ from 44.0 to 48.6 (+4.6), while also raising unary performance from 73.8 to 79.6 for instruments, from 57.8 to 63.9 for verbs, and from 59.5 to 64.9 for targets. SPIRIT-TS pushes these gains further and achieves the best overall results, reaching 49.9 in $AP_{IV}$, 60.1 in $AP_{IT}$, and 51.5 in $AP_{IVT}$. Relative to DinoV3-L-T, this corresponds to gains of +9.1, +8.8, and +7.5, respectively, together with substantial improvements at the unary level. These gains are also accompanied by relatively low standard deviations across folds, especially when compared with RDV and RiT, indicating that the improvements are stable and not tied to a particular train-validation partition. Overall, the cross-validation results show that SPIRIT improves both recognition accuracy and robustness to split variability, with the clearest advantage appearing in pairwise reasoning and final triplet composition.

\begin{table}[!htbp]
\centering
\setlength{\tabcolsep}{7pt}
\caption{\label{all-centers}Results on the triplet recognition task using the official \textit{all-centers} split of MultiBypass-4C-T40. Values are reported as AP (\%). Parentheses indicate resolution; default is 224 unless specified.}
\resizebox{\columnwidth}{!}{
\begin{tabular}{lcccccc}
\toprule
\multirow{2}{*}{\textbf{Method}} &
\multicolumn{3}{c}{\textbf{Component Detection}} &
\multicolumn{3}{c}{\textbf{Triplet Association}} \\
\cmidrule(lr){2-4} \cmidrule(lr){5-7}
& $AP_I$ & $AP_V$ & $AP_T$ & $AP_{IV}$ & $AP_{IT}$ & $AP_{IVT}$ \\
\midrule
\multicolumn{4}{l}{\textit{Triplet Specific Models}} \\
RDV & 42.8 & 32.8 & 37.6 & 21.2 & 28.0 & 21.6 \\
RiT & 46.3 & 34.9 & 39.8 & 24.0 & 32.0 & 26.4 \\

SelfD (224) & 67.2 & 53.3 & 56.5 & 37.8 & 48.0 & 40.6 \\

SelfD (384) & 61.4 & 50.7 & 54.3 & 34.0 & 44.5 & 36.5 \\

SelfD (Ens) & 68.6	& 57.2	& 57.5	& 39.7	& 50.2	& 43.0 \\

TERL (224) & 55.0 & 44.6 & 48.6 & 28.8 & 38.5 & 31.3 \\

TERL (384) & 61.7 & 49.9 & 51.0 & 32.6 & 42.0 & 34.5 \\

TERL (Ens) & 63.1	& 51.6	& 53.6	& 34.4	& 44.3	& 37.0 \\

CurConMix (224) & 65.5	& 55.1	& 56.6	& 38.5	& 48.2	&  40.2 \\

CurConMix (384) & 64.9	& 50.5	& 54.9	& 37.2	& 46.6	& 38.7 \\

CurConMix (Ens) & 68.4	& 56.5	& 58.8	& 40.7	& 50.6	& 42.4 \\

\midrule
\multicolumn{4}{l}{\textit{Foundation Models}} \\
SurgeNetXL & 61.4 & 49.6 & 51.0 & 32.7 & 44.2 & 36.3 \\
LemonFM & 66.2	& 54.3	& 55.4	& 36.3	& 47.6	& 39.7 \\
LemonFM-T & 59.2 & 49.6	& 52.7	& 33.5	& 44.4	& 38.1 \\
DinoV3-L & 70.0 & 57.3 & 55.2 & 40.3 & 50.8 & 42.5 \\
DinoV3-L-T & 66.6 & 53.9 & 52.3 & 37.6 & 47.4 & 41.4 \\

\midrule

\multicolumn{4}{l}{\textit{Our Models}} \\
SPIRIT-S & 78.2 & 62.6 & 58.4 & 46.5 & 55.6 & 47.9 \\
\rowcolor{magenta!15}
SPIRIT-TS & 79.7 & 66.8 & 62.9 & 49.9 & 59.7 & 51.9 \\
\bottomrule
\end{tabular}
}
\end{table}

\begin{table*}[h]
\centering
\setlength{\tabcolsep}{18pt}
\caption{\label{results_challenge}Results on the surgical action triplet recognition task using the official challenge split of MultiBypass-4C-T40. Values are reported as AP (\%) on the public test set / hidden test set. Parentheses indicate resolution; default is 224 unless specified.}
\resizebox{\textwidth}{!}{
\begin{tabular}{lcccccc}
\toprule
\multirow{2}{*}{\textbf{Method}} &
\multicolumn{3}{c}{\textbf{Component Detection}} &
\multicolumn{3}{c}{\textbf{Triplet Association}} \\
\cmidrule(lr){2-4} \cmidrule(lr){5-7}
& $AP_I$ & $AP_V$ & $AP_T$ & $AP_{IV}$ & $AP_{IT}$ & $AP_{IVT}$ \\
\midrule
\multicolumn{4}{l}{\textit{Triplet Specific Models}} \\
RDV & 20.7 / 18.5 & 17.3 / 16.7 & 19.5 / 20.2 & 9.5 / 9.3 & 11.2 / 11.4 & 9.0 / 8.6 \\
RiT & 24.7 / 25.8 & 20.7 / 20.4 & 24.3 / 23.8 & 11.3 / 11.8 & 16.3 / 15.3 & 12.7 / 11.7 \\

SelfD (224) & 65.0 / 47.9 & 47.3 / 36.1 & 46.3 / 45.8 & 32.2 / 24.3 & 41.2 / 33.3 & 34.0 / 25.3 \\
SelfD (384) & 70.5 / 53.2 & 52.9 / 41.8 & 51.5 / 48.5 & 36.7 / 27.8 & 46.4 / 37.9 & 39.4 / 29.7 \\
SelfD (Ens) & 72.3 / 55.7 & 56.2 / 44.2 & 53.6 / 52.6 & 39.8 / 30.2 & 48.9 / 40.5 & 42.3 / 32.2 \\

TERL (224) & 58.6 / 45.3 & 44.4 / 35.4 & 44.2 / 45.4 & 29.6 / 22.6 & 36.8 / 32.0 & 29.8 / 24.1 \\
TERL (384) & 63.3 / 47.8 & 46.8 / 38.1 & 47.7 / 45.5 & 33.3 / 24.8 & 42.1 / 34.9 & 35.4 / 26.8 \\
TERL (Ens) & 65.6 / 50.2 & 49.7 / 40.5 & 49.2 / 48.4 & 35.3 / 26.3 & 43.3 / 36.7 & 36.5 / 28.4 \\

CurConMix (224) & 67.8 / 51.0 & 52.3 / 39.3 & 49.3 / 47.6 & 36.3 / 26.8 & 45.0 / 35.6 & 38.0 / 28.1 \\
CurConMix (384) & 72.0 / 55.8 & 54.8 / 43.1 & 52.0 / 49.9 & 39.6 / 29.9 & 48.4 / 39.3 & 41.1 / 31.7 \\
CurConMix (Ens) & 71.9 / 55.9 & 55.8 / 44.2 & 52.2 / 51.0 & 39.7 / 30.4 & 48.4 / 39.8 & 41.5 / 32.1 \\

\midrule
\multicolumn{4}{l}{\textit{Foundation Models}} \\
SurgeNetXL & 60.0 / 48.3 & 45.0 / 39.2 & 45.5 / 44.0 & 32.4 / 25.4 & 41.3 / 35.6 & 34.7 / 28.6 \\
LemonFM & 67.8 / 57.9	& 51.9 / 47.1 &	51.9 / 50.5	& 37.0  / 31.7	& 45.8 / 42.7 &	38.8 / 35.0 \\
LemonFM-T & 63.8 / 54.8 & 47.4 / 45.5	& 50.3 / 48.8	& 34.1 / 30.1	& 44.1 / 40.1 & 37.9 / 33.4 \\
DinoV3-L & 69.7 / 55.3 & 52.5 / 44.4 & 48.9 / 49.7 & 36.5 / 29.4 & 45.3 / 41.1 & 38.3 / 32.3 \\
DinoV3-L-T & 65.8 / 56.1 & 50.7 / 45.6 & 48.9 / 47.8 & 35.7 / 32.0 & 44.5 / 41.5 & 39.2 / 33.2 \\

\midrule
\multicolumn{4}{l}{\textit{Our Models}} \\
SPIRIT-S & 72.8 / 60.4 & 56.8 / 49.5 & 52.3 / 54.1 & 41.2 / 36.1 & 50.6 / 47.1 & 42.8 / 39.3 \\
\rowcolor{magenta!15}
SPIRIT-TS &  75.9 / 67.2 &	64.8 / 54.8 & 55.5 / 56.3 &	45.8 / 39.7 & 55.0 / 49.8 & 47.3 / 42.3	\\
\bottomrule

\end{tabular}
}
\end{table*}

\subsection{All-centers Results}
Table~\ref{all-centers} reports results on the official \emph{all-centers} split of MultiBypass-4C-T40, which serves as the primary fixed-split benchmark in this study. Because all four centers contribute to both training and evaluation, this setting is less about unseen-center transfer and more about how well different modeling strategies capture fine-grained surgical interactions in a shared multi-centric environment.

Among the \textbf{\emph{Triplet Specific Models}}, a clear performance hierarchy is visible. The earliest methods, RDV and RiT, remain substantially weaker than all later approaches across unary, pairwise, and final triplet prediction, suggesting that lightweight triplet-recognition pipelines struggle to capture the complexity of Roux-en-Y gastric bypass. More recent task-specific methods, including SelfD, TERL, and CurConMix, improve this picture considerably. In particular, increasing the resolution from 224 to 384 often helps, and the ensemble variants are consistently the strongest within each method family. This indicates that dedicated triplet-recognition architectures can benefit substantially from stronger optimization, richer supervision, and model aggregation. Nevertheless, even the strongest methods in this category, such as SelfD (Ens) and CurConMix (Ens), remain below the best-performing models overall, suggesting that triplet-specific design alone is not sufficient if the underlying representation is not strong enough.

A different pattern emerges among the \textbf{\emph{Foundation Models}}. SurgeNetXL already provides a clear jump over the earlier triplet-specific methods, and both LemonFM and DinoV3-L improve further across nearly all prediction levels. This reinforces the importance of strong pretrained visual representations for fine-grained surgical interaction analysis. At the same time, the results also expose a limitation of backbone-only improvements. The temporal extensions LemonFM-T and DinoV3-L-T do not improve over their frame-based counterparts on this split; in fact, both are weaker overall. This suggests that simply adding temporal context at the backbone level does not automatically translate into better triplet reasoning. In a task where the key difficulty lies in organizing interactions rather than merely observing more frames, temporal information needs to be used in a more structured manner.

The advantage of the \textbf{\emph{SPIRIT Variants}} is therefore especially informative. Rather than relying only on stronger appearance features or generic temporal encoding, SPIRIT uses temporal evidence to build better component representations, then refines them through explicit pairwise reasoning before final triplet composition. Even the student variant, SPIRIT-S, moves well beyond the strongest non-SPIRIT baselines, improving over DinoV3-L from 40.3 to 46.5 in $AP_{IV}$, from 50.8 to 55.6 in $AP_{IT}$, and from 42.5 to 47.9 in $AP_{IVT}$. These gains are important because they appear first at the intermediate relational level, before propagating to the final triplet score. The best overall performance is obtained by SPIRIT-TS, which reaches 79.7, 66.8, and 62.9 for $AP_I$, $AP_V$, and $AP_T$, together with 49.9, 59.7, and 51.9 for $AP_{IV}$, $AP_{IT}$, and $AP_{IVT}$. Relative to DinoV3-L, this corresponds to gains of +9.7, +9.5, and +7.7 at the unary level, and +9.6, +8.9, and +9.4 at the relation and triplet levels. Overall, these results suggest that once sufficiently strong visual features are available, the remaining performance bottleneck lies in how effectively the model organizes them into meaningful interaction structure. In the \emph{all-centers} setting, SPIRIT’s explicit modeling of interaction semantics, instrument affordance, and higher-order triplet composition provides a clear and systematic advantage over both dedicated triplet-specific architectures and large pretrained vision models.

\subsection{Challenge Results}
Table~\ref{results_challenge} presents results on the official \emph{challenge} split, which is the most demanding evaluation setting in this study. In this protocol, training is restricted to videos from C1 and C2, while testing is divided into a public set drawn from the same centers and a hidden set that additionally includes unseen videos from C3 and C4. The split is therefore designed to separate standard held-out evaluation from a stronger test of cross-center transfer. In contrast to the \emph{all-centers} setting, success here depends not only on learning fine-grained triplet structure, but on doing so in a way that remains reliable when visual and procedural characteristics shift across institutions.

Among the \textbf{\emph{Triplet Specific Models}}, the same broad hierarchy remains visible, but the overall degradation is substantial. RDV and RiT again perform very poorly, indicating that their interaction modeling is too weak to withstand strong distribution shift. More recent methods, such as SelfD, TERL, and CurConMix, are considerably stronger, and their ensemble variants consistently outperform their single-resolution counterparts. However, even these improved triplet-specific baselines still show a marked drop from the public to the hidden test set, especially at the pairwise and final triplet levels. This suggests that while stronger triplet-specific learning improves in-distribution performance, it does not fully address the harder problem of transferring interaction semantics across centers.

\begin{table*}[!htbp]
\centering
\setlength{\tabcolsep}{4pt}
\caption{\label{results_hidden_centerwise}Per-center performance on the hidden test set of MultiBypass-4C-T40. Each center contains 5 videos. Values are reported as AP (\%). Parentheses indicate resolution; default is 224 unless specified.}
\resizebox{\textwidth}{!}{
\begin{tabular}{
l!{\hspace{3pt}\vrule width 0.8pt\hspace{3pt}}
cccccc!{\hspace{3pt}\vrule width 0.8pt\hspace{3pt}}
cccccc!{\hspace{3pt}\vrule width 0.8pt\hspace{3pt}}
cccccc
}
\toprule
\multirow{2}{*}{\textbf{Method}} &
\multicolumn{6}{c!{\hspace{3pt}\vrule width 0.8pt\hspace{3pt}}}{\textbf{Center1}} &
\multicolumn{6}{c!{\hspace{3pt}\vrule width 0.8pt\hspace{3pt}}}{\textbf{Center3}} &
\multicolumn{6}{c}{\textbf{Center4}} \\
\cmidrule(lr){2-7} \cmidrule(lr){8-13} \cmidrule(lr){14-19}
& $AP_I$ & $AP_V$ & $AP_T$ & $AP_{IV}$ & $AP_{IT}$ & $AP_{IVT}$
& $AP_I$ & $AP_V$ & $AP_T$ & $AP_{IV}$ & $AP_{IT}$ & $AP_{IVT}$
& $AP_I$ & $AP_V$ & $AP_T$ & $AP_{IV}$ & $AP_{IT}$ & $AP_{IVT}$ \\
\midrule
\multicolumn{4}{l}{\textit{Triplet Specific Models}} \\
RDV       & 26.7 & 23.6 & 28.3 & 13.3 & 18.2 & 13.3 & 18.1 & 15.9 & 18.9 & 9.2 & 9.8 & 7.8 & 16.9 & 13.9 & 17.4 & 7.8 & 7.3 & 5.7 \\
RiT              & 38.3 & 32.1 & 33.1 & 17.7 & 23.7 & 17.7 & 21.3 & 18.1 & 23.5 & 10.5 & 13.7 & 10.8 & 18.6 & 14.7 & 19.3 & 7.7 & 7.9 & 6.0 \\

SelfD (224) & 62.0 & 45.1 & 53.5 & 32.1 & 44.6 & 35.0 & 44.5 & 36.7 & 50.9 & 24.4 & 32.3 & 24.1 & 41.9 & 30.7 & 39.9 & 19.3 & 22.0 & 15.9 \\

SelfD (384) & 67.2 & 50.7 & 57.0 & 34.7 & 49.5 & 38.1 & 49.4 & 41.9 & 54.1 & 28.3 & 36.9 & 29.4 & 46.4 & 35.5 & 42.1 & 23.2 & 25.6 & 20.2 \\

SelfD (Ens) &  68.9 & 53.2 & 60.4 & 38.0 & 51.6 & 42.1 & 51.2 & 45.2 & 56.7 & 30.4 & 38.9 & 31.3 & 50.3 & 36.7 & 45.5 & 24.7 & 28.1 & 21.2 \\

TERL (224) &  60.3 & 46.1 & 54.0 & 29.9 & 43.8 & 32.6 & 43.6 & 36.8 & 50.2 & 23.7 & 29.9 & 23.4 & 37.2 & 26.9 & 34.7 & 16.8 & 19.5 & 14.5 \\

TERL (384) &  63.7 & 47.3 & 54.0 & 32.3 & 47.0 & 36.8 & 41.6 & 38.2 & 51.0 & 24.5 & 31.6 & 25.2 & 43.4 & 31.8 & 38.2 & 20.6 & 23.8 & 16.8 \\

TERL (Ens) &  65.7 & 50.1 & 57.1 & 33.4 & 48.9 & 37.8 & 45.6 & 42.1 & 53.5 & 27.2 & 34.0 & 27.4 & 44.2 & 32.2 & 39.2 & 21.2 & 24.7 & 18.1 \\

CurConMix (224) &  61.1 & 46.2 & 55.2 & 32.4 & 46.6 & 37.0 & 49.6 & 42.1 & 53.0 & 27.7 & 33.9 & 27.4 & 47.3 & 35.1 & 41.5 & 23.0 & 24.8 & 18.9 \\

CurConMix (384) &  66.6 & 49.8 & 58.2 & 35.4 & 49.8 & 39.7 & 52.5 & 45.2 & 56.5 & 29.9 & 38.0 & 30.7 & 51.9 & 37.6 & 42.8 & 26.3 & 28.6 & 22.4 \\

CurConMix (Ens) &  66.0 & 50.9 & 58.6 & 35.9 & 50.5 & 40.7 & 54.0 & 47.4 & 57.0 & 31.0 & 38.3 & 31.2 & 52.0 & 38.5 & 44.5 & 26.7 & 28.7 & 22.4 \\

\midrule
\multicolumn{4}{l}{\textit{Foundation Models}} \\
SurgeNetXL       & 63.3 & 49.9 & 51.9 & 34.0 & 47.0 & 37.3 & 43.6 & 39.2 & 49.3 & 24.5 & 33.8 & 28.7 & 42.8 & 31.9 & 41.5 & 20.9 & 24.9 & 19.1 \\
LemonFM & 68.2 & 55 & 58.6 & 36.9 & 50.6 & 40.2 & 53.3 & 48.2 & 54.4 & 31.3 & 40.7 & 33.8 & 54.9 & 40.7 & 46 & 28.1 & 33.2 & 27.2 \\
LemonFM-T &  65.6 & 50.6 & 55 & 34.4 & 48 & 38.6 & 48.6 & 46.3 & 54.6 & 27.9 & 37.2 & 31.7 & 55.4 & 42.5 & 44.3 & 29.1 & 33.4 & 26.1 \\
DinoV3-L         & 66.8 & 51.3 & 57.5 & 34.5 & 50.0 & 38.5 & 50.8 & 44.9 & 55.3 & 29.3 & 39.4 & 31.7 & 49.2 & 39.5 & 44.2 & 26.2 & 30.7 & 24.0 \\
DinoV3-L-T       & 69.1 & 53.4 & 54.2 & 37.8 & 49.6 & 40.3 & 51.6 & 45.9 & 54.7 & 30.4 & 38.1 & 32.6 & 52.1 & 40.6 & 46.2 & 29.9 & 34.5 & 27.0 \\

\midrule
\multicolumn{4}{l}{\textit{Our Models}} \\
SPIRIT-S          & 69.7 & 55.9 & 61.8 & 42.0 & 55.9 & 44.6 & 54.4 & 49.6 & 58.2 & 33.9 & 43.3 & 37.2 & 57.6 & 45.4 & 52.5 & 33.5 & 40.2 & 32.7 \\
\rowcolor{magenta!15}
SPIRIT-TS & 74.0 & 59.2 & 63.3 & 43.3 & 57.7 & 46.7 & 58.6 & 53.5 & 59.7 & 37.3 & 45.4 & 39.0 & 72.3 & 52.7 & 57.2 & 39.1 & 44.1 & 36.4 \\
\bottomrule
\end{tabular}
}
\end{table*}

A second trend emerges within the \textbf{\emph{Foundation Models}}. Relative to the triplet-specific baselines, SurgeNetXL already provides a substantial improvement, and the stronger pretrained models, LemonFM and DinoV3-L, push performance further across both unary recognition and relation prediction. Unlike the \emph{all-centers} split, where temporal backbone variants were not consistently helpful, the \emph{challenge} split shows a more nuanced picture: temporal context appears to provide some benefit under stronger distribution shift, particularly on the hidden test set. For example, DinoV3-L-T improves over DinoV3-L on hidden-test $AP_{IV}$, $AP_{IT}$, and $AP_{IVT}$, suggesting that short-term temporal evidence can support cross-center transfer when the model must disambiguate interactions under unfamiliar visual and procedural conditions. However, these gains remain modest and do not alter the overall ranking of the strongest methods. This indicates that stronger spatio-temporal encoding alone is not enough to solve the core challenge of unseen-center generalization. The main difficulty is not only to observe more context, but to organize that context into coherent instrument-verb-target structure.

The clearest difference emerges in the \textbf{\emph{SPIRIT Variants}}. SPIRIT-S already surpasses all triplet-specific and foundation-model baselines on both the public and hidden test sets, reaching 41.2 / 36.1 in $AP_{IV}$, 50.6 / 47.1 in $AP_{IT}$, and 42.8 / 39.3 in $AP_{IVT}$. These gains are especially meaningful because they appear at the relational level, indicating that the model is better at preserving interaction semantics under domain shift. SPIRIT-TS strengthens this trend further and achieves the best overall performance across nearly all prediction levels, including 45.8 / 39.7 in $AP_{IV}$, 55.0 / 49.8 in $AP_{IT}$, and 47.3 / 42.3 in $AP_{IVT}$. Relative to the strongest non-SPIRIT baselines, the gains are especially pronounced at the pairwise and final triplet levels: on the public test set, $AP_{IVT}$ improves from 42.3 with SelfD (Ens) to 47.3, while on the hidden test set it improves from 35.0 with LemonFM to 42.3. Similar trends are visible for $AP_{IV}$ and $AP_{IT}$, showing that the advantage of SPIRIT is rooted in stronger relational reasoning rather than only better unary recognition. Overall, the \emph{challenge} split shows that SPIRIT’s advantage comes not only from stronger features, but from more robust relational reasoning under institutional variation.

\subsection{Per-center Triplet Performance}
To examine hidden-test generalization more closely, Table~\ref{results_hidden_centerwise} reports results separately for each hidden-test center. We do not include C2 in this analysis because its videos are already part of the public MultiBypass140~\citep{Lavanchy2024} release. The three reported centers play different roles in the challenge protocol: C1 contributes hidden-test videos from a center that is also represented in training, whereas C3 and C4 are entirely absent during training. This distinction is important because it allows us to separate performance on a familiar institutional distribution from performance on genuinely unseen centers. A single hidden-test average can hide this difference, whereas the per-center breakdown reveals how consistently each method generalizes across sites.

Within the \textbf{\emph{Triplet Specific Models}}, the center-wise breakdown reveals two clear effects. First, C1 is consistently the easiest center, whereas C3 and C4 are markedly more difficult. This is expected because, although the hidden-test videos from C1 are not used for training, they still originate from a center whose broader visual and procedural characteristics are represented in the training set. Second, the loss in performance is not uniform across the prediction hierarchy. The unary scores degrade relatively less, while the pairwise and triplet scores decline much more sharply. This indicates that center shift disrupts relational understanding more strongly than isolated component recognition. In practical terms, the model may still recognize which instrument is present, which action is plausible, or which anatomical structure is visible, but struggle to associate them into the correct instrument-verb-target interaction once surgical style, tissue handling, exposure pattern, and camera viewpoint differ from the centers seen during training.

A related pattern is also visible in the \textbf{\emph{Foundation Models}}. Although these models improve substantially over the earlier triplet-specific baselines, the gap between C1 and the unseen centers remains pronounced, especially at the pairwise and triplet levels. The comparison between C3 and C4 adds a further nuance: both centers are entirely unseen during training, yet the magnitude of performance drop is not identical across them. This indicates that unseen-center transfer is not a single uniform phenomenon. Different centers introduce different combinations of visual, anatomical, and procedural variation, and these variations affect relation prediction to different extents. Importantly, this sensitivity is most visible in $AP_{IV}$, $AP_{IT}$, and $AP_{IVT}$ rather than in the unary scores, reinforcing the view that the hardest part of generalization lies in preserving interaction structure rather than simply recognizing entities in isolation.

Against this background, the advantage of the \textbf{\emph{SPIRIT Variants}} becomes even more informative than the raw scores alone suggest. SPIRIT-TS remains the strongest method on every reported center and at every prediction level, but its most meaningful gains again appear in the pairwise and final triplet measures, especially on the unseen centers. On C3, for example, it improves $AP_{IVT}$ to 39.0, compared with 33.8 for LemonFM and 32.6 for DinoV3-L-T, while also raising $AP_{IV}$ and $AP_{IT}$ to 37.3 and 45.4. On C4, where transfer is hardest at the interaction level, SPIRIT-TS reaches 36.4 in $AP_{IVT}$, compared with 27.2 for LemonFM, 27.0 for DinoV3-L-T, and 24.0 for DinoV3-L, together with clear gains in both $AP_{IV}$ and $AP_{IT}$. These improvements are not confined to the final classifier; they indicate that SPIRIT preserves more reliable interaction semantics and instrument-affordance structure under center shift, which then translates into better triplet composition.

The same tendency is already visible with SPIRIT-S, which consistently outperforms the strongest non-SPIRIT baselines on C1, C3, and C4. This shows that the gain is rooted in the structured modeling itself, while teacher-student training further amplifies the benefit, particularly on the unseen centers where robust relational cues matter most. Overall, the per-center analysis strengthens the main claim of the paper: hidden-test transfer in surgical triplet recognition is fundamentally a relational generalization problem, and SPIRIT’s explicit modeling of pairwise interactions and higher-order triplet structure is especially well suited to that challenge.


\subsection{Ablation Study}

We perform an ablation study to assess whether each SPIRIT component improves the level of representation it is intended to model, and whether their combination is necessary for optimal triplet recognition. Starting from the plain multi-task baseline without PIC, TGR or scene context, performance is lowest across all outputs, with $AP_{IV}=36.6$, $AP_{IT}=48.1$ and $AP_{IVT}=39.1$. Adding relational structure through PIC and TGR, even without scene context, already yields a substantial improvement, increasing $AP_{IVT}$ to 46.3 (+7.2 points), together with gains to $AP_{IV}=43.3$ (+6.7) and $AP_{IT}=54.9$ (+6.8). This indicates that explicit interaction modeling is a major source of improvement over flat multi-task learning.

\begin{table}[!h]
\centering
\setlength{\tabcolsep}{3pt}
\caption{\label{tab:ablation_SPIRIT_ticks}Ablation of SPIRIT components on the MultiBypass-4C-T40 \emph{all-centers} split using SPIRIT-S. Values are reported as AP (\%) for unary, pairwise and triplet prediction. PIC: pairwise interaction coupling; TGR: triplet graph reasoning; Scene: scene context enabled.}
\resizebox{\columnwidth}{!}{
\begin{tabular}{ccc ccc ccc}
\toprule
\multicolumn{3}{c}{\textbf{Modules}} &
\multicolumn{3}{c}{\textbf{Component Detection}} &
\multicolumn{3}{c}{\textbf{Triplet Association}} \\
\cmidrule(lr){1-3} \cmidrule(lr){4-6} \cmidrule(lr){7-9}
\textbf{PIC} & \textbf{TGR} & \textbf{Scene}
& $AP_I$ & $AP_V$ & $AP_T$ & $AP_{IV}$ & $AP_{IT}$ & $AP_{IVT}$ \\
\midrule
\xmark      & \xmark      & \xmark      & 69.6 & 54.9 & 52.2 & 36.6 & 48.1 & 39.1 \\
\checkmark  & \checkmark  & \xmark      & 74.4 & 61.4 & 57.0 & 43.3 & 54.9 & 46.3 \\
\checkmark  & \xmark      & \checkmark  & 77.5 & 62.7 & 60.1 & 44.3 & 54.4 & 45.1 \\
\xmark      & \checkmark  & \checkmark  & 78.5 & 62.0 & 57.1 & 44.6 & 54.4 & 45.2 \\
\checkmark  & \checkmark  & \checkmark  & 78.2 & 62.6 & 58.4 & 46.5 & 55.6 & 47.9 \\
\bottomrule
\end{tabular}
}

\end{table}

We then isolate the contributions of PIC and TGR. Enabling PIC with scene context, but removing TGR, produces strong gains in unary and pairwise predictions, reaching $AP_I=77.5$ (+7.9), $AP_V=62.7$ (+7.8), $AP_T=60.1$ (+7.9), $AP_{IV}=44.3$ (+7.7) and $AP_{IT}=54.4$ (+6.3), but yields a lower triplet score of $AP_{IVT}=45.1$. Conversely, retaining TGR with scene context but removing PIC gives similar pairwise scores ($AP_{IV}=44.6$, $AP_{IT}=54.4$) and a comparable triplet score of $AP_{IVT}=45.2$, again below the full model. These findings suggest that PIC primarily strengthens intermediate pairwise interaction representations, whereas TGR contributes more directly to the final triplet composition.

When all three components are enabled, the full SPIRIT model achieves the best overall performance, with $AP_{IV}=46.5$, $AP_{IT}=55.6$ and $AP_{IVT}=47.9$. Relative to the plain baseline, this corresponds to gains of +9.9 points on $AP_{IV}$, +7.5 on $AP_{IT}$ and +8.8 on $AP_{IVT}$, while also improving component detection from 69.6 to 78.2 for instruments, 54.9 to 62.6 for verbs and 52.2 to 58.4 for targets. Overall, the ablation supports the intended design of SPIRIT: PIC improves pairwise interaction learning, TGR strengthens triplet-level reasoning, and scene context provides complementary global cues that further refine final IVT prediction.

\begin{table}[h]
\centering
\setlength{\tabcolsep}{12pt}
\caption{\label{results_challenge_hitk}Top-$N$ triplet recognition results on the official challenge split of MultiBypass-4C-T40. Each entry is reported as Hit@N (\%) on the public test set / hidden test set for final triplet prediction ($IVT$). Parentheses indicate resolution; default is 224 unless specified.}
\resizebox{\columnwidth}{!}{
\begin{tabular}{lccc}
\toprule
\textbf{Method} & $\textbf{Hit@5}_{IVT}$ & $\textbf{Hit@10}_{IVT}$ & $\textbf{Hit@20}_{IVT}$ \\

\midrule
\multicolumn{4}{l}{\textit{Triplet Specific Models}} \\
RDV & 38.1 / 36.5 & 56.7 / 54.4 & 77.5 / 73.9 \\
RiT & 50.6 / 46.5 & 70.7 / 63.2 & 86.9 / 81.1 \\

SelfD (224) & 91.4 / 87.9 & 94.4 / 91.7 & 95.8 / 93.9 \\
SelfD (384) & 93.0 / 89.8 & 95.9 / 93.9 & 97.0 / 95.9 \\
SelfD (Ens) & 94.1 / 91.4 & 96.8 / 95.2 & 97.8 / 97.0 \\

TERL (224) & 87.9 / 86.6 & 93.1 / 92.6 & 96.1 / 95.9 \\
TERL (384) & 90.3 / 87.4 & 94.4 / 92.6 & 96.7 / 95.7 \\
TERL (Ens) & 91.8 / 89.3 & 95.6 / 94.3 & 97.3 / 96.9 \\

CurConMix (224) & 92.7 / 89.7 & 96.1 / 94.3 & 97.3 / 96.4 \\
CurConMix (384) & 94.3 / 91.9 & 97.2 / 96.1 & 98.1 / 97.7 \\
CurConMix (Ens) & 94.2 / 91.9 & 97.2 / 96.2 & 98.2 / 97.9 \\

\midrule
\multicolumn{4}{l}{\textit{Foundation Models}} \\
SurgeNetXL & 87.0 / 84.7 & 93.3 / 92.3 & 97.4 / 96.6 \\
LemonFM & 89.1 / 88.2 &	94.4 / 94.0 &	97.5 / 97.2 \\
LemonFM-T & 89.2 / 88.4	& 95.5 / 94.5	& 98.2	/ 97.5 \\
DinoV3-L & 91.2 / 89.9 & 95.2 / 94.7 & 97.3 / 97.4 \\
DinoV3-L-T & 90.8 / 88.8 & 95.7 / 94.0 & 97.8 / 97.2 \\
\midrule
\multicolumn{4}{l}{\textit{Our Models}} \\
SPIRIT-S & 92.6 / 90.7 & 95.9 / 95.3 & 97.7 / 97.8 \\
\rowcolor{magenta!15}
SPIRIT-TS & 95.3 / 93.5 & 97.4 / 96.8 & 98.5 / 98.5 \\
\bottomrule
\end{tabular}
}
\end{table}

\begin{figure*}[!htbp]
\centering
    \includegraphics[width=1.00\textwidth]{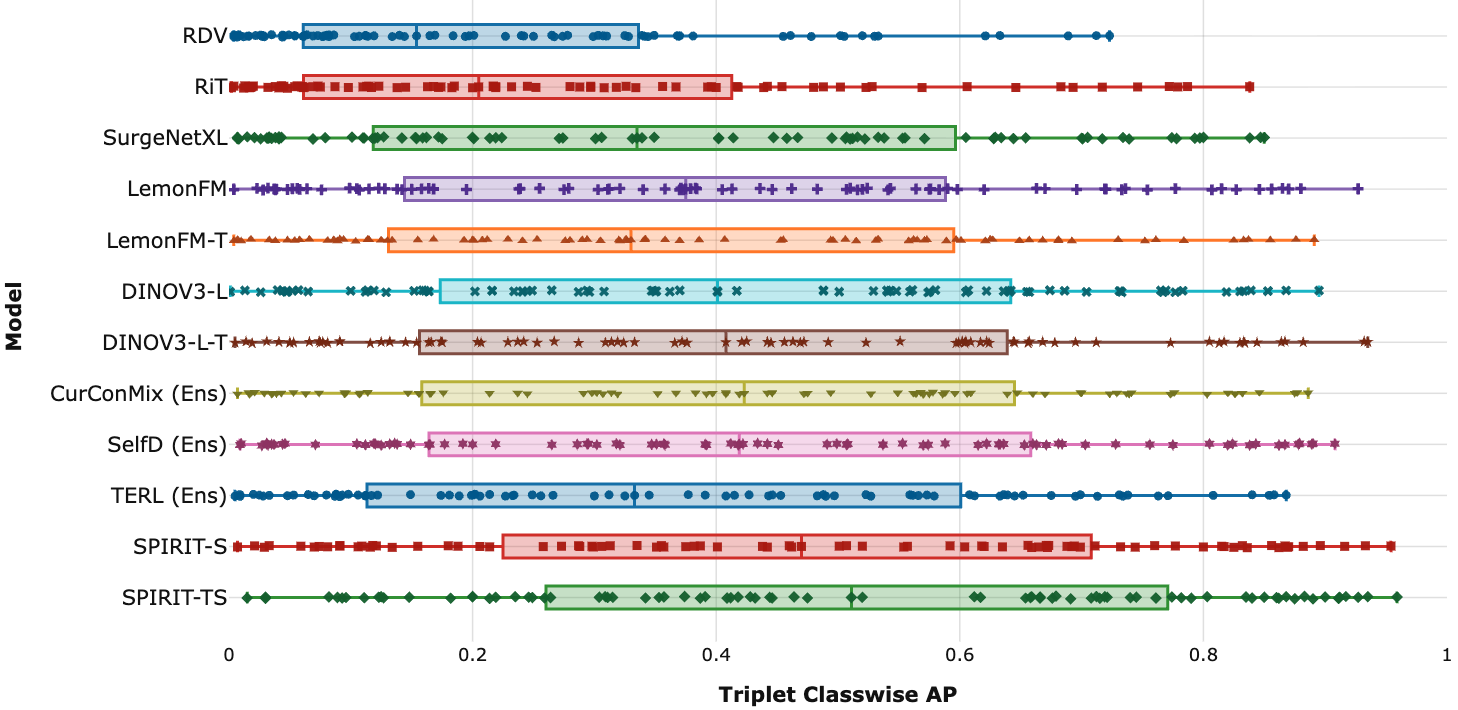} 
    \caption{Boxplot of per-class $AP_{IVT}$ across models on the \emph{all-centers} split of MultiBypass-4C-T40 dataset.}
    \label{fig:ac_box_plot}
\end{figure*}

\subsection{Top-N Triplet Recognition Performance}
While the challenge results quantify recognition quality through AP, they do not directly show how often the correct triplet appears among the highest-ranked predictions. We therefore report Hit@N to complement mAP with a more explicit ranking-based view of triplet retrieval performance. For each frame, the model outputs a ranked list of triplet classes, and a prediction is counted as correct if any ground-truth positive triplet label appears among the top-$N$ predictions. Hit@N is then averaged over all evaluated frames containing at least one positive label.

As shown in Table~\ref{results_challenge_hitk}, SPIRIT-TS now achieves the strongest performance across all reported Top-$N$ settings on both the public and hidden test sets. It reaches 95.3 / 93.5 at Hit@5, 97.4 / 96.8 at Hit@10, and 98.5 / 98.5 at Hit@20, indicating that the correct triplet is ranked very highly with remarkable consistency. Relative to the strongest non-SPIRIT baseline for each metric, these results correspond to gains of +1.0 / +1.6 points at Hit@5 over CurConMix (384), +0.2 / +0.6 points at Hit@10 over CurConMix (384) and CurConMix (Ens), respectively, and +0.3 / +0.6 points at Hit@20 over CurConMix (Ens). The margins are naturally smaller at larger values of $N$, where several strong methods already retrieve the correct triplet with high reliability, but SPIRIT-TS remains the best in every case.

These results reinforce the findings from the AP-based evaluation. The advantage of SPIRIT is not limited to improving average recognition quality, but also extends to ranking quality, especially in the practically important low-$N$ regime where only a small number of candidate triplets are considered. Taken together, the Top-$N$ results show that SPIRIT not only produces stronger triplet predictions overall, but also places the correct interaction more reliably among the highest-ranked candidates under both standard and hidden-test evaluation.

\begin{table}[!htbp]
\centering
\scriptsize
\setlength{\tabcolsep}{4pt}
\caption{\label{tab:ablation_temporal_only}Temporal modeling ablation on MultiBypass-4C-T40 (AP \%) dataset for \emph{all-centers} split.}
\resizebox{\columnwidth}{!}{
\begin{tabular}{lcccccccc}
\toprule
\textbf{Method} & \textbf{Temporal} & $I$ & $V$ & $T$ & $IV$ & $IT$ & $IVT$ \\
\midrule
SPIRIT-S     & \xmark & 73.0 & 59.1 & 53.3 & 41.1 & 51.0 & 41.9 \\
SPIRIT-S     & \checkmark & 78.2 & 62.6 & 58.4 & 46.5 & 55.6 & 47.9 \\
SPIRIT-TS    & \xmark & 74.9 & 61.4 & 56.3 & 43.9 & 53.8 & 45.0 \\
SPIRIT-TS    & \checkmark & 79.7 & 66.8 & 62.9 & 49.9 & 59.7 & 51.9 \\
\bottomrule
\end{tabular}
}
\end{table}

\begin{figure*}[h]
\centering
    \includegraphics[width=0.90\textwidth]{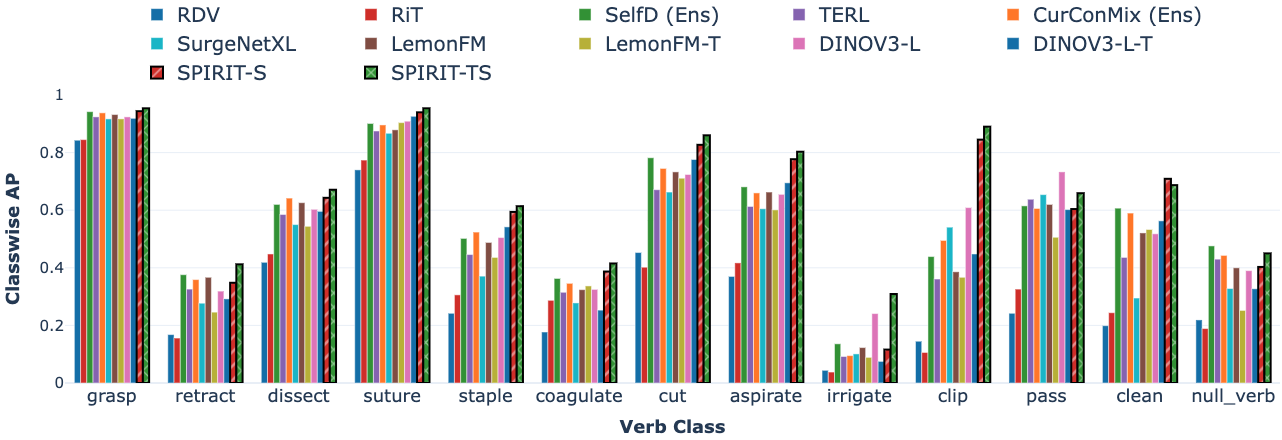} 
    \caption{Per-class performance comparison across models on verb component on the \emph{all-centers} split of MultiBypass-4C-T40 dataset.}
    \label{fig:per_class_verb_perf}
\end{figure*}

\subsection{Effect of temporal modeling}

We further perform a temporal modeling ablation to assess whether the gains of SPIRIT depend on access to short-term spatio-temporal context beyond single-frame reasoning. As shown in Table~\ref{tab:ablation_temporal_only}, enabling temporal modeling consistently improves performance for both the student and teacher-student variants across all prediction levels. For SPIRIT-S, temporal modeling increases $AP_{IVT}$ from 41.9 to 47.9 (+6.0), together with gains of +5.4 on $AP_{IV}$ and +4.6 on $AP_{IT}$. A similar pattern is observed for SPIRIT-TS, where $AP_{IVT}$ rises from 45.0 to 51.9 (+6.9), while $AP_{IV}$ and $AP_{IT}$ improve by +6.0 and +5.9, respectively. The unary predictions also improve consistently in both settings, indicating that temporal context helps disambiguate not only the final triplet, but also its individual instrument, verb and target components.

These gains are particularly meaningful because both pairwise relations in SPIRIT depend on temporal context. The instrument-target branch must determine \emph{where} the action is being applied, which often becomes clearer only when the interaction is observed over multiple frames rather than from a single image. Likewise, the instrument-verb branch must determine \emph{what} action the instrument is performing, and this is often better distinguished from the temporal evolution of the tool motion than from static appearance alone. By incorporating spatio-temporal context into the unary features, SPIRIT produces more informative instrument, verb and target representations, which in turn improves both interaction semantics and instrument affordance modeling. This stronger pairwise reasoning then supports more reliable triplet composition, leading to higher final triplet recognition performance. Overall, these results show that temporal modeling is not merely an auxiliary enhancement, but a key factor in improving both component-level recognition and the structured reasoning required for coherent triplet prediction.

\subsection{Per-class Triplet Performance Variation}
To assess whether the gains of SPIRIT are confined to a small subset of triplet classes or distributed more broadly across the label space, we compare the distribution of per-class $AP_{IVT}$ across methods using the box plots in Fig.~\ref{fig:ac_box_plot}. Both SPIRIT variants show a clear upward shift relative to the compared baselines, with SPIRIT-TS attaining the highest median and SPIRIT-S also consistently outperforming the remaining methods. This indicates that the improvement brought by SPIRIT is not driven only by a few easy or dominant classes, but generalizes across a broader portion of the triplet taxonomy.

At the same time, the spread of the box plots shows that substantial class-level variability remains, which is expected given the long-tail distribution and semantic diversity of fine-grained surgical interactions. Some triplet classes remain consistently more difficult than others across all methods, reflecting persistent challenges in recognizing rare interactions and visually ambiguous compositions. Nevertheless, the higher median and overall upward shift observed for SPIRIT-TS, followed by SPIRIT-S, suggest that explicit relational modeling improves class-wise recognition in a more uniform manner rather than benefiting only a narrow subset of triplets. In this sense, Fig.~\ref{fig:ac_box_plot} complements the aggregate AP results by showing that the gains of SPIRIT are reflected not only in the mean performance, but also in the broader distribution of per-class recognition quality.

\subsection{Per-class Verb Performance}

To better understand how different models capture action semantics beyond aggregate mAP, we analyze per-class verb recognition performance using the class-wise $AP_V$ distribution shown in Fig.~\ref{fig:per_class_verb_perf}. This analysis is particularly important in Roux-en-Y gastric bypass, where verbs differ substantially in procedural complexity. Actions such as \emph{grasp} and \emph{retract} are comparatively easier to recognize because they are frequent, visually salient, and often characterized by stable instrument-anatomy configurations. By contrast, \emph{suture} and \emph{staple} are intrinsically more difficult. Suturing is a temporally extended maneuver that unfolds over multiple phases, including needle positioning, tissue approximation, passage, tightening and release, and often involves coordinated use of more than one instrument. Stapling is likewise more complex than a simple local interaction, because it typically occurs at critical procedural stages, requires precise alignment of the stapler with the target anatomy, and may be preceded and followed by substantial preparatory and verification motions. In both cases, the visual appearance of the action is therefore distributed over time and depends strongly on context, making recognition harder than for simpler verbs whose meaning is apparent from a single local interaction.

The resulting trends are consistent with this distinction. While all strong models perform well on common verbs such as \emph{grasp}, SPIRIT shows clearer gains on verbs that require richer temporal and relational understanding. Compared with DinoV3-L, SPIRIT-TS improves from 0.505 to 0.614 on \emph{staple}, from 0.603 to 0.671 on \emph{dissect}, from 0.325 to 0.415 on \emph{coagulate}, and from 0.609 to 0.890 on \emph{clip}, while also maintaining very strong performance on \emph{suture} (0.954). At the same time, not all classes follow the same pattern: for \emph{pass}, the strongest score is obtained by the plain DinoV3-L backbone (0.733), exceeding both DinoV3-L-T and the SPIRIT variants, which suggests that some verbs may depend more strongly on distinctive local appearance cues than on explicit relational composition. Rare verbs such as \emph{irrigate} also remain difficult for all methods, reflecting the long-tail nature of the label space. Overall, Fig.~\ref{fig:per_class_verb_perf} shows that the benefit of SPIRIT is especially meaningful for verbs whose recognition depends on coordinated instrument use, longer temporal context and procedure-specific interaction semantics.

\begin{figure*}[!htbp]
\centering
    \includegraphics[width=1.00\textwidth]{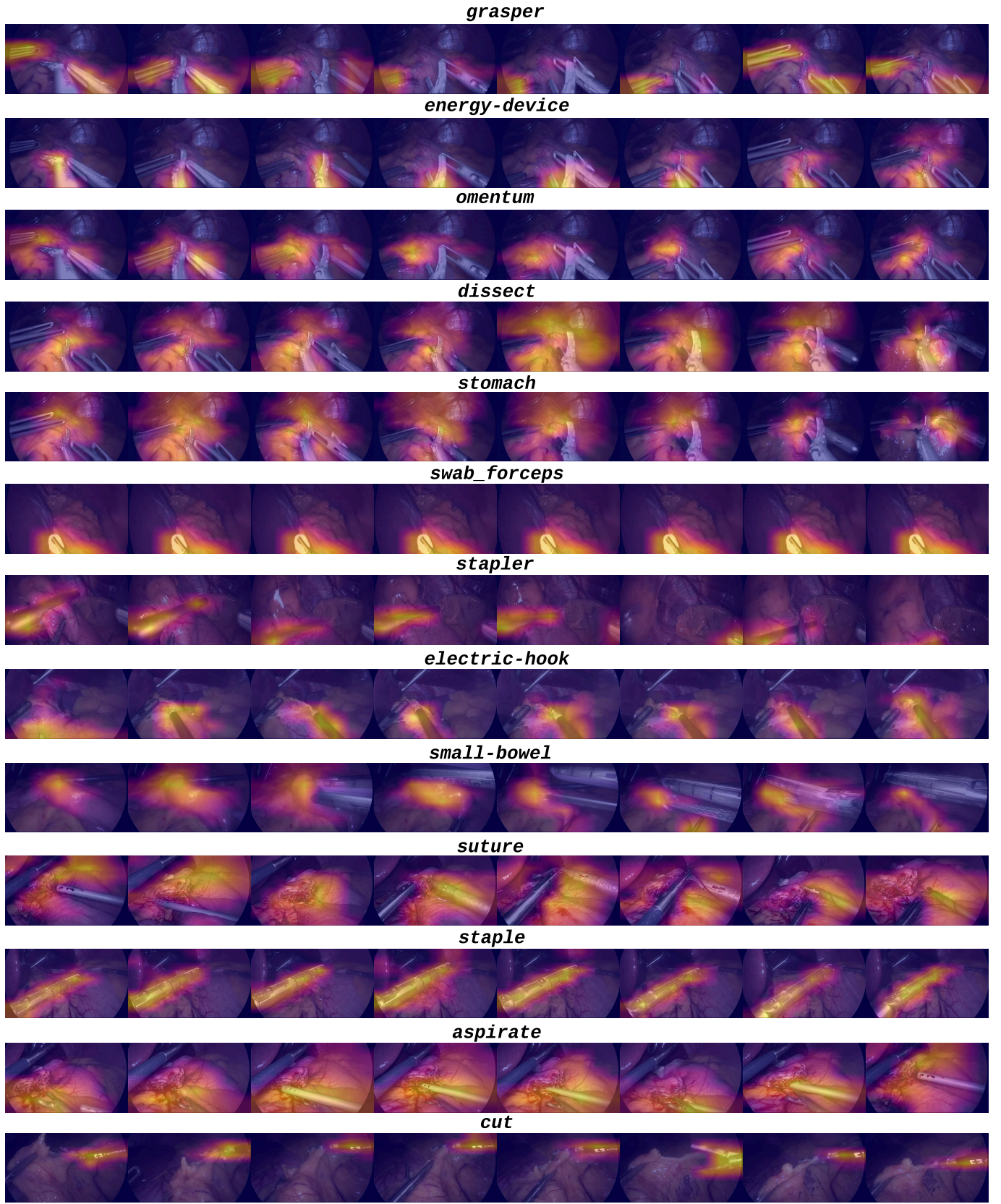} 
    \caption{Visualization of component-wise spatio-temporal cross-attention maps in the TUF block of SPIRIT on the MultiBypass-4C-T40 dataset.}
    \label{fig:ac_attn_map_per_component}
\end{figure*}

\subsection{Qualitative Results}
Figure~\ref{fig:ac_attn_map_per_component} visualizes the component-wise attention maps obtained from cross-attention between the text-conditioned class queries and the spatio-temporal visual tokens of an input clip. These maps provide qualitative evidence that SPIRIT does not simply recognize isolated class labels, but learns class-specific spatial focus patterns that evolve across time for instruments, verbs and targets. In particular, the highlighted regions show that the unary branches are already aligned with the temporally structured visual evidence most relevant to downstream relational reasoning.

Several examples illustrate this behavior. For the instrument branches, the model localizes both the very frequent \emph{grasper} class and the substantially less frequent \emph{energy-device} class, even though the latter appears much less often in the training data. This suggests that the learned unary representations are not restricted to dominant instrument categories, but remain spatially meaningful even for comparatively underrepresented tools. For the verb branch, the attention map for \emph{dissect} concentrates on the region where the action is being performed, while also spreading over the local trajectory of the interacting instrument, suggesting that the model captures not only the acted-upon anatomy but also the motion context associated with the verb. Finally, for the target branches, the attention maps for \emph{omentum} and \emph{stomach} localize the corresponding anatomical structures with good specificity, indicating that the model learns target-aware representations that can support subsequent pairwise interaction coupling and triplet composition. Overall, these qualitative results suggest that SPIRIT learns semantically meaningful spatial grounding for the individual triplet components, which is consistent with the improvements observed in the quantitative analysis.


\subsection{Rare vs Frequent Triplet Performance}
Evaluating only the overall triplet mAP can obscure an important aspect of model behavior, namely whether performance gains are driven mainly by common interactions or whether they also extend to the long tail of infrequent triplet classes. This distinction is especially relevant in MultiBypass-4C-T40, where the triplet distribution is highly imbalanced and several clinically meaningful interactions occur only sparsely. To examine this effect more explicitly, we divide triplet classes into \emph{rare} and \emph{frequent} groups based on their frame-level support in the training set. Triplets with fewer than 200 frame instances are treated as rare, while the remaining classes are considered frequent. Table~\ref{results_rare_vs_nonrare} reports the resulting $AP_{IVT}$ for both groups on the \emph{all-centers} split.

A clear trend emerges across all three method groups. First, every model performs substantially better on frequent triplets than on rare ones, confirming that long-tail triplet recognition remains one of the main difficulties of this benchmark. Within the \textbf{\emph{Triplet Specific Models}}, stronger recent methods such as SelfD, TERL, and CurConMix improve considerably over RDV and RiT in both groups, with their ensemble variants generally providing the best results. 
Within the \textbf{\emph{Foundation Models}}, DinoV3-L gives the strongest rare-triplet performance at 26.6, outperforming both LemonFM variants despite only moderate differences on the frequent subset. This suggests that stronger visual pretraining alone already helps the long tail to some extent, but does not close the gap fully.

\begin{table}[h]
\centering
\setlength{\tabcolsep}{24pt}
\caption{\label{results_rare_vs_nonrare}Results on $IVT$ triplet prediction for rare and frequent triplet classes on the MultiBypass-4C-T40 \emph{all-centers} split. Parentheses indicate resolution; the default is 224 unless specified.}
\resizebox{\columnwidth}{!}{
\begin{tabular}{lcc}
\toprule
\textbf{Method} & $\textbf{AP}_{IVT}^{\text{Frequent}}$ & $\textbf{AP}_{IVT}^{\text{Rare}}$ \\
\midrule
\multicolumn{3}{l}{\textit{Triplet Specific Models}} \\
RDV & 25.9 & 10.3 \\
RiT & 32.7 & 10.1 \\
SelfD (224) & 50.1 & 22.6 \\
SelfD (384) & 46.1 & 17.9 \\
SelfD (Ens) & 52.5 & 26.0 \\
TERL (224) & 40.4 & 19.4 \\
TERL (384) & 43.9 & 22.3 \\
TERL (Ens) & 45.9 & 25.4 \\
CurConMix (224) & 50.0 & 23.0 \\
CurConMix (384) & 47.8 & 21.2 \\
CurConMix (Ens) & 52.1 & 24.4 \\
\midrule
\multicolumn{3}{l}{\textit{Foundation Models}} \\
SurgeNetXL & 43.2 & 18.5 \\
LemonFM & 47.3 & 19.8 \\
LemonFM-T & 45.6 & 18.5 \\
DinoV3-L & 48.6 & 26.6 \\
DinoV3-L-T & 48.2 & 23.6 \\
\midrule
\multicolumn{3}{l}{\textit{Our Models}} \\
SPIRIT-S & 54.2 & 31.5 \\
\rowcolor{magenta!15}
SPIRIT-TS & 58.6 & 34.5 \\
\bottomrule
\end{tabular}
}
\end{table}

The most important pattern is again visible in the \textbf{\emph{SPIRIT Variants}}. SPIRIT-S and SPIRIT-TS achieve the best performance on both frequent and rare triplets, with SPIRIT-TS reaching 58.6 and 34.5, respectively. Relative to DinoV3-L, this corresponds to an improvement of +10.0 points on frequent triplets and +7.9 points on rare triplets. The rare-triplet result is particularly important because it shows that the benefit of SPIRIT is not confined to dominant interaction patterns. Instead, explicit modeling of pairwise interaction semantics and instrument affordance appears to help the model generalize better when training examples are sparse and direct visual repetition is limited. Overall, these results indicate that SPIRIT strengthens performance not only on well-represented triplets, but also on the long-tail portion of the label space, which is often the more clinically challenging part of the task.

\subsection{Computational Complexity Analysis}

In addition to recognition accuracy, we analyze the computational complexity of all compared methods to better characterize their practical training and deployment cost. This is important because models with similar predictive performance can differ substantially in optimization footprint, forward-pass computation, and runtime efficiency, especially in our setting where the comparison spans lightweight triplet-specific architectures, large foundation-model baselines, temporal extensions, and ensemble variants. To provide a fair comparison, all measurements are obtained under a unified benchmarking protocol on NVIDIA A100 GPUs.


We report four complementary metrics: \emph{Trainable Params (M)}, \emph{FLOPs (GFLOPs/sample)}, \emph{Training Cost (ms/iter)}, and \emph{Inference Latency (ms/sample)}. Trainable parameters count only the weights updated during optimization, FLOPs measure per-sample forward-pass compute, and the runtime metrics capture practical training and deployment cost. Training cost is measured in milliseconds per iteration using each model's default batch size, whereas inference latency is measured with batch size 1 in milliseconds per sample. To ensure stable timing, measurements are recorded after warm-up and averaged over steady-state iterations. Ensemble variants are reported as the summed cost of their constituent models, reflecting their effective resource requirement when used jointly.

The results show several clear trends. Lightweight triplet-specific baselines such as RDV and TERL (224), together with SurgeNetXL, remain the most efficient in runtime and inference latency, making them attractive when resource usage is the primary constraint. As expected, increasing the input resolution from 224 to 384 substantially increases compute and training cost for SelfD, TERL, and CurConMix. Larger foundation-model baselines also incur a markedly higher optimization footprint than these lightweight alternatives, with DinoV3-L and LemonFM requiring substantially more trainable parameters and wall-clock training time.

Temporal modeling generally increases runtime cost, but its effect on FLOPs is not uniform across architectures. For example, both LemonFM-T and DinoV3-L-T are much slower than their spatial counterparts, while only DinoV3-L-T shows a large increase in FLOPs and inference latency. The SPIRIT variants follow the same broad trend of higher computational demand due to temporal reasoning and structured relational modeling. At the same time, SPIRIT-S and SPIRIT-TS have nearly identical parameter count, FLOPs, training cost, and inference latency, showing that the improvement of SPIRIT-TS is achieved without materially increasing architectural complexity over SPIRIT-S. This indicates that the performance gain of SPIRIT-TS comes from more effective learning rather than from scaling model size or compute budget.

\begin{table}[h]
\centering
\caption{Computational complexity comparison on NVIDIA A100 GPUs. Trainable parameters are reported in millions (M), FLOPs in GFLOPs per sample, training cost in ms/iter using each model's default training batch size, and inference latency in ms/sample with batch size 1. Ensemble variants are reported as the summed cost of their constituent models.}
\label{comp_complexity}
\setlength{\tabcolsep}{8.5pt} 
\resizebox{\columnwidth}{!}{
\begin{tabular}{lcccc}
\toprule
\textbf{Method} &
\makecell{\textbf{Trainable} \\ \textbf{Params (M)}} &
\makecell{\textbf{FLOPs} \\ \textbf{(G/sample)}} &
\makecell{\textbf{Train Cost} \\ \textbf{(ms/iter)}} &
\makecell{\textbf{Inf. Latency} \\ \textbf{(ms/sample)}} \\
\midrule
\multicolumn{5}{l}{\textit{Triplet Specific Models}} \\
RDV                & 16.37  & 2.19   & 40.58   & 13.50 \\
RiT                & 16.37  & 17.49  & 49.43   & 14.73 \\
SelfD (224)        & 87.23  & 15.47  & 512.50  & 14.41 \\
SelfD (384)        & 87.36  & 47.19  & 1603.78 & 16.21 \\
SelfD (Ens)        & 370.44 & 128.91 & 4148.50 & 45.07 \\
TERL (224)         & 28.80  & 4.51   & 54.93   & 9.04 \\
TERL (384)         & 89.11  & 47.21  & 118.46  & 13.82 \\
TERL (Ens)         & 117.91 & 51.72  & 173.39  & 22.86 \\
CurConMix (224)    & 87.23  & 15.47  & 498.54  & 14.74 \\
CurConMix (384)    & 87.36  & 47.19  & 1578.86 & 14.52 \\
CurConMix (Ens)    & 174.59 & 62.66  & 2077.40 & 29.26 \\
\midrule
\multicolumn{5}{l}{\textit{Foundation Models}} \\
SurgeNetXL         & 23.30  & 4.12   & 38.18   & 6.68 \\
LemonFM            & 196.42 & 275.23 & 286.90  & 23.07 \\
LemonFM-T          & 212.17 & 275.28 & 1121.99 & 23.31 \\
DinoV3-L           & 303.28 & 60.90  & 216.11  & 13.28 \\
DinoV3-L-T         & 311.68 & 487.27 & 1544.84 & 40.18 \\
\midrule
\multicolumn{5}{l}{\textit{Our Models}} \\
SPIRIT-S           & 103.21 & 487.47 & 920.34  & 46.58 \\
SPIRIT-TS          & 103.21 & 487.47 & 921.78  & 46.68 \\
\bottomrule
\end{tabular}
}
\end{table}

\subsection{Statistical Significance Analysis}
To assess whether the improvement of \textbf{SPIRIT-TS} over \textbf{DinoV3-L-T} is statistically reliable, we perform a paired video-level significance analysis on the \emph{challenge} test split of MultiBypass-4C-T40 using $AP_{IVT}$ as the primary metric. We choose DinoV3-L-T as the baseline because it is the strongest and most relevant reference model, and SPIRIT-TS is built on top of the same backbone family with additional modeling components. Since statistical testing requires multiple paired observations rather than a single aggregate score, we compute the video-level overall $AP_{IVT}$ for each of the 9 test videos and compare the two models on a per-video basis.

Across the 9 challenge-test videos, SPIRIT-TS improves over DinoV3-L-T by an average of +6.53 percentage points in video-level overall $AP_{IVT}$. A paired permutation test yields a p-value of $0.00355$, indicating that this improvement is unlikely to have occurred by chance. In addition, bootstrap resampling gives a 95\% confidence interval of [4.89, 8.01] percentage points for the mean paired improvement. These results show that the gain of SPIRIT-TS is statistically significant and consistently observed across videos.

%% file: sections/06-conclusion.tex
\section{Conclusion}
In this work, we addressed the problem of learning surgical action triplet representations that generalize reliably across institutions by proposing \textbf{SPIRIT}, a structured framework that models surgical actions progressively through spatio-temporal unary representations, pairwise interaction coupling, higher-order triplet graph reasoning, and multi-head distillation. To evaluate this setting rigorously, we established \textbf{MultiBypass-4C-T40}, a multi-centric benchmark for fine-grained surgical action triplet recognition in Roux-en-Y gastric bypass, spanning four geographically distinct centers with dense triplet annotations accompanied by phase- and step-level labels. By extending prior multi-centric workflow analysis from coarse-grained procedural labels to fine-grained instrument-verb-target interactions, the dataset enables explicit study of triplet recognition under realistic institutional variation. Across the \emph{all-centers}, \emph{cross-validation}, and \emph{challenge} settings defined on this benchmark, SPIRIT consistently outperformed strong triplet-specific baselines and recent surgical foundation models, with the largest gains appearing at the pairwise and final triplet levels. These results show that explicitly modeling intermediate interaction structure improves robustness under center shift and leads to more transferable triplet representations. Together, SPIRIT and MultiBypass-4C-T40 provide a methodological and benchmarking foundation for advancing fine-grained surgical scene understanding toward more generalizable and clinically relevant surgical AI.

%% file: sections/07-acknowledgement.tex
\section*{Acknowledgment}
This work has received funding from the European Union (ERC, CompSURG, 101088553). Views and opinions expressed are however those of the authors only and do not necessarily reflect those of the European Union or the European Research Council. Neither the European Union nor the granting authority can be held responsible for them. 
This work was also partially supported by French state funds managed by the ANR under Grants ANR-22-FAI1-0001 (project DAIOR), ANR-10-IAHU-02 (IHU Strasbourg) and by the Interdisciplinary Thematic Institute HealthTech (ITI 2021-2028 program of the University of Strasbourg, CNRS and Inserm) via the IdEx Unistra (ANR-10-IDEX-0002) and SFRI (STRATUS project, ANR-20-SFRI-0012).
Joël L. Lavanchy was funded by the Swiss National Science Foundation (P5R5PM 21766), the Novartis Foundation for medical-biological Research (23C162) and by the Vontobel Foundation (0867/2024).
This work was also granted access to the servers/HPC resources managed by CAMMA, IHU Strasbourg, Unistra Mesocentre.

\section*{Declaration}
During the preparation of this work the author used GPT4 in order to perform minor grammar checks. After using this tool, the author reviewed and edited the content as needed and take full responsibility for the content of the published article.

%% file: sections/supp.tex
\onecolumn

\begin{center}
{\Large\bfseries ===== Appendix =====}
\end{center}


\setcounter{table}{0}
\renewcommand{\thetable}{A\arabic{table}}
\setcounter{figure}{0}
\renewcommand{\thefigure}{A\arabic{figure}}

\begin{table*}[!htbp]
\centering
\small
\setlength{\tabcolsep}{4pt}
\renewcommand{\arraystretch}{1.2}
\caption{Center-wise triplet frequency table. Green cells denote triplet frequencies; red-shaded cells denote absent classes. Abbreviations used in table entries: SB-A=small bowel anastomosis, SBS-A=small bowel stomach anastomosis, RAD=right angle dissector, IA=irrigator aspirator.}
\label{tab:center_triplet_presence}
\begin{tabular}{>{\raggedright\arraybackslash}p{0.28\linewidth}>{\centering\arraybackslash}p{0.035\linewidth}>{\centering\arraybackslash}p{0.035\linewidth}>{\centering\arraybackslash}p{0.035\linewidth}>{\centering\arraybackslash}p{0.035\linewidth}|>{\raggedright\arraybackslash}p{0.28\linewidth}>{\centering\arraybackslash}p{0.035\linewidth}>{\centering\arraybackslash}p{0.035\linewidth}>{\centering\arraybackslash}p{0.035\linewidth}>{\centering\arraybackslash}p{0.035\linewidth}}
\hline
Triplet & C1 & C2 & C3 & C4 & Triplet & C1 & C2 & C3 & C4 \\
\hline
0. grasper,grasp,small\_bowel & \cellcolor{green!20}\textcolor{black}{16369} & \cellcolor{green!20}\textcolor{black}{13338} & \cellcolor{green!20}\textcolor{black}{3895} & \cellcolor{green!20}\textcolor{black}{8301} & 45. energy\_device,dissect,omentum & \cellcolor{green!20}\textcolor{black}{764} & \cellcolor{green!20}\textcolor{black}{803} & \cellcolor{green!20}\textcolor{black}{305} & \cellcolor{green!20}\textcolor{black}{662} \\
1. grasper,grasp,stomach & \cellcolor{green!20}\textcolor{black}{17137} & \cellcolor{green!20}\textcolor{black}{10347} & \cellcolor{green!20}\textcolor{black}{4282} & \cellcolor{green!20}\textcolor{black}{4626} & 46. energy\_device,dissect,mesentery & \cellcolor{green!20}\textcolor{black}{35} & \cellcolor{green!20}\textcolor{black}{404} & \cellcolor{green!20}\textcolor{black}{45} & \cellcolor{green!20}\textcolor{black}{123} \\
2. grasper,grasp,thread & \cellcolor{green!20}\textcolor{black}{24512} & \cellcolor{green!20}\textcolor{black}{9591} & \cellcolor{green!20}\textcolor{black}{7928} & \cellcolor{green!20}\textcolor{black}{2211} & 47. energy\_device,cut,small\_bowel & \cellcolor{red!20}\textcolor{black}{0} & \cellcolor{green!20}\textcolor{black}{360} & \cellcolor{red!20}\textcolor{black}{0} & \cellcolor{green!20}\textcolor{black}{192} \\
3. grasper,grasp,omentum & \cellcolor{green!20}\textcolor{black}{8913} & \cellcolor{green!20}\textcolor{black}{4763} & \cellcolor{green!20}\textcolor{black}{1835} & \cellcolor{green!20}\textcolor{black}{4772} & 48. energy\_device,cut,stomach & \cellcolor{green!20}\textcolor{black}{136} & \cellcolor{green!20}\textcolor{black}{177} & \cellcolor{red!20}\textcolor{black}{0} & \cellcolor{green!20}\textcolor{black}{188} \\
4. grasper,grasp,liver & \cellcolor{green!20}\textcolor{black}{61} & \cellcolor{green!20}\textcolor{black}{117} & \cellcolor{red!20}\textcolor{black}{0} & \cellcolor{red!20}\textcolor{black}{0} & 49. energy\_device,cut,omentum & \cellcolor{green!20}\textcolor{black}{2467} & \cellcolor{green!20}\textcolor{black}{2120} & \cellcolor{green!20}\textcolor{black}{811} & \cellcolor{green!20}\textcolor{black}{840} \\
5. grasper,grasp,needle & \cellcolor{green!20}\textcolor{black}{3161} & \cellcolor{green!20}\textcolor{black}{3277} & \cellcolor{green!20}\textcolor{black}{1064} & \cellcolor{green!20}\textcolor{black}{266} & 50. energy\_device,cut,mesentery & \cellcolor{green!20}\textcolor{black}{163} & \cellcolor{green!20}\textcolor{black}{429} & \cellcolor{green!20}\textcolor{black}{112} & \cellcolor{green!20}\textcolor{black}{233} \\
6. grasper,grasp,mesentery & \cellcolor{green!20}\textcolor{black}{4296} & \cellcolor{green!20}\textcolor{black}{812} & \cellcolor{green!20}\textcolor{black}{609} & \cellcolor{green!20}\textcolor{black}{103} & 51. energy\_device,cut,adhesion & \cellcolor{green!20}\textcolor{black}{152} & \cellcolor{green!20}\textcolor{black}{391} & \cellcolor{red!20}\textcolor{black}{0} & \cellcolor{red!20}\textcolor{black}{0} \\
7. grasper,grasp,colon & \cellcolor{green!20}\textcolor{black}{75} & \cellcolor{green!20}\textcolor{black}{55} & \cellcolor{green!20}\textcolor{black}{60} & \cellcolor{green!20}\textcolor{black}{39} & 52. energy\_device,coagulate,stomach & \cellcolor{green!20}\textcolor{black}{422} & \cellcolor{green!20}\textcolor{black}{147} & \cellcolor{green!20}\textcolor{black}{35} & \cellcolor{green!20}\textcolor{black}{33} \\
8. grasper,retract,small\_bowel & \cellcolor{green!20}\textcolor{black}{1576} & \cellcolor{green!20}\textcolor{black}{1372} & \cellcolor{green!20}\textcolor{black}{696} & \cellcolor{green!20}\textcolor{black}{597} & 53. energy\_device,coagulate,omentum & \cellcolor{green!20}\textcolor{black}{197} & \cellcolor{green!20}\textcolor{black}{73} & \cellcolor{green!20}\textcolor{black}{81} & \cellcolor{green!20}\textcolor{black}{186} \\
9. grasper,retract,stomach & \cellcolor{green!20}\textcolor{black}{1920} & \cellcolor{green!20}\textcolor{black}{2256} & \cellcolor{green!20}\textcolor{black}{798} & \cellcolor{green!20}\textcolor{black}{2105} & 54. IA,retract,small\_bowel & \cellcolor{red!20}\textcolor{black}{0} & \cellcolor{green!20}\textcolor{black}{42} & \cellcolor{green!20}\textcolor{black}{33} & \cellcolor{green!20}\textcolor{black}{136} \\
10. grasper,retract,omentum & \cellcolor{green!20}\textcolor{black}{1281} & \cellcolor{green!20}\textcolor{black}{1629} & \cellcolor{green!20}\textcolor{black}{680} & \cellcolor{green!20}\textcolor{black}{330} & 55. IA,retract,stomach & \cellcolor{green!20}\textcolor{black}{74} & \cellcolor{green!20}\textcolor{black}{27} & \cellcolor{green!20}\textcolor{black}{19} & \cellcolor{green!20}\textcolor{black}{57} \\
11. grasper,retract,liver & \cellcolor{green!20}\textcolor{black}{1432} & \cellcolor{green!20}\textcolor{black}{1480} & \cellcolor{green!20}\textcolor{black}{18} & \cellcolor{red!20}\textcolor{black}{0} & 56. IA,aspirate,fluid & \cellcolor{green!20}\textcolor{black}{1894} & \cellcolor{green!20}\textcolor{black}{2593} & \cellcolor{green!20}\textcolor{black}{932} & \cellcolor{green!20}\textcolor{black}{1410} \\
12. grasper,retract,mesentery & \cellcolor{green!20}\textcolor{black}{577} & \cellcolor{green!20}\textcolor{black}{252} & \cellcolor{green!20}\textcolor{black}{302} & \cellcolor{green!20}\textcolor{black}{45} & 57. IA,irrigate,fluid & \cellcolor{green!20}\textcolor{black}{6} & \cellcolor{green!20}\textcolor{black}{27} & \cellcolor{green!20}\textcolor{black}{176} & \cellcolor{green!20}\textcolor{black}{174} \\
13. grasper,retract,sponge & \cellcolor{green!20}\textcolor{black}{121} & \cellcolor{red!20}\textcolor{black}{0} & \cellcolor{red!20}\textcolor{black}{0} & \cellcolor{red!20}\textcolor{black}{0} & 58. needle\_driver,grasp,thread & \cellcolor{green!20}\textcolor{black}{638} & \cellcolor{green!20}\textcolor{black}{733} & \cellcolor{green!20}\textcolor{black}{315} & \cellcolor{green!20}\textcolor{black}{96} \\
14. grasper,retract,colon & \cellcolor{green!20}\textcolor{black}{57} & \cellcolor{green!20}\textcolor{black}{27} & \cellcolor{green!20}\textcolor{black}{10} & \cellcolor{green!20}\textcolor{black}{4} & 59. needle\_driver,grasp,needle & \cellcolor{green!20}\textcolor{black}{1025} & \cellcolor{green!20}\textcolor{black}{814} & \cellcolor{green!20}\textcolor{black}{388} & \cellcolor{green!20}\textcolor{black}{63} \\
15. grasper,dissect,small\_bowel & \cellcolor{green!20}\textcolor{black}{532} & \cellcolor{green!20}\textcolor{black}{35} & \cellcolor{red!20}\textcolor{black}{0} & \cellcolor{red!20}\textcolor{black}{0} & 60. needle\_driver,retract,small\_bowel & \cellcolor{green!20}\textcolor{black}{114} & \cellcolor{green!20}\textcolor{black}{50} & \cellcolor{green!20}\textcolor{black}{31} & \cellcolor{green!20}\textcolor{black}{23} \\
16. grasper,dissect,stomach & \cellcolor{green!20}\textcolor{black}{2691} & \cellcolor{green!20}\textcolor{black}{566} & \cellcolor{green!20}\textcolor{black}{102} & \cellcolor{green!20}\textcolor{black}{246} & 61. needle\_driver,suture,small\_bowel & \cellcolor{green!20}\textcolor{black}{268} & \cellcolor{green!20}\textcolor{black}{873} & \cellcolor{green!20}\textcolor{black}{20} & \cellcolor{green!20}\textcolor{black}{37} \\
17. grasper,dissect,omentum & \cellcolor{green!20}\textcolor{black}{562} & \cellcolor{green!20}\textcolor{black}{400} & \cellcolor{green!20}\textcolor{black}{42} & \cellcolor{green!20}\textcolor{black}{830} & 62. needle\_driver,suture,mesentery & \cellcolor{green!20}\textcolor{black}{8897} & \cellcolor{red!20}\textcolor{black}{0} & \cellcolor{green!20}\textcolor{black}{908} & \cellcolor{red!20}\textcolor{black}{0} \\
18. grasper,dissect,mesentery & \cellcolor{green!20}\textcolor{black}{413} & \cellcolor{green!20}\textcolor{black}{190} & \cellcolor{green!20}\textcolor{black}{26} & \cellcolor{green!20}\textcolor{black}{167} & 63. needle\_driver,suture,SBS-A & \cellcolor{green!20}\textcolor{black}{12828} & \cellcolor{green!20}\textcolor{black}{11348} & \cellcolor{green!20}\textcolor{black}{3821} & \cellcolor{red!20}\textcolor{black}{0} \\
19. grasper,suture,small\_bowel & \cellcolor{green!20}\textcolor{black}{86} & \cellcolor{green!20}\textcolor{black}{498} & \cellcolor{green!20}\textcolor{black}{6} & \cellcolor{green!20}\textcolor{black}{528} & 64. needle\_driver,suture,SB-A & \cellcolor{green!20}\textcolor{black}{9057} & \cellcolor{green!20}\textcolor{black}{7506} & \cellcolor{green!20}\textcolor{black}{1655} & \cellcolor{green!20}\textcolor{black}{2034} \\
20. grasper,suture,mesentery & \cellcolor{green!20}\textcolor{black}{5285} & \cellcolor{red!20}\textcolor{black}{0} & \cellcolor{green!20}\textcolor{black}{375} & \cellcolor{red!20}\textcolor{black}{0} & 65. clipper,clip,small\_bowel & \cellcolor{red!20}\textcolor{black}{0} & \cellcolor{red!20}\textcolor{black}{0} & \cellcolor{red!20}\textcolor{black}{0} & \cellcolor{green!20}\textcolor{black}{110} \\
21. grasper,suture,SBS-A & \cellcolor{green!20}\textcolor{black}{9151} & \cellcolor{green!20}\textcolor{black}{7176} & \cellcolor{green!20}\textcolor{black}{1432} & \cellcolor{red!20}\textcolor{black}{0} & 66. clipper,clip,stomach & \cellcolor{green!20}\textcolor{black}{45} & \cellcolor{green!20}\textcolor{black}{41} & \cellcolor{green!20}\textcolor{black}{39} & \cellcolor{green!20}\textcolor{black}{312} \\
22. grasper,suture,SB-A & \cellcolor{green!20}\textcolor{black}{5723} & \cellcolor{green!20}\textcolor{black}{4416} & \cellcolor{green!20}\textcolor{black}{801} & \cellcolor{green!20}\textcolor{black}{1069} & 67. bipolar\_forceps,grasp,sponge & \cellcolor{green!20}\textcolor{black}{107} & \cellcolor{red!20}\textcolor{black}{0} & \cellcolor{red!20}\textcolor{black}{0} & \cellcolor{red!20}\textcolor{black}{0} \\
23. stapler,grasp,small\_bowel & \cellcolor{green!20}\textcolor{black}{2130} & \cellcolor{green!20}\textcolor{black}{2441} & \cellcolor{green!20}\textcolor{black}{287} & \cellcolor{green!20}\textcolor{black}{209} & 68. bipolar\_forceps,coagulate,stomach & \cellcolor{green!20}\textcolor{black}{1192} & \cellcolor{red!20}\textcolor{black}{0} & \cellcolor{red!20}\textcolor{black}{0} & \cellcolor{red!20}\textcolor{black}{0} \\
24. stapler,grasp,stomach & \cellcolor{green!20}\textcolor{black}{1780} & \cellcolor{green!20}\textcolor{black}{1144} & \cellcolor{green!20}\textcolor{black}{632} & \cellcolor{green!20}\textcolor{black}{727} & 69. suture\_passer,pass,thread & \cellcolor{green!20}\textcolor{black}{630} & \cellcolor{red!20}\textcolor{black}{0} & \cellcolor{green!20}\textcolor{black}{119} & \cellcolor{red!20}\textcolor{black}{0} \\
25. stapler,grasp,SBS-A & \cellcolor{green!20}\textcolor{black}{636} & \cellcolor{green!20}\textcolor{black}{143} & \cellcolor{red!20}\textcolor{black}{0} & \cellcolor{red!20}\textcolor{black}{0} & 70. swab\_forceps,retract,small\_bowel & \cellcolor{red!20}\textcolor{black}{0} & \cellcolor{green!20}\textcolor{black}{169} & \cellcolor{red!20}\textcolor{black}{0} & \cellcolor{red!20}\textcolor{black}{0} \\
26. stapler,grasp,SB-A & \cellcolor{green!20}\textcolor{black}{625} & \cellcolor{green!20}\textcolor{black}{319} & \cellcolor{green!20}\textcolor{black}{216} & \cellcolor{green!20}\textcolor{black}{211} & 71. swab\_forceps,retract,liver & \cellcolor{red!20}\textcolor{black}{0} & \cellcolor{green!20}\textcolor{black}{343} & \cellcolor{red!20}\textcolor{black}{0} & \cellcolor{red!20}\textcolor{black}{0} \\
27. stapler,retract,stomach & \cellcolor{red!20}\textcolor{black}{0} & \cellcolor{green!20}\textcolor{black}{97} & \cellcolor{red!20}\textcolor{black}{0} & \cellcolor{red!20}\textcolor{black}{0} & 72. swab\_forceps,clean,small\_bowel & \cellcolor{red!20}\textcolor{black}{0} & \cellcolor{green!20}\textcolor{black}{472} & \cellcolor{green!20}\textcolor{black}{23} & \cellcolor{red!20}\textcolor{black}{0} \\
28. stapler,staple,small\_bowel & \cellcolor{green!20}\textcolor{black}{167} & \cellcolor{green!20}\textcolor{black}{211} & \cellcolor{green!20}\textcolor{black}{42} & \cellcolor{green!20}\textcolor{black}{102} & 73. swab\_forceps,clean,stomach & \cellcolor{red!20}\textcolor{black}{0} & \cellcolor{green!20}\textcolor{black}{299} & \cellcolor{green!20}\textcolor{black}{111} & \cellcolor{green!20}\textcolor{black}{71} \\
29. stapler,staple,stomach & \cellcolor{green!20}\textcolor{black}{732} & \cellcolor{green!20}\textcolor{black}{958} & \cellcolor{green!20}\textcolor{black}{347} & \cellcolor{green!20}\textcolor{black}{624} & 74. gauze,clean,stomach & \cellcolor{green!20}\textcolor{black}{3425} & \cellcolor{red!20}\textcolor{black}{0} & \cellcolor{green!20}\textcolor{black}{604} & \cellcolor{red!20}\textcolor{black}{0} \\
30. stapler,staple,SBS-A & \cellcolor{green!20}\textcolor{black}{197} & \cellcolor{green!20}\textcolor{black}{171} & \cellcolor{red!20}\textcolor{black}{0} & \cellcolor{red!20}\textcolor{black}{0} & 75. grasper,null\_verb,null\_target & \cellcolor{green!20}\textcolor{black}{13825} & \cellcolor{green!20}\textcolor{black}{9022} & \cellcolor{green!20}\textcolor{black}{2551} & \cellcolor{green!20}\textcolor{black}{1998} \\
31. stapler,staple,SB-A & \cellcolor{green!20}\textcolor{black}{245} & \cellcolor{green!20}\textcolor{black}{270} & \cellcolor{green!20}\textcolor{black}{77} & \cellcolor{green!20}\textcolor{black}{99} & 76. stapler,null\_verb,null\_target & \cellcolor{green!20}\textcolor{black}{909} & \cellcolor{green!20}\textcolor{black}{930} & \cellcolor{green!20}\textcolor{black}{267} & \cellcolor{green!20}\textcolor{black}{103} \\
32. electric\_hook,dissect,small\_bowel & \cellcolor{green!20}\textcolor{black}{788} & \cellcolor{red!20}\textcolor{black}{0} & \cellcolor{green!20}\textcolor{black}{133} & \cellcolor{red!20}\textcolor{black}{0} & 77. electric\_hook,null\_verb,null\_target & \cellcolor{green!20}\textcolor{black}{466} & \cellcolor{green!20}\textcolor{black}{140} & \cellcolor{green!20}\textcolor{black}{102} & \cellcolor{red!20}\textcolor{black}{0} \\
33. electric\_hook,dissect,stomach & \cellcolor{green!20}\textcolor{black}{1080} & \cellcolor{red!20}\textcolor{black}{0} & \cellcolor{green!20}\textcolor{black}{137} & \cellcolor{red!20}\textcolor{black}{0} & 78. RAD,null\_verb,null\_target & \cellcolor{red!20}\textcolor{black}{0} & \cellcolor{green!20}\textcolor{black}{126} & \cellcolor{green!20}\textcolor{black}{39} & \cellcolor{green!20}\textcolor{black}{3} \\
34. electric\_hook,dissect,omentum & \cellcolor{green!20}\textcolor{black}{1283} & \cellcolor{red!20}\textcolor{black}{0} & \cellcolor{red!20}\textcolor{black}{0} & \cellcolor{red!20}\textcolor{black}{0} & 79. energy\_device,null\_verb,null\_target & \cellcolor{green!20}\textcolor{black}{586} & \cellcolor{green!20}\textcolor{black}{1222} & \cellcolor{green!20}\textcolor{black}{249} & \cellcolor{green!20}\textcolor{black}{137} \\
35. electric\_hook,coagulate,spleen & \cellcolor{red!20}\textcolor{black}{0} & \cellcolor{green!20}\textcolor{black}{272} & \cellcolor{red!20}\textcolor{black}{0} & \cellcolor{red!20}\textcolor{black}{0} & 80. IA,null\_verb,null\_target & \cellcolor{green!20}\textcolor{black}{44} & \cellcolor{green!20}\textcolor{black}{245} & \cellcolor{green!20}\textcolor{black}{68} & \cellcolor{green!20}\textcolor{black}{75} \\
36. RAD,grasp,stomach & \cellcolor{red!20}\textcolor{black}{0} & \cellcolor{green!20}\textcolor{black}{141} & \cellcolor{red!20}\textcolor{black}{0} & \cellcolor{green!20}\textcolor{black}{122} & 81. needle\_driver,null\_verb,null\_target & \cellcolor{green!20}\textcolor{black}{421} & \cellcolor{green!20}\textcolor{black}{95} & \cellcolor{green!20}\textcolor{black}{159} & \cellcolor{green!20}\textcolor{black}{2} \\
37. RAD,dissect,small\_bowel & \cellcolor{green!20}\textcolor{black}{21} & \cellcolor{green!20}\textcolor{black}{398} & \cellcolor{green!20}\textcolor{black}{18} & \cellcolor{green!20}\textcolor{black}{103} & 82. clipper,null\_verb,null\_target & \cellcolor{green!20}\textcolor{black}{29} & \cellcolor{green!20}\textcolor{black}{66} & \cellcolor{green!20}\textcolor{black}{21} & \cellcolor{green!20}\textcolor{black}{19} \\
38. RAD,dissect,stomach & \cellcolor{green!20}\textcolor{black}{6} & \cellcolor{green!20}\textcolor{black}{193} & \cellcolor{green!20}\textcolor{black}{21} & \cellcolor{green!20}\textcolor{black}{51} & 83. bipolar\_forceps,null\_verb,null\_target & \cellcolor{green!20}\textcolor{black}{339} & \cellcolor{red!20}\textcolor{black}{0} & \cellcolor{red!20}\textcolor{black}{0} & \cellcolor{red!20}\textcolor{black}{0} \\
39. energy\_device,grasp,omentum & \cellcolor{green!20}\textcolor{black}{64} & \cellcolor{green!20}\textcolor{black}{51} & \cellcolor{green!20}\textcolor{black}{17} & \cellcolor{red!20}\textcolor{black}{0} & 84. swab\_forceps,null\_verb,null\_target & \cellcolor{red!20}\textcolor{black}{0} & \cellcolor{green!20}\textcolor{black}{305} & \cellcolor{red!20}\textcolor{black}{0} & \cellcolor{red!20}\textcolor{black}{0} \\
40. energy\_device,retract,small\_bowel & \cellcolor{green!20}\textcolor{black}{7} & \cellcolor{green!20}\textcolor{black}{92} & \cellcolor{green!20}\textcolor{black}{6} & \cellcolor{red!20}\textcolor{black}{0} &  &  &  &  &  \\
41. energy\_device,retract,stomach & \cellcolor{green!20}\textcolor{black}{107} & \cellcolor{green!20}\textcolor{black}{367} & \cellcolor{green!20}\textcolor{black}{57} & \cellcolor{green!20}\textcolor{black}{51} &  &  &  &  &  \\
42. energy\_device,retract,omentum & \cellcolor{green!20}\textcolor{black}{295} & \cellcolor{green!20}\textcolor{black}{216} & \cellcolor{green!20}\textcolor{black}{14} & \cellcolor{green!20}\textcolor{black}{62} &  &  &  &  &  \\
43. energy\_device,dissect,small\_bowel & \cellcolor{red!20}\textcolor{black}{0} & \cellcolor{green!20}\textcolor{black}{415} & \cellcolor{green!20}\textcolor{black}{51} & \cellcolor{green!20}\textcolor{black}{199} &  &  &  &  &  \\
44. energy\_device,dissect,stomach & \cellcolor{green!20}\textcolor{black}{810} & \cellcolor{green!20}\textcolor{black}{1191} & \cellcolor{green!20}\textcolor{black}{351} & \cellcolor{green!20}\textcolor{black}{521} &  &  &  &  &  \\
\hline
\end{tabular}
\end{table*}

\newlength{\tripletbarwidth}
\newlength{\tripletfillwidth}
\settowidth{\tripletbarwidth}{\scriptsize 83. bipolar\_forceps,null\_verb,null\_target}
\addtolength{\tripletbarwidth}{0.45em}
\newcommand{\tripletcell}[2]{%
  \begingroup
  \setlength{\tripletfillwidth}{\dimexpr \tripletbarwidth * #2 / 25064\relax}%
  {\scriptsize
  \makebox[\tripletbarwidth][l]{%
    \rlap{%
      \textcolor{black!12}{\rule{\tripletbarwidth}{1.7ex}}%
      \hspace{-\tripletbarwidth}%
      \textcolor{black!28}{\rule{\tripletfillwidth}{1.7ex}}%
    }%
    \hspace{0.18em}\textcolor{black}{#1}%
  }%
  \hspace{0.35em}\textcolor{black!60}{#2}}%
  \endgroup
}


\begin{table*}[t]
\centering
\footnotesize
\setlength{\tabcolsep}{3.5pt}
\renewcommand{\arraystretch}{1.12}
\caption{Class-wise triplet mAP across 12 models on the hidden test set of the \emph{challenge} split (Part 1: IDs 0--53). Green, violet, and orange cells indicate the best, second-best, and third-best values in each row, respectively; red cells denote missing entries shown as 0; and row-wise maxima are bold. Model abbreviations: RDV=RDV, RiT=RiT, SNXL=SurgeNetXL, LFM=LemonFM, LFM-T=LemonFM-T, DinoV3=DinoV3, DinoV3-LT=DinoV3-L-T, SelfD-E=SelfD (Ens), TERL-E=TERL (Ens), CCM-E=CurConMix (Ens), SPRT-S=SPIRIT-S, SPRT-TS=SPIRIT-TS. Triplet abbreviations: SB-A=small bowel anastomosis, SBS-A=small bowel stomach anastomosis, RAD=right angle dissector, IA=irrigator aspirator. Each triplet cell includes a fixed-width bar proportional to class frequency, with the raw count shown at the end.}
\label{tab:triplet_model_map_hidden_part1}
\resizebox{\textwidth}{!}{
\begin{tabular}{p{0.34\textwidth}cccccccccccc}
\hline
Triplet & RDV & RiT & SNXL & LFM & LFM-T & DinoV3 & DinoV3-LT & SelfD-E & TERL-E & CCM-E & SPRT-S & SPRT-TS \\
\hline
\tripletcell{0. grasper,grasp,small\_bowel}{25064}  & \textcolor{black}{0.383} & \textcolor{black}{0.452} & \textcolor{black}{0.809} & \cellcolor{orange!30}\textcolor{black}{0.844} & \textcolor{black}{0.825} & \textcolor{black}{0.821} & \textcolor{black}{0.824} & \textcolor{black}{0.837} & \textcolor{black}{0.805} & \textcolor{black}{0.831} & \cellcolor{blue!20}\textcolor{black}{0.862} & \cellcolor{green!25}\textcolor{black}{\textbf{0.890}} \\
\tripletcell{1. grasper,grasp,stomach}{16368}  & \textcolor{black}{0.433} & \textcolor{black}{0.539} & \textcolor{black}{0.788} & \textcolor{black}{0.839} & \textcolor{black}{0.807} & \cellcolor{orange!30}\textcolor{black}{0.844} & \textcolor{black}{0.815} & \textcolor{black}{0.842} & \textcolor{black}{0.811} & \textcolor{black}{0.839} & \cellcolor{blue!20}\textcolor{black}{0.857} & \cellcolor{green!25}\textcolor{black}{\textbf{0.863}} \\
\tripletcell{2. grasper,grasp,thread}{19504}  & \textcolor{black}{0.360} & \textcolor{black}{0.446} & \textcolor{black}{0.669} & \textcolor{black}{0.701} & \textcolor{black}{0.699} & \textcolor{black}{0.749} & \textcolor{black}{0.749} & \cellcolor{orange!30}\textcolor{black}{0.781} & \textcolor{black}{0.705} & \textcolor{black}{0.756} & \cellcolor{blue!20}\textcolor{black}{0.786} & \cellcolor{green!25}\textcolor{black}{\textbf{0.842}} \\
\tripletcell{3. grasper,grasp,omentum}{9875}  & \textcolor{black}{0.320} & \textcolor{black}{0.392} & \textcolor{black}{0.631} & \cellcolor{orange!30}\textcolor{black}{0.650} & \textcolor{black}{0.636} & \textcolor{black}{0.637} & \textcolor{black}{0.603} & \textcolor{black}{0.646} & \textcolor{black}{0.623} & \textcolor{black}{0.647} & \cellcolor{blue!20}\textcolor{black}{0.688} & \cellcolor{green!25}\textcolor{black}{\textbf{0.692}} \\
\tripletcell{4. grasper,grasp,liver}{61}  & \textcolor{black}{0.010} & \textcolor{black}{0.027} & \textcolor{black}{0.015} & \cellcolor{orange!30}\textcolor{black}{0.044} & \textcolor{black}{0.032} & \textcolor{black}{0.015} & \textcolor{black}{0.035} & \textcolor{black}{0.010} & \textcolor{black}{0.013} & \textcolor{black}{0.013} & \cellcolor{green!25}\textcolor{black}{\textbf{0.155}} & \cellcolor{blue!20}\textcolor{black}{0.085} \\
\tripletcell{5. grasper,grasp,needle}{2144}  & \textcolor{black}{0.075} & \textcolor{black}{0.091} & \textcolor{black}{0.300} & \textcolor{black}{0.364} & \textcolor{black}{0.165} & \cellcolor{orange!30}\textcolor{black}{0.422} & \textcolor{black}{0.312} & \textcolor{black}{0.362} & \textcolor{black}{0.375} & \textcolor{black}{0.383} & \cellcolor{blue!20}\textcolor{black}{0.426} & \cellcolor{green!25}\textcolor{black}{\textbf{0.482}} \\
\tripletcell{6. grasper,grasp,mesentery}{2689}  & \textcolor{black}{0.129} & \textcolor{black}{0.158} & \textcolor{black}{0.187} & \textcolor{black}{0.232} & \textcolor{black}{0.217} & \textcolor{black}{0.250} & \textcolor{black}{0.226} & \cellcolor{blue!20}\textcolor{black}{0.276} & \textcolor{black}{0.217} & \cellcolor{orange!30}\textcolor{black}{0.263} & \cellcolor{orange!30}\textcolor{black}{0.263} & \cellcolor{green!25}\textcolor{black}{\textbf{0.289}} \\
\tripletcell{7. grasper,grasp,colon}{122}  & \textcolor{black}{0.010} & \textcolor{black}{0.011} & \textcolor{black}{0.050} & \cellcolor{blue!20}\textcolor{black}{0.055} & \textcolor{black}{0.034} & \textcolor{black}{0.008} & \textcolor{black}{0.016} & \textcolor{black}{0.039} & \textcolor{black}{0.030} & \cellcolor{orange!30}\textcolor{black}{0.051} & \cellcolor{green!25}\textcolor{black}{\textbf{0.116}} & \textcolor{black}{0.011} \\
\tripletcell{8. grasper,retract,small\_bowel}{2026}  & \textcolor{black}{0.051} & \textcolor{black}{0.053} & \textcolor{black}{0.100} & \textcolor{black}{0.107} & \textcolor{black}{0.085} & \textcolor{black}{0.118} & \textcolor{black}{0.119} & \cellcolor{orange!30}\textcolor{black}{0.140} & \textcolor{black}{0.089} & \textcolor{black}{0.123} & \cellcolor{green!25}\textcolor{black}{\textbf{0.159}} & \cellcolor{blue!20}\textcolor{black}{0.150} \\
\tripletcell{9. grasper,retract,stomach}{3361}  & \textcolor{black}{0.130} & \textcolor{black}{0.117} & \textcolor{black}{0.341} & \cellcolor{green!25}\textcolor{black}{\textbf{0.430}} & \textcolor{black}{0.341} & \textcolor{black}{0.357} & \textcolor{black}{0.336} & \cellcolor{blue!20}\textcolor{black}{0.421} & \textcolor{black}{0.255} & \textcolor{black}{0.398} & \textcolor{black}{0.314} & \cellcolor{orange!30}\textcolor{black}{0.411} \\
\tripletcell{10. grasper,retract,omentum}{1336}  & \textcolor{black}{0.045} & \textcolor{black}{0.069} & \cellcolor{orange!30}\textcolor{black}{0.108} & \textcolor{black}{0.069} & \textcolor{black}{0.082} & \cellcolor{green!25}\textcolor{black}{\textbf{0.126}} & \cellcolor{blue!20}\textcolor{black}{0.117} & \textcolor{black}{0.083} & \textcolor{black}{0.094} & \textcolor{black}{0.086} & \textcolor{black}{0.069} & \textcolor{black}{0.097} \\
\tripletcell{11. grasper,retract,liver}{191}  & \textcolor{black}{0.032} & \textcolor{black}{0.022} & \textcolor{black}{0.198} & \textcolor{black}{0.221} & \textcolor{black}{0.265} & \textcolor{black}{0.225} & \textcolor{black}{0.184} & \cellcolor{orange!30}\textcolor{black}{0.274} & \textcolor{black}{0.169} & \textcolor{black}{0.246} & \cellcolor{green!25}\textcolor{black}{\textbf{0.347}} & \cellcolor{blue!20}\textcolor{black}{0.314} \\
\tripletcell{12. grasper,retract,mesentery}{496}  & \textcolor{black}{0.089} & \textcolor{black}{0.069} & \cellcolor{orange!30}\textcolor{black}{0.110} & \textcolor{black}{0.105} & \textcolor{black}{0.105} & \cellcolor{blue!20}\textcolor{black}{0.117} & \textcolor{black}{0.067} & \textcolor{black}{0.088} & \cellcolor{green!25}\textcolor{black}{\textbf{0.133}} & \textcolor{black}{0.092} & \textcolor{black}{0.099} & \textcolor{black}{0.099} \\
\tripletcell{13. grasper,retract,sponge}{0}  & \cellcolor{red!20}\textcolor{black}{0} & \cellcolor{red!20}\textcolor{black}{0} & \cellcolor{red!20}\textcolor{black}{0} & \cellcolor{red!20}\textcolor{black}{0} & \cellcolor{red!20}\textcolor{black}{0} & \cellcolor{red!20}\textcolor{black}{0} & \cellcolor{red!20}\textcolor{black}{0} & \cellcolor{red!20}\textcolor{black}{0} & \cellcolor{red!20}\textcolor{black}{0} & \cellcolor{red!20}\textcolor{black}{0} & \cellcolor{red!20}\textcolor{black}{0} & \cellcolor{red!20}\textcolor{black}{0} \\
\tripletcell{14. grasper,retract,colon}{14}  & \textcolor{black}{0.044} & \textcolor{black}{0.010} & \textcolor{black}{0.006} & \textcolor{black}{0.029} & \textcolor{black}{0.042} & \cellcolor{orange!30}\textcolor{black}{0.045} & \cellcolor{green!25}\textcolor{black}{\textbf{0.096}} & \textcolor{black}{0.043} & \textcolor{black}{0.041} & \textcolor{black}{0.034} & \textcolor{black}{0.003} & \cellcolor{blue!20}\textcolor{black}{0.052} \\
\tripletcell{15. grasper,dissect,small\_bowel}{100}  & \textcolor{black}{0.028} & \textcolor{black}{0.074} & \textcolor{black}{0.434} & \textcolor{black}{0.388} & \textcolor{black}{0.262} & \textcolor{black}{0.411} & \textcolor{black}{0.302} & \cellcolor{orange!30}\textcolor{black}{0.464} & \textcolor{black}{0.399} & \cellcolor{blue!20}\textcolor{black}{0.476} & \textcolor{black}{0.388} & \cellcolor{green!25}\textcolor{black}{\textbf{0.557}} \\
\tripletcell{16. grasper,dissect,stomach}{829}  & \textcolor{black}{0.051} & \textcolor{black}{0.065} & \textcolor{black}{0.365} & \textcolor{black}{0.362} & \textcolor{black}{0.369} & \textcolor{black}{0.381} & \cellcolor{green!25}\textcolor{black}{\textbf{0.401}} & \textcolor{black}{0.359} & \textcolor{black}{0.320} & \textcolor{black}{0.370} & \cellcolor{orange!30}\textcolor{black}{0.383} & \cellcolor{blue!20}\textcolor{black}{0.387} \\
\tripletcell{17. grasper,dissect,omentum}{1537}  & \textcolor{black}{0.069} & \textcolor{black}{0.056} & \textcolor{black}{0.167} & \cellcolor{orange!30}\textcolor{black}{0.283} & \textcolor{black}{0.227} & \textcolor{black}{0.232} & \textcolor{black}{0.220} & \textcolor{black}{0.218} & \textcolor{black}{0.228} & \textcolor{black}{0.260} & \cellcolor{blue!20}\textcolor{black}{0.357} & \cellcolor{green!25}\textcolor{black}{\textbf{0.398}} \\
\tripletcell{18. grasper,dissect,mesentery}{283}  & \textcolor{black}{0.084} & \textcolor{black}{0.040} & \textcolor{black}{0.296} & \textcolor{black}{0.296} & \textcolor{black}{0.349} & \textcolor{black}{0.175} & \cellcolor{orange!30}\textcolor{black}{0.448} & \textcolor{black}{0.304} & \textcolor{black}{0.311} & \textcolor{black}{0.348} & \cellcolor{blue!20}\textcolor{black}{0.559} & \cellcolor{green!25}\textcolor{black}{\textbf{0.592}} \\
\tripletcell{19. grasper,suture,small\_bowel}{620}  & \textcolor{black}{0.040} & \textcolor{black}{0.050} & \textcolor{black}{0.056} & \cellcolor{orange!30}\textcolor{black}{0.194} & \textcolor{black}{0.066} & \textcolor{black}{0.044} & \textcolor{black}{0.072} & \textcolor{black}{0.040} & \textcolor{black}{0.031} & \textcolor{black}{0.028} & \cellcolor{green!25}\textcolor{black}{\textbf{0.392}} & \cellcolor{blue!20}\textcolor{black}{0.312} \\
\tripletcell{20. grasper,suture,mesentery}{1735}  & \textcolor{black}{0.222} & \textcolor{black}{0.486} & \textcolor{black}{0.668} & \textcolor{black}{0.638} & \textcolor{black}{0.697} & \cellcolor{green!25}\textcolor{black}{\textbf{0.794}} & \textcolor{black}{0.662} & \textcolor{black}{0.652} & \cellcolor{blue!20}\textcolor{black}{0.782} & \textcolor{black}{0.636} & \cellcolor{orange!30}\textcolor{black}{0.757} & \textcolor{black}{0.740} \\
\tripletcell{21. grasper,suture,SBS-A}{3073}  & \textcolor{black}{0.450} & \textcolor{black}{0.439} & \textcolor{black}{0.609} & \textcolor{black}{0.657} & \textcolor{black}{0.661} & \cellcolor{orange!30}\textcolor{black}{0.700} & \textcolor{black}{0.699} & \textcolor{black}{0.619} & \textcolor{black}{0.682} & \textcolor{black}{0.583} & \cellcolor{blue!20}\textcolor{black}{0.762} & \cellcolor{green!25}\textcolor{black}{\textbf{0.791}} \\
\tripletcell{22. grasper,suture,SB-A}{3839}  & \textcolor{black}{0.310} & \textcolor{black}{0.380} & \textcolor{black}{0.711} & \textcolor{black}{0.771} & \textcolor{black}{0.764} & \textcolor{black}{0.726} & \textcolor{black}{0.734} & \textcolor{black}{0.780} & \textcolor{black}{0.670} & \cellcolor{orange!30}\textcolor{black}{0.782} & \cellcolor{blue!20}\textcolor{black}{0.841} & \cellcolor{green!25}\textcolor{black}{\textbf{0.858}} \\
\tripletcell{23. stapler,grasp,small\_bowel}{1115}  & \textcolor{black}{0.051} & \textcolor{black}{0.073} & \textcolor{black}{0.381} & \textcolor{black}{0.467} & \cellcolor{orange!30}\textcolor{black}{0.476} & \textcolor{black}{0.394} & \cellcolor{green!25}\textcolor{black}{\textbf{0.506}} & \textcolor{black}{0.474} & \textcolor{black}{0.435} & \textcolor{black}{0.475} & \textcolor{black}{0.408} & \cellcolor{blue!20}\textcolor{black}{0.486} \\
\tripletcell{24. stapler,grasp,stomach}{1991}  & \textcolor{black}{0.104} & \textcolor{black}{0.176} & \textcolor{black}{0.591} & \cellcolor{blue!20}\textcolor{black}{0.698} & \textcolor{black}{0.660} & \textcolor{black}{0.649} & \textcolor{black}{0.675} & \textcolor{black}{0.656} & \textcolor{black}{0.666} & \cellcolor{orange!30}\textcolor{black}{0.680} & \textcolor{black}{0.678} & \cellcolor{green!25}\textcolor{black}{\textbf{0.703}} \\
\tripletcell{25. stapler,grasp,SBS-A}{229}  & \textcolor{black}{0.057} & \textcolor{black}{0.170} & \textcolor{black}{0.752} & \textcolor{black}{0.787} & \textcolor{black}{0.751} & \textcolor{black}{0.717} & \textcolor{black}{0.718} & \cellcolor{blue!20}\textcolor{black}{0.830} & \textcolor{black}{0.703} & \cellcolor{green!25}\textcolor{black}{\textbf{0.838}} & \cellcolor{orange!30}\textcolor{black}{0.804} & \textcolor{black}{0.803} \\
\tripletcell{26. stapler,grasp,SB-A}{684}  & \textcolor{black}{0.026} & \textcolor{black}{0.036} & \textcolor{black}{0.284} & \cellcolor{blue!20}\textcolor{black}{0.475} & \textcolor{black}{0.377} & \cellcolor{orange!30}\textcolor{black}{0.406} & \textcolor{black}{0.373} & \textcolor{black}{0.389} & \textcolor{black}{0.398} & \textcolor{black}{0.400} & \cellcolor{green!25}\textcolor{black}{\textbf{0.495}} & \textcolor{black}{0.392} \\
\tripletcell{27. stapler,retract,stomach}{0}  & \cellcolor{red!20}\textcolor{black}{0} & \cellcolor{red!20}\textcolor{black}{0} & \cellcolor{red!20}\textcolor{black}{0} & \cellcolor{red!20}\textcolor{black}{0} & \cellcolor{red!20}\textcolor{black}{0} & \cellcolor{red!20}\textcolor{black}{0} & \cellcolor{red!20}\textcolor{black}{0} & \cellcolor{red!20}\textcolor{black}{0} & \cellcolor{red!20}\textcolor{black}{0} & \cellcolor{red!20}\textcolor{black}{0} & \cellcolor{red!20}\textcolor{black}{0} & \cellcolor{red!20}\textcolor{black}{0} \\
\tripletcell{28. stapler,staple,small\_bowel}{210}  & \textcolor{black}{0.028} & \textcolor{black}{0.028} & \textcolor{black}{0.326} & \textcolor{black}{0.338} & \cellcolor{blue!20}\textcolor{black}{0.413} & \textcolor{black}{0.265} & \textcolor{black}{0.342} & \textcolor{black}{0.311} & \textcolor{black}{0.152} & \textcolor{black}{0.309} & \cellcolor{orange!30}\textcolor{black}{0.403} & \cellcolor{green!25}\textcolor{black}{\textbf{0.481}} \\
\tripletcell{29. stapler,staple,stomach}{1081}  & \textcolor{black}{0.049} & \textcolor{black}{0.118} & \textcolor{black}{0.332} & \textcolor{black}{0.431} & \cellcolor{orange!30}\textcolor{black}{0.530} & \textcolor{black}{0.464} & \cellcolor{blue!20}\textcolor{black}{0.532} & \textcolor{black}{0.446} & \textcolor{black}{0.344} & \textcolor{black}{0.422} & \textcolor{black}{0.524} & \cellcolor{green!25}\textcolor{black}{\textbf{0.539}} \\
\tripletcell{30. stapler,staple,SBS-A}{38}  & \textcolor{black}{0.007} & \textcolor{black}{0.013} & \textcolor{black}{0.392} & \textcolor{black}{0.263} & \cellcolor{orange!30}\textcolor{black}{0.566} & \textcolor{black}{0.412} & \textcolor{black}{0.331} & \cellcolor{blue!20}\textcolor{black}{0.574} & \textcolor{black}{0.442} & \textcolor{black}{0.523} & \cellcolor{green!25}\textcolor{black}{\textbf{0.669}} & \textcolor{black}{0.504} \\
\tripletcell{31. stapler,staple,SB-A}{239}  & \textcolor{black}{0.017} & \textcolor{black}{0.023} & \textcolor{black}{0.357} & \textcolor{black}{0.358} & \cellcolor{orange!30}\textcolor{black}{0.404} & \textcolor{black}{0.325} & \cellcolor{green!25}\textcolor{black}{\textbf{0.562}} & \textcolor{black}{0.376} & \textcolor{black}{0.236} & \textcolor{black}{0.403} & \cellcolor{blue!20}\textcolor{black}{0.429} & \cellcolor{green!25}\textcolor{black}{\textbf{0.562}} \\
\tripletcell{32. electric\_hook,dissect,small\_bowel}{383}  & \textcolor{black}{0.050} & \textcolor{black}{0.127} & \textcolor{black}{0.517} & \cellcolor{green!25}\textcolor{black}{\textbf{0.714}} & \textcolor{black}{0.547} & \cellcolor{blue!20}\textcolor{black}{0.632} & \textcolor{black}{0.535} & \cellcolor{orange!30}\textcolor{black}{0.602} & \textcolor{black}{0.508} & \textcolor{black}{0.501} & \textcolor{black}{0.553} & \textcolor{black}{0.580} \\
\tripletcell{33. electric\_hook,dissect,stomach}{298}  & \textcolor{black}{0.072} & \textcolor{black}{0.123} & \textcolor{black}{0.433} & \cellcolor{orange!30}\textcolor{black}{0.492} & \textcolor{black}{0.372} & \textcolor{black}{0.480} & \textcolor{black}{0.350} & \cellcolor{green!25}\textcolor{black}{\textbf{0.508}} & \textcolor{black}{0.405} & \cellcolor{blue!20}\textcolor{black}{0.507} & \textcolor{black}{0.384} & \textcolor{black}{0.352} \\
\tripletcell{34. electric\_hook,dissect,omentum}{573}  & \textcolor{black}{0.169} & \textcolor{black}{0.242} & \textcolor{black}{0.748} & \textcolor{black}{0.802} & \cellcolor{orange!30}\textcolor{black}{0.809} & \textcolor{black}{0.743} & \cellcolor{blue!20}\textcolor{black}{0.812} & \textcolor{black}{0.800} & \textcolor{black}{0.779} & \textcolor{black}{0.798} & \textcolor{black}{0.789} & \cellcolor{green!25}\textcolor{black}{\textbf{0.834}} \\
\tripletcell{35. electric\_hook,coagulate,spleen}{0}  & \cellcolor{red!20}\textcolor{black}{0} & \cellcolor{red!20}\textcolor{black}{0} & \cellcolor{red!20}\textcolor{black}{0} & \cellcolor{red!20}\textcolor{black}{0} & \cellcolor{red!20}\textcolor{black}{0} & \cellcolor{red!20}\textcolor{black}{0} & \cellcolor{red!20}\textcolor{black}{0} & \cellcolor{red!20}\textcolor{black}{0} & \cellcolor{red!20}\textcolor{black}{0} & \cellcolor{red!20}\textcolor{black}{0} & \cellcolor{red!20}\textcolor{black}{0} & \cellcolor{red!20}\textcolor{black}{0} \\
\tripletcell{36. RAD,grasp,stomach}{122}  & \textcolor{black}{0.024} & \textcolor{black}{0.039} & \textcolor{black}{0.082} & \textcolor{black}{0.301} & \cellcolor{blue!20}\textcolor{black}{0.589} & \textcolor{black}{0.076} & \textcolor{black}{0.343} & \textcolor{black}{0.036} & \textcolor{black}{0.021} & \textcolor{black}{0.199} & \cellcolor{orange!30}\textcolor{black}{0.544} & \cellcolor{green!25}\textcolor{black}{\textbf{0.716}} \\
\tripletcell{37. RAD,dissect,small\_bowel}{121}  & \textcolor{black}{0.007} & \textcolor{black}{0.026} & \textcolor{black}{0.191} & \cellcolor{orange!30}\textcolor{black}{0.329} & \textcolor{black}{0.113} & \textcolor{black}{0.235} & \textcolor{black}{0.098} & \textcolor{black}{0.176} & \textcolor{black}{0.097} & \textcolor{black}{0.110} & \cellcolor{green!25}\textcolor{black}{\textbf{0.572}} & \cellcolor{blue!20}\textcolor{black}{0.549} \\
\tripletcell{38. RAD,dissect,stomach}{72}  & \textcolor{black}{0.061} & \textcolor{black}{0.023} & \textcolor{black}{0.054} & \textcolor{black}{0.349} & \textcolor{black}{0.183} & \textcolor{black}{0.248} & \cellcolor{orange!30}\textcolor{black}{0.354} & \textcolor{black}{0.065} & \textcolor{black}{0.030} & \textcolor{black}{0.038} & \cellcolor{blue!20}\textcolor{black}{0.388} & \cellcolor{green!25}\textcolor{black}{\textbf{0.581}} \\
\tripletcell{39. energy\_device,grasp,omentum}{30}  & \textcolor{black}{0.100} & \textcolor{black}{0.028} & \textcolor{black}{0.028} & \cellcolor{blue!20}\textcolor{black}{0.249} & \textcolor{black}{0.140} & \textcolor{black}{0.097} & \textcolor{black}{0.134} & \cellcolor{green!25}\textcolor{black}{\textbf{0.264}} & \textcolor{black}{0.143} & \textcolor{black}{0.204} & \textcolor{black}{0.195} & \cellcolor{orange!30}\textcolor{black}{0.241} \\
\tripletcell{40. energy\_device,retract,small\_bowel}{13}  & \textcolor{black}{0.006} & \textcolor{black}{0.014} & \textcolor{black}{0.037} & \textcolor{black}{0.029} & \cellcolor{blue!20}\textcolor{black}{0.069} & \cellcolor{orange!30}\textcolor{black}{0.068} & \textcolor{black}{0.013} & \textcolor{black}{0.044} & \textcolor{black}{0.018} & \textcolor{black}{0.039} & \textcolor{black}{0.061} & \cellcolor{green!25}\textcolor{black}{\textbf{0.091}} \\
\tripletcell{41. energy\_device,retract,stomach}{113}  & \textcolor{black}{0.008} & \textcolor{black}{0.012} & \textcolor{black}{0.010} & \textcolor{black}{0.014} & \textcolor{black}{0.011} & \textcolor{black}{0.010} & \textcolor{black}{0.010} & \textcolor{black}{0.017} & \textcolor{black}{0.018} & \cellcolor{orange!30}\textcolor{black}{0.019} & \cellcolor{blue!20}\textcolor{black}{0.030} & \cellcolor{green!25}\textcolor{black}{\textbf{0.032}} \\
\tripletcell{42. energy\_device,retract,omentum}{97}  & \textcolor{black}{0.006} & \textcolor{black}{0.009} & \textcolor{black}{0.030} & \textcolor{black}{0.056} & \cellcolor{orange!30}\textcolor{black}{0.103} & \textcolor{black}{0.057} & \cellcolor{green!25}\textcolor{black}{\textbf{0.162}} & \textcolor{black}{0.031} & \textcolor{black}{0.017} & \textcolor{black}{0.047} & \textcolor{black}{0.052} & \cellcolor{blue!20}\textcolor{black}{0.135} \\
\tripletcell{43. energy\_device,dissect,small\_bowel}{250}  & \textcolor{black}{0.012} & \textcolor{black}{0.020} & \textcolor{black}{0.216} & \cellcolor{green!25}\textcolor{black}{\textbf{0.340}} & \cellcolor{blue!20}\textcolor{black}{0.322} & \textcolor{black}{0.280} & \cellcolor{orange!30}\textcolor{black}{0.293} & \textcolor{black}{0.138} & \textcolor{black}{0.079} & \textcolor{black}{0.171} & \textcolor{black}{0.227} & \textcolor{black}{0.272} \\
\tripletcell{44. energy\_device,dissect,stomach}{919}  & \textcolor{black}{0.070} & \textcolor{black}{0.094} & \textcolor{black}{0.328} & \textcolor{black}{0.333} & \cellcolor{orange!30}\textcolor{black}{0.360} & \textcolor{black}{0.345} & \cellcolor{green!25}\textcolor{black}{\textbf{0.407}} & \textcolor{black}{0.346} & \textcolor{black}{0.313} & \textcolor{black}{0.312} & \textcolor{black}{0.342} & \cellcolor{blue!20}\textcolor{black}{0.393} \\
\tripletcell{45. energy\_device,dissect,omentum}{1107}  & \textcolor{black}{0.034} & \textcolor{black}{0.109} & \textcolor{black}{0.235} & \textcolor{black}{0.241} & \cellcolor{orange!30}\textcolor{black}{0.261} & \textcolor{black}{0.221} & \textcolor{black}{0.244} & \textcolor{black}{0.240} & \textcolor{black}{0.182} & \textcolor{black}{0.192} & \cellcolor{blue!20}\textcolor{black}{0.310} & \cellcolor{green!25}\textcolor{black}{\textbf{0.350}} \\
\tripletcell{46. energy\_device,dissect,mesentery}{168}  & \textcolor{black}{0.024} & \textcolor{black}{0.049} & \textcolor{black}{0.228} & \textcolor{black}{0.336} & \textcolor{black}{0.324} & \textcolor{black}{0.221} & \textcolor{black}{0.278} & \cellcolor{orange!30}\textcolor{black}{0.353} & \textcolor{black}{0.166} & \textcolor{black}{0.250} & \cellcolor{green!25}\textcolor{black}{\textbf{0.576}} & \cellcolor{blue!20}\textcolor{black}{0.479} \\
\tripletcell{47. energy\_device,cut,small\_bowel}{192}  & \textcolor{black}{0.024} & \textcolor{black}{0.053} & \textcolor{black}{0.328} & \cellcolor{orange!30}\textcolor{black}{0.566} & \cellcolor{blue!20}\textcolor{black}{0.572} & \textcolor{black}{0.141} & \textcolor{black}{0.484} & \textcolor{black}{0.274} & \textcolor{black}{0.189} & \textcolor{black}{0.351} & \textcolor{black}{0.525} & \cellcolor{green!25}\textcolor{black}{\textbf{0.785}} \\
\tripletcell{48. energy\_device,cut,stomach}{188}  & \textcolor{black}{0.045} & \textcolor{black}{0.047} & \textcolor{black}{0.306} & \cellcolor{blue!20}\textcolor{black}{0.600} & \cellcolor{green!25}\textcolor{black}{\textbf{0.681}} & \textcolor{black}{0.333} & \textcolor{black}{0.339} & \textcolor{black}{0.276} & \textcolor{black}{0.121} & \textcolor{black}{0.256} & \textcolor{black}{0.427} & \cellcolor{orange!30}\textcolor{black}{0.584} \\
\tripletcell{49. energy\_device,cut,omentum}{2045}  & \textcolor{black}{0.229} & \textcolor{black}{0.264} & \textcolor{black}{0.502} & \cellcolor{orange!30}\textcolor{black}{0.572} & \cellcolor{blue!20}\textcolor{black}{0.580} & \textcolor{black}{0.510} & \textcolor{black}{0.504} & \textcolor{black}{0.477} & \textcolor{black}{0.451} & \textcolor{black}{0.496} & \textcolor{black}{0.556} & \cellcolor{green!25}\textcolor{black}{\textbf{0.614}} \\
\tripletcell{50. energy\_device,cut,mesentery}{439}  & \textcolor{black}{0.067} & \textcolor{black}{0.072} & \textcolor{black}{0.412} & \cellcolor{blue!20}\textcolor{black}{0.728} & \textcolor{black}{0.548} & \cellcolor{orange!30}\textcolor{black}{0.615} & \textcolor{black}{0.582} & \textcolor{black}{0.495} & \textcolor{black}{0.362} & \textcolor{black}{0.563} & \textcolor{black}{0.607} & \cellcolor{green!25}\textcolor{black}{\textbf{0.735}} \\
\tripletcell{51. energy\_device,cut,adhesion}{92}  & \textcolor{black}{0.047} & \textcolor{black}{0.039} & \textcolor{black}{0.108} & \textcolor{black}{0.421} & \textcolor{black}{0.365} & \textcolor{black}{0.377} & \textcolor{black}{0.123} & \cellcolor{green!25}\textcolor{black}{\textbf{0.594}} & \cellcolor{blue!20}\textcolor{black}{0.533} & \cellcolor{orange!30}\textcolor{black}{0.465} & \textcolor{black}{0.181} & \textcolor{black}{0.357} \\
\tripletcell{52. energy\_device,coagulate,stomach}{91}  & \textcolor{black}{0.007} & \textcolor{black}{0.015} & \textcolor{black}{0.102} & \textcolor{black}{0.114} & \cellcolor{blue!20}\textcolor{black}{0.158} & \textcolor{black}{0.071} & \cellcolor{green!25}\textcolor{black}{\textbf{0.166}} & \textcolor{black}{0.074} & \textcolor{black}{0.045} & \textcolor{black}{0.049} & \cellcolor{orange!30}\textcolor{black}{0.129} & \textcolor{black}{0.064} \\
\tripletcell{53. energy\_device,coagulate,omentum}{267}  & \textcolor{black}{0.017} & \textcolor{black}{0.023} & \textcolor{black}{0.184} & \textcolor{black}{0.146} & \cellcolor{green!25}\textcolor{black}{\textbf{0.217}} & \textcolor{black}{0.119} & \textcolor{black}{0.157} & \cellcolor{orange!30}\textcolor{black}{0.189} & \cellcolor{blue!20}\textcolor{black}{0.190} & \textcolor{black}{0.148} & \textcolor{black}{0.074} & \textcolor{black}{0.145} \\
\hline
\end{tabular}
}
\end{table*}

\begin{table*}[!htbp]
\centering
\footnotesize
\setlength{\tabcolsep}{3.5pt}
\renewcommand{\arraystretch}{1.12}
\caption{Class-wise triplet mAP across 12 models on the hidden test set of the \emph{challenge} split (Part 1: IDs 0--53). Green, violet, and orange cells indicate the best, second-best, and third-best values in each row, respectively; red cells denote missing entries shown as 0; and row-wise maxima are bold. Model abbreviations: RDV=RDV, RiT=RiT, SNXL=SurgeNetXL, LFM=LemonFM, LFM-T=LemonFM-T, DinoV3=DinoV3, DinoV3-LT=DinoV3-L-T, SelfD-E=SelfD (Ens), TERL-E=TERL (Ens), CCM-E=CurConMix (Ens), SPRT-S=SPIRIT-S, SPRT-TS=SPIRIT-TS. Triplet abbreviations: SB-A=small bowel anastomosis, SBS-A=small bowel stomach anastomosis, RAD=right angle dissector, IA=irrigator aspirator. Each triplet cell includes a fixed-width bar proportional to class frequency, with the raw count shown at the end.}
\label{tab:triplet_model_map_hidden_part2}
\resizebox{\textwidth}{!}{
\begin{tabular}{p{0.34\textwidth}cccccccccccc}
\hline
Triplet & RDV & RiT & SNXL & LFM & LFM-T & DinoV3 & DinoV3-LT & SelfD-E & TERL-E & CCM-E & SPRT-S & SPRT-TS \\
\hline
\tripletcell{54. IA,retract,small\_bowel}{169}  & \textcolor{black}{0.013} & \textcolor{black}{0.023} & \textcolor{black}{0.041} & \textcolor{black}{0.051} & \cellcolor{orange!30}\textcolor{black}{0.082} & \textcolor{black}{0.029} & \cellcolor{green!25}\textcolor{black}{\textbf{0.157}} & \textcolor{black}{0.068} & \textcolor{black}{0.030} & \cellcolor{blue!20}\textcolor{black}{0.095} & \textcolor{black}{0.046} & \textcolor{black}{0.058} \\
\tripletcell{55. IA,retract,stomach}{76}  & \textcolor{black}{0.010} & \textcolor{black}{0.009} & \textcolor{black}{0.031} & \textcolor{black}{0.039} & \cellcolor{orange!30}\textcolor{black}{0.118} & \textcolor{black}{0.023} & \textcolor{black}{0.030} & \textcolor{black}{0.016} & \textcolor{black}{0.014} & \textcolor{black}{0.014} & \cellcolor{green!25}\textcolor{black}{\textbf{0.159}} & \cellcolor{blue!20}\textcolor{black}{0.127} \\
\tripletcell{56. IA,aspirate,fluid}{2944}  & \textcolor{black}{0.089} & \textcolor{black}{0.145} & \textcolor{black}{0.432} & \textcolor{black}{0.520} & \textcolor{black}{0.467} & \textcolor{black}{0.392} & \cellcolor{orange!30}\textcolor{black}{0.523} & \textcolor{black}{0.409} & \textcolor{black}{0.355} & \textcolor{black}{0.414} & \cellcolor{blue!20}\textcolor{black}{0.558} & \cellcolor{green!25}\textcolor{black}{\textbf{0.660}} \\
\tripletcell{57. IA,irrigate,fluid}{356}  & \textcolor{black}{0.020} & \textcolor{black}{0.014} & \textcolor{black}{0.054} & \textcolor{black}{0.085} & \cellcolor{green!25}\textcolor{black}{\textbf{0.124}} & \textcolor{black}{0.024} & \textcolor{black}{0.050} & \textcolor{black}{0.039} & \textcolor{black}{0.042} & \textcolor{black}{0.057} & \cellcolor{blue!20}\textcolor{black}{0.115} & \cellcolor{orange!30}\textcolor{black}{0.104} \\
\tripletcell{58. needle\_driver,grasp,thread}{596}  & \textcolor{black}{0.023} & \textcolor{black}{0.025} & \textcolor{black}{0.037} & \textcolor{black}{0.066} & \textcolor{black}{0.058} & \textcolor{black}{0.068} & \cellcolor{orange!30}\textcolor{black}{0.103} & \textcolor{black}{0.081} & \textcolor{black}{0.042} & \textcolor{black}{0.077} & \cellcolor{blue!20}\textcolor{black}{0.136} & \cellcolor{green!25}\textcolor{black}{\textbf{0.142}} \\
\tripletcell{59. needle\_driver,grasp,needle}{664}  & \textcolor{black}{0.023} & \textcolor{black}{0.023} & \textcolor{black}{0.050} & \textcolor{black}{0.064} & \textcolor{black}{0.045} & \textcolor{black}{0.080} & \textcolor{black}{0.097} & \cellcolor{blue!20}\textcolor{black}{0.164} & \textcolor{black}{0.110} & \cellcolor{orange!30}\textcolor{black}{0.147} & \textcolor{black}{0.116} & \cellcolor{green!25}\textcolor{black}{\textbf{0.190}} \\
\tripletcell{60. needle\_driver,retract,small\_bowel}{116}  & \textcolor{black}{0.005} & \textcolor{black}{0.004} & \textcolor{black}{0.010} & \cellcolor{orange!30}\textcolor{black}{0.018} & \textcolor{black}{0.015} & \textcolor{black}{0.010} & \textcolor{black}{0.017} & \cellcolor{blue!20}\textcolor{black}{0.028} & \textcolor{black}{0.006} & \textcolor{black}{0.011} & \cellcolor{green!25}\textcolor{black}{\textbf{0.029}} & \textcolor{black}{0.014} \\
\tripletcell{61. needle\_driver,suture,small\_bowel}{236}  & \textcolor{black}{0.039} & \textcolor{black}{0.022} & \textcolor{black}{0.050} & \textcolor{black}{0.073} & \textcolor{black}{0.106} & \cellcolor{orange!30}\textcolor{black}{0.153} & \textcolor{black}{0.074} & \textcolor{black}{0.038} & \textcolor{black}{0.073} & \textcolor{black}{0.042} & \cellcolor{green!25}\textcolor{black}{\textbf{0.220}} & \cellcolor{blue!20}\textcolor{black}{0.154} \\
\tripletcell{62. needle\_driver,suture,mesentery}{3202}  & \textcolor{black}{0.300} & \textcolor{black}{0.452} & \textcolor{black}{0.786} & \textcolor{black}{0.782} & \cellcolor{orange!30}\textcolor{black}{0.877} & \textcolor{black}{0.842} & \textcolor{black}{0.844} & \textcolor{black}{0.811} & \textcolor{black}{0.813} & \textcolor{black}{0.787} & \cellcolor{blue!20}\textcolor{black}{0.890} & \cellcolor{green!25}\textcolor{black}{\textbf{0.921}} \\
\tripletcell{63. needle\_driver,suture,SBS-A}{6620}  & \textcolor{black}{0.416} & \textcolor{black}{0.617} & \textcolor{black}{0.786} & \textcolor{black}{0.824} & \cellcolor{green!25}\textcolor{black}{\textbf{0.867}} & \textcolor{black}{0.833} & \textcolor{black}{0.849} & \textcolor{black}{0.810} & \textcolor{black}{0.779} & \textcolor{black}{0.797} & \cellcolor{orange!30}\textcolor{black}{0.852} & \cellcolor{blue!20}\textcolor{black}{0.860} \\
\tripletcell{64. needle\_driver,suture,SB-A}{6770}  & \textcolor{black}{0.262} & \textcolor{black}{0.316} & \textcolor{black}{0.795} & \textcolor{black}{0.858} & \textcolor{black}{0.861} & \textcolor{black}{0.855} & \cellcolor{blue!20}\textcolor{black}{0.880} & \textcolor{black}{0.843} & \textcolor{black}{0.801} & \textcolor{black}{0.823} & \cellcolor{orange!30}\textcolor{black}{0.871} & \cellcolor{green!25}\textcolor{black}{\textbf{0.912}} \\
\tripletcell{65. clipper,clip,small\_bowel}{110}  & \textcolor{black}{0.006} & \textcolor{black}{0.025} & \textcolor{black}{0.015} & \cellcolor{orange!30}\textcolor{black}{0.037} & \textcolor{black}{0.013} & \textcolor{black}{0.005} & \textcolor{black}{0.030} & \textcolor{black}{0.025} & \textcolor{black}{0.009} & \textcolor{black}{0.004} & \cellcolor{blue!20}\textcolor{black}{0.210} & \cellcolor{green!25}\textcolor{black}{\textbf{0.350}} \\
\tripletcell{66. clipper,clip,stomach}{351}  & \textcolor{black}{0.060} & \textcolor{black}{0.039} & \textcolor{black}{0.019} & \textcolor{black}{0.048} & \textcolor{black}{0.047} & \textcolor{black}{0.048} & \textcolor{black}{0.051} & \textcolor{black}{0.053} & \textcolor{black}{0.118} & \cellcolor{orange!30}\textcolor{black}{0.164} & \cellcolor{blue!20}\textcolor{black}{0.211} & \cellcolor{green!25}\textcolor{black}{\textbf{0.234}} \\
\tripletcell{67. bipolar\_forceps,grasp,sponge}{11}  & \textcolor{black}{0.003} & \textcolor{black}{0.002} & \textcolor{black}{0.003} & \textcolor{black}{0.003} & \textcolor{black}{0.003} & \textcolor{black}{0.003} & \cellcolor{orange!30}\textcolor{black}{0.014} & \textcolor{black}{0.003} & \textcolor{black}{0.002} & \textcolor{black}{0.002} & \cellcolor{green!25}\textcolor{black}{\textbf{0.156}} & \cellcolor{blue!20}\textcolor{black}{0.033} \\
\tripletcell{68. bipolar\_forceps,coagulate,stomach}{279}  & \textcolor{black}{0.071} & \textcolor{black}{0.449} & \textcolor{black}{0.569} & \textcolor{black}{0.683} & \textcolor{black}{0.677} & \cellcolor{blue!20}\textcolor{black}{0.729} & \textcolor{black}{0.636} & \textcolor{black}{0.701} & \textcolor{black}{0.689} & \textcolor{black}{0.708} & \cellcolor{orange!30}\textcolor{black}{0.719} & \cellcolor{green!25}\textcolor{black}{\textbf{0.781}} \\
\tripletcell{69. suture\_passer,pass,thread}{310}  & \textcolor{black}{0.141} & \textcolor{black}{0.309} & \textcolor{black}{0.607} & \cellcolor{blue!20}\textcolor{black}{0.736} & \textcolor{black}{0.591} & \textcolor{black}{0.645} & \textcolor{black}{0.620} & \cellcolor{orange!30}\textcolor{black}{0.678} & \textcolor{black}{0.604} & \textcolor{black}{0.551} & \textcolor{black}{0.646} & \cellcolor{green!25}\textcolor{black}{\textbf{0.739}} \\
\tripletcell{70. swab\_forceps,retract,small\_bowel}{0}  & \cellcolor{red!20}\textcolor{black}{0} & \cellcolor{red!20}\textcolor{black}{0} & \cellcolor{red!20}\textcolor{black}{0} & \cellcolor{red!20}\textcolor{black}{0} & \cellcolor{red!20}\textcolor{black}{0} & \cellcolor{red!20}\textcolor{black}{0} & \cellcolor{red!20}\textcolor{black}{0} & \cellcolor{red!20}\textcolor{black}{0} & \cellcolor{red!20}\textcolor{black}{0} & \cellcolor{red!20}\textcolor{black}{0} & \cellcolor{red!20}\textcolor{black}{0} & \cellcolor{red!20}\textcolor{black}{0} \\
\tripletcell{71. swab\_forceps,retract,liver}{0}  & \cellcolor{red!20}\textcolor{black}{0} & \cellcolor{red!20}\textcolor{black}{0} & \cellcolor{red!20}\textcolor{black}{0} & \cellcolor{red!20}\textcolor{black}{0} & \cellcolor{red!20}\textcolor{black}{0} & \cellcolor{red!20}\textcolor{black}{0} & \cellcolor{red!20}\textcolor{black}{0} & \cellcolor{red!20}\textcolor{black}{0} & \cellcolor{red!20}\textcolor{black}{0} & \cellcolor{red!20}\textcolor{black}{0} & \cellcolor{red!20}\textcolor{black}{0} & \cellcolor{red!20}\textcolor{black}{0} \\
\tripletcell{72. swab\_forceps,clean,small\_bowel}{23}  & \textcolor{black}{0.006} & \textcolor{black}{0.040} & \textcolor{black}{0.555} & \cellcolor{orange!30}\textcolor{black}{0.831} & \textcolor{black}{0.505} & \textcolor{black}{0.815} & \textcolor{black}{0.333} & \textcolor{black}{0.324} & \textcolor{black}{0.435} & \textcolor{black}{0.478} & \cellcolor{blue!20}\textcolor{black}{0.868} & \cellcolor{green!25}\textcolor{black}{\textbf{0.911}} \\
\tripletcell{73. swab\_forceps,clean,stomach}{182}  & \textcolor{black}{0.016} & \textcolor{black}{0.012} & \textcolor{black}{0.315} & \textcolor{black}{0.546} & \textcolor{black}{0.550} & \textcolor{black}{0.544} & \cellcolor{blue!20}\textcolor{black}{0.642} & \textcolor{black}{0.340} & \textcolor{black}{0.301} & \textcolor{black}{0.471} & \cellcolor{orange!30}\textcolor{black}{0.558} & \cellcolor{green!25}\textcolor{black}{\textbf{0.715}} \\
\tripletcell{74. gauze,clean,stomach}{2732}  & \textcolor{black}{0.273} & \cellcolor{blue!20}\textcolor{black}{0.306} & \cellcolor{orange!30}\textcolor{black}{0.304} & \cellcolor{green!25}\textcolor{black}{\textbf{0.333}} & \cellcolor{green!25}\textcolor{black}{\textbf{0.333}} & \textcolor{black}{0.303} & \cellcolor{blue!20}\textcolor{black}{0.306} & \textcolor{black}{0.260} & \textcolor{black}{0.290} & \textcolor{black}{0.269} & \textcolor{black}{0.282} & \textcolor{black}{0.269} \\
\tripletcell{75. grasper,null\_verb,null\_target}{8399}  & \textcolor{black}{0.132} & \textcolor{black}{0.130} & \textcolor{black}{0.269} & \textcolor{black}{0.296} & \textcolor{black}{0.284} & \textcolor{black}{0.315} & \textcolor{black}{0.326} & \cellcolor{blue!20}\textcolor{black}{0.368} & \textcolor{black}{0.325} & \textcolor{black}{0.361} & \cellcolor{orange!30}\textcolor{black}{0.366} & \cellcolor{green!25}\textcolor{black}{\textbf{0.445}} \\
\tripletcell{76. stapler,null\_verb,null\_target}{641}  & \textcolor{black}{0.052} & \textcolor{black}{0.071} & \textcolor{black}{0.319} & \textcolor{black}{0.275} & \textcolor{black}{0.179} & \cellcolor{orange!30}\textcolor{black}{0.432} & \textcolor{black}{0.311} & \textcolor{black}{0.396} & \textcolor{black}{0.352} & \textcolor{black}{0.418} & \cellcolor{blue!20}\textcolor{black}{0.493} & \cellcolor{green!25}\textcolor{black}{\textbf{0.526}} \\
\tripletcell{77. electric\_hook,null\_verb,null\_target}{169}  & \textcolor{black}{0.009} & \textcolor{black}{0.015} & \textcolor{black}{0.071} & \textcolor{black}{0.071} & \textcolor{black}{0.054} & \textcolor{black}{0.119} & \textcolor{black}{0.099} & \cellcolor{orange!30}\textcolor{black}{0.123} & \textcolor{black}{0.060} & \textcolor{black}{0.091} & \cellcolor{green!25}\textcolor{black}{\textbf{0.148}} & \cellcolor{blue!20}\textcolor{black}{0.134} \\
\tripletcell{78. RAD,null\_verb,null\_target}{42}  & \textcolor{black}{0.003} & \textcolor{black}{0.003} & \textcolor{black}{0.015} & \cellcolor{orange!30}\textcolor{black}{0.040} & \textcolor{black}{0.008} & \cellcolor{blue!20}\textcolor{black}{0.148} & \cellcolor{green!25}\textcolor{black}{\textbf{0.290}} & \textcolor{black}{0.006} & \textcolor{black}{0.009} & \textcolor{black}{0.020} & \textcolor{black}{0.018} & \textcolor{black}{0.038} \\
\tripletcell{79. energy\_device,null\_verb,null\_target}{477}  & \textcolor{black}{0.027} & \textcolor{black}{0.031} & \textcolor{black}{0.172} & \cellcolor{blue!20}\textcolor{black}{0.282} & \textcolor{black}{0.140} & \textcolor{black}{0.221} & \textcolor{black}{0.237} & \textcolor{black}{0.254} & \textcolor{black}{0.189} & \cellcolor{orange!30}\textcolor{black}{0.257} & \cellcolor{green!25}\textcolor{black}{\textbf{0.367}} & \cellcolor{green!25}\textcolor{black}{\textbf{0.367}} \\
\tripletcell{80. IA,null\_verb,null\_target}{159}  & \textcolor{black}{0.004} & \textcolor{black}{0.004} & \textcolor{black}{0.025} & \cellcolor{orange!30}\textcolor{black}{0.079} & \textcolor{black}{0.061} & \textcolor{black}{0.022} & \textcolor{black}{0.075} & \textcolor{black}{0.037} & \textcolor{black}{0.034} & \textcolor{black}{0.029} & \cellcolor{green!25}\textcolor{black}{\textbf{0.162}} & \cellcolor{blue!20}\textcolor{black}{0.139} \\
\tripletcell{81. needle\_driver,null\_verb,null\_target}{392}  & \textcolor{black}{0.014} & \textcolor{black}{0.009} & \textcolor{black}{0.035} & \textcolor{black}{0.049} & \textcolor{black}{0.029} & \textcolor{black}{0.027} & \cellcolor{blue!20}\textcolor{black}{0.073} & \textcolor{black}{0.031} & \textcolor{black}{0.036} & \textcolor{black}{0.042} & \cellcolor{orange!30}\textcolor{black}{0.058} & \cellcolor{green!25}\textcolor{black}{\textbf{0.089}} \\
\tripletcell{82. clipper,null\_verb,null\_target}{58}  & \textcolor{black}{0.003} & \textcolor{black}{0.002} & \textcolor{black}{0.003} & \textcolor{black}{0.008} & \textcolor{black}{0.003} & \textcolor{black}{0.003} & \textcolor{black}{0.005} & \textcolor{black}{0.004} & \textcolor{black}{0.003} & \cellcolor{orange!30}\textcolor{black}{0.037} & \cellcolor{green!25}\textcolor{black}{\textbf{0.054}} & \cellcolor{blue!20}\textcolor{black}{0.042} \\
\tripletcell{83. bipolar\_forceps,null\_verb,null\_target}{116}  & \textcolor{black}{0.008} & \textcolor{black}{0.011} & \textcolor{black}{0.143} & \textcolor{black}{0.099} & \textcolor{black}{0.074} & \textcolor{black}{0.108} & \textcolor{black}{0.097} & \textcolor{black}{0.162} & \textcolor{black}{0.076} & \cellcolor{blue!20}\textcolor{black}{0.176} & \cellcolor{green!25}\textcolor{black}{\textbf{0.202}} & \cellcolor{orange!30}\textcolor{black}{0.170} \\
\tripletcell{84. swab\_forceps,null\_verb,null\_target}{0}  & \cellcolor{red!20}\textcolor{black}{0} & \cellcolor{red!20}\textcolor{black}{0} & \cellcolor{red!20}\textcolor{black}{0} & \cellcolor{red!20}\textcolor{black}{0} & \cellcolor{red!20}\textcolor{black}{0} & \cellcolor{red!20}\textcolor{black}{0} & \cellcolor{red!20}\textcolor{black}{0} & \cellcolor{red!20}\textcolor{black}{0} & \cellcolor{red!20}\textcolor{black}{0} & \cellcolor{red!20}\textcolor{black}{0} & \cellcolor{red!20}\textcolor{black}{0} & \cellcolor{red!20}\textcolor{black}{0} \\
\hline
\end{tabular}
}
\end{table*}

\begin{table*}[!htbp]
\centering
\setlength{\tabcolsep}{14pt}
\caption{\label{results_crossval_cholecT45}Cross-validation results for surgical action triplet recognition on CholecT45. Values are reported as mean $\pm$ standard deviation AP (\%). SPIRIT-TS achieves the highest triplet-level performance ($AP_{IVT}$), indicating that the proposed relational formulation remains effective beyond MultiBypass-4C-T40.}
\resizebox{\textwidth}{!}{
\begin{tabular}{lcccccc}
\toprule
\multirow{2}{*}{\textbf{Method}} &
\multicolumn{3}{c}{\textbf{Component Detection}} &
\multicolumn{3}{c}{\textbf{Triplet Association}} \\
\cmidrule(lr){2-4} \cmidrule(lr){5-7}
& $AP_I$ & $AP_V$ & $AP_T$ & $AP_{IV}$ & $AP_{IT}$ & $AP_{IVT}$ \\
\midrule
RDV~\citep{nwoye2022rendezvous} & 89.3 $\pm$ 2.1 & 62.0 $\pm$ 1.3 & 40.0 $\pm$ 1.4 & 34.0 $\pm$ 3.3 & 30.8 $\pm$ 2.1 & 29.4 $\pm$ 2.8 \\
RiT~\citep{sharma2023rendezvous} & 88.6 $\pm$ 2.6 & 64.0 $\pm$ 2.5 & 43.4 $\pm$ 1.4 & 38.3 $\pm$ 3.5 & 36.9 $\pm$ 1.0 & 29.7 $\pm$ 2.6 \\
TDN~\citep{chen2023surgical} & 91.2 $\pm$ 1.9 & 65.3 $\pm$ 2.8 & 43.7 $\pm$ 1.6 & - & - & 33.8 $\pm$ 2.5 \\
MT4MTL-KD~\citep{gui2023mt4mtl} & 93.1 $\pm$ 2.1 & 71.8 $\pm$ 3.4 & 48.8 $\pm$ 3.8 & 44.9 $\pm$ 2.4 & 43.1 $\pm$ 2.0 & 37.1 $\pm$ 0.5 \\
SelfD~\citep{yamlahi2023self} & 90.3 $\pm$ 2.3 & 67.4 $\pm$ 1.5 & 47.9 $\pm$ 1.8 & 43.7 $\pm$ 4.1 & 42.9 $\pm$ 1.6 & 37.1 $\pm$ 1.9 \\
TERL-T~\citep{gui2024tail} & 93.5 $\pm$ 1.5 & 71.4 $\pm$ 2.2 & 47.2 $\pm$ 2.6 & 44.7 $\pm$ 3.8 & 42.0 $\pm$ 2.4 & 35.7 $\pm$ 1.6 \\
TERL-B (224)~\citep{gui2024tail} & 93.9 $\pm$ 2.0 & 70.8 $\pm$ 2.3 & 49.4 $\pm$ 4.7 & 43.9 $\pm$ 3.4 & 43.6 $\pm$ 2.6 & 35.6 $\pm$ 1.4 \\
TERL-B (384)~\citep{gui2024tail} & 94.1 $\pm$ 2.3 & 73.0 $\pm$ 1.4 & 51.1 $\pm$ 3.8 & 46.5 $\pm$ 4.9 & 44.9 $\pm$ 1.8 & 37.7 $\pm$ 1.5 \\
TERL (Ens)~\citep{gui2024tail} & 94.6 $\pm$ 1.9 & 73.5 $\pm$ 1.9 & 50.8 $\pm$ 3.3 & 47.3 $\pm$ 4.1 & 45.3 $\pm$ 1.9 & 38.5 $\pm$ 1.1 \\
CurConMix-T~\citep{jeon2025curconmix} & 90.4 $\pm$ 2.1 & 67.8 $\pm$ 1.8 & 48.3 $\pm$ 3.4 & 43.3 $\pm$ 2.9 & 43.3 $\pm$ 1.8 & 37.7 $\pm$ 2.1 \\
CurConMix-B~\citep{jeon2025curconmix} & 90.4 $\pm$ 3.0 & 68.2 $\pm$ 1.5 & 49.7 $\pm$ 2.5 & 44.8 $\pm$ 5.4 & 45.3 $\pm$ 2.4 & 38.8 $\pm$ 2.8 \\
CurConMix-B (384)~\citep{jeon2025curconmix} & 90.9 $\pm$ 2.0 & 68.3 $\pm$ 1.3 & 49.8 $\pm$ 3.2 & 45.2 $\pm$ 4.2 & 45.1 $\pm$ 1.1 & 39.1 $\pm$ 2.0 \\
CurConMix-Ens~\citep{jeon2025curconmix} & 91.7 $\pm$ 2.2 & 69.5 $\pm$ 0.4 & 51.3 $\pm$ 2.9 & 46.3 $\pm$ 5.0 & 47.1 $\pm$ 1.6 & 40.7 $\pm$ 2.1 \\
MEJO-B~\citep{zhang2025mejo} & 93.9 $\pm$ 2.2 & 72.9 $\pm$ 1.8 & 51.6 $\pm$ 5.7 & 47.8 $\pm$ 6.4 & 46.3 $\pm$ 1.8 & 41.2 $\pm$ 2.6 \\
\midrule
SPIRIT-S (\textit{Ours}) & 94.7 $\pm$ 1.6 & 70.1 $\pm$ 1.6 & 47.2 $\pm$ 4.2 & 48.1 $\pm$ 4.0 & 47.3 $\pm$ 3.1 & 40.1 $\pm$ 2.7 \\
\rowcolor{magenta!15}
SPIRIT-TS (\textit{Ours}) & 96.0 $\pm$ 1.2 & 73.8 $\pm$ 1.4 & 50.5 $\pm$ 3.3 & 50.7 $\pm$ 2.9 & 49.8 $\pm$ 1.9 & 43.7 $\pm$ 2.7 \\

\bottomrule
\end{tabular}
}
\end{table*}

\begin{table*}[!t]
\centering
\caption{Training hyperparameter summary across compared models. We follow the training hyperparameters proposed by the original authors for each method. For model families with multiple variants (e.g., 224, 384, and ensemble settings), we report the shared/default training configuration at the family level. For CurConMix and SelfD, which use multi-stage SSL/teacher$\to$student pipelines, we report the final (student) stage.}
\label{tab:training_hparams_all}
\setlength{\tabcolsep}{6pt}
\renewcommand{\arraystretch}{1.15}
\resizebox{\textwidth}{!}{%
\begin{tabular}{lcccccccc}
\toprule
\textbf{Model} & \textbf{Backbone} & \textbf{LR Scheduler} & \textbf{Batch Size} & \textbf{Optimizer} & \textbf{Weight Decay} & \textbf{Num Epochs} & \textbf{LR} & \textbf{Backbone LR} \\
\midrule
RDV         & ResNet18 & step & 32 & SGD & $1\times10^{-6}$ & 50 & $1\times10^{-1}$ & $1\times10^{-1}$ \\
RiT         & ResNet18 & step & 16 & SGD & $1\times10^{-5}$ & 50 & $1\times10^{-1}$ & $1\times10^{-1}$ \\

SelfD       & Swin & cosine & 64 & Adam & $1\times10^{-6}$ & 30 & $2\times10^{-4}$ & $2\times10^{-4}$ \\
TERL        & Swin & step & 16$^\S$ & SGD & $1\times10^{-5}$ & 20 & $1\times10^{-3}$ & $1\times10^{-3}$ \\
CurConMix   & Swin & cosine & 64 & AdamW & $1\times10^{-6}$ & 40 & $2\times10^{-4}$ & $2\times10^{-4}$ \\

SurgeNetXL  & CaFormerS18 & cosine & 16 & AdamW & $6\times10^{-2}$ & 40 & $1\times10^{-4}$ & $2\times10^{-5}$ \\
LemonFM     & ConvNeXt-L  & cosine & 16 & AdamW & $6\times10^{-2}$ & 30 & $1\times10^{-4}$ & $2\times10^{-5}$ \\
LemonFM-T   & ConvNeXt-L  & cosine & 16 & AdamW & $6\times10^{-2}$ & 30 & $1\times10^{-4}$ & $2\times10^{-5}$ \\

DinoV3-L    & ViT-L         & cosine & 16 & AdamW & $1\times10^{-4}$ & 30 & $1\times10^{-4}$ & $2\times10^{-5}$ \\
DinoV3-L-T  & ViT-L         & cosine & 16 & AdamW & $1\times10^{-4}$ & 30 & $1\times10^{-4}$ & $2\times10^{-5}$ \\
SPIRIT-S    & DinoV3-L-txt  & cosine & 16 & AdamW & $1\times10^{-4}$ & 30 & $1\times10^{-4}$ & $2\times10^{-5}$ \\
SPIRIT-TS   & DinoV3-L-txt  & cosine & 16 & AdamW & $1\times10^{-4}$ & 30 & $1\times10^{-4}$ & $2\times10^{-5}$ \\
\bottomrule
\end{tabular}%
}
\vspace{2pt}
\footnotesize{$^\S$TERL: batch size 4 due to memory constraints at 384px.}
\end{table*}